\documentclass{article}

\usepackage[british]{babel} 

\usepackage{media9}
\usepackage{graphicx}
\usepackage{pdflscape}

\usepackage[letterpaper,top=2cm,bottom=2cm,left=3cm,right=3cm,marginparwidth=1.75cm]{geometry}

\usepackage[T1]{fontenc}
\usepackage[utf8]{inputenc}

\usepackage{tipa}

\usepackage{amsmath}
\usepackage{bm}
\usepackage{graphicx}
\usepackage{hyperref}
\usepackage{float}
\usepackage{subcaption}
\usepackage{siunitx}
\usepackage{enumitem}
\newcounter{example}

\usepackage{xcolor}

\usepackage{pdflscape} 
\usepackage{booktabs} 

\usepackage[utf8]{inputenc}
\usepackage[T1]{fontenc}
\usepackage{lmodern}
\usepackage{rotating}

\usepackage[sort]{natbib}
\usepackage{kbordermatrix}

\usepackage[utf8]{inputenc} 
\usepackage[T1]{fontenc}   

\title{Polish phonology and morphology through the lens of distributional semantics}
\author{Paula Orzechowska \& R. Harald Baayen\\ \ \\
Adam Mickiewicz University \& University of T\"{u}bingen}
\begin{document}
\maketitle

\begin{abstract}

\noindent
This study investigates the relationship between the phonological and morphological structure of Polish words and their meanings using Distributional Semantics.  Polish is characterized by a rich set of consonant clusters, many of which are triggered by the intervention of morphology.  Previous research has focused on the study of the phonological properties of clusters, their constituent morphemes and parsability.  In the present analysis, we ask whether there is a relationship between the form properties of words containing such clusters and their meanings.  More specifically, is not only the morphological, but also the phonological and morphonological structure of complex words mirrored in semantic space? Can the grammatical functions of morphemes and their decomposability be inferred from semantic vectors without explicitly consulting the morphological structure of words? Can phonotactic patterns be modelled without consulting the phonemic representation of syllables? We address these questions for Polish, a language characterized by non-trivial morphology and an impressive inventory of morphologically-motivated consonant clusters. Instead of determining morpheme boundaries within a cluster, which has been the main goal of many previous studies, we use statistical and computational techniques, such as t-SNE, Linear Discriminant Analysis and Linear Discriminative Learning, and demonstrate that --- apart from encoding rich morphosyntactic information (e.g. tense, number, case) --- semantic vectors capture information on sub-lexical linguistic units such as phoneme strings. First, phonotactic complexity, morphotactic transparency, and a wide range of morphosyntactic categories available in Polish (case, gender, aspect, tense, number) can be predicted from embeddings without requiring any information about the forms of words. Second, we argue that computational modelling with the discriminative lexicon model using embeddings can provide highly accurate predictions for comprehension and production, exactly because of the existence of extensive information in semantic space that is to a considerable extent isomorphic with structure in the form space.  \\ \ \\


\end{abstract}

\section{Introduction}\label{sec:intro}

\citet{Martinet-1949-double-articulation, Martinet-1965-linguistique} characterized language as a system with dual articulation: a phonological system that accounts for the forms of the basic units of language, and a syntactic system that accounts for how these basic units are arranged in meaningful ways. Phonological units and their structure are seen as being subject to the constraints of the human vocal tract. The link between units of form and their meanings is seen as arbitrary \citep{DeSaussure-1966-course}. The syntactic system is construed as operating on abstract symbols, with rules specifying how form units are arranged into utterances, irrespective of the specifics of the meanings of these form units.  

In recent years, evidence has been accumulating that this strict division of labour between phonetics and phonology, on the one hand, and syntax, on the other, is artefactual.  One line of research suggests that the link between form and meaning is far less arbitrary, and that iconicity is much more prevalent than we thought \citep[see, e.g.][]{Dingemanse-blasi-2015-arbitrariness, Winter-etal-2024-iconicity,Varda-etal-2025-arbitrariness}.  Another line of research has provided evidence that the specifics of words' meanings co-determine their phonetic realization. \citet{Gahl-2008-time-thyme} and \citet{Lohmann-2018-cut-homophons} have shown that in the corpora of spontaneous speech English homophones have different distributions of spoken word durations. \citet{Gahl-baayen-2024-time-thyme}, using embeddings from distributional semantics and a computational model that will be introduced in more detail below, have recently demonstrated that differences in homophones' spoken word duration can be predicted to a considerable extent by the extent to which a homophone's form is supported by its meaning.  Furthermore, \citet{Plag-homann-kunter-2017-homophony} have shown that the distribution of the durations of word-final /s/ in English varies with the semantics (e.g. non-morphemic, genitive plural, noun plural) that the suffix /s/ realizes.  Tone in Mandarin Chinese and stress in English have likewise been found to co-vary systematically with word meaning \citep{Chuang-etal-2023-do-words-sing, Chuang-etal-2025-tone-mand, lu2025realization}.  These empirical findings challenge the axioms of the arbitrariness of the sign and the dual articulation of language, and linguistic theories that build on these axioms.

In this study, we provide further evidence challenging these axioms by documenting surprisingly strong systematic correspondences between form and meaning for two datasets, each featuring several thousand Polish words with complex phonotactics and morphology.  In what follows, we first provide a detailed description of the phonology, morphology, and morphonology of Polish (Section \ref{sec:intro_phon&morph}).  Subsequently, we show that the Discriminative Lexicon Model \citep[DLM]{Baayen-etal-2019-DLM}, a computational model that predicts meaning from form (comprehension) and form from meaning (production), performs with high accuracy without being informed about stems, morphemes, inflectional classes, and morphosyntactic features (Section \ref{sec:modelling_dlm}).  The question of how this is possible is addressed in the next part of our study.  Analyses using Linear Discriminant Analysis and t-distributed Stochastic Neighbour Embeddings reveal that the morphosyntactic features (case, number, gender, person, aspect, tense), the phonotactic features (size and phonological markedness of word-initial consonant clusters, phonological characteristics of a consonantal prefix), and the morphonotactic features (parsability of a prefix) are predictable from their embeddings with accuracies that are far above majority baselines (Sections \ref{sec:analyses} and \ref{sec:orthogonalization_fastText}).  In the general discussion (Section \ref{sec:gendisc}), we argue that the reflection of morphosyntax and (mor)phonotactics in semantic space are the reason why the DLM can perform so well.  We also suggest that these pervasive similarities may help explain why recent deep learning models of language \citep[e.g.][]{Vaswani-etal-2017-attention} have been so successful.

\section{Polish phonology and morphology}\label{sec:intro_phon&morph}   

Central themes in the study of word structure have been the words' phonological (phonotactic) and morphological (morphotactic) patterns within words \citep[for an overview, see][]{Taylor-2015-word-hanbook}). The first set of patterns is mainly related to phonotactics as the most fundamental form of the syntax of words \cite[e.g.][]{Harris-1954-distributional-structure, Trask-1996-dict}. The second set of patterns accounts for the admissible ordering of morphemes in words and their phonological properties \cite[e.g.][]{Booij-2001-LP, Halle-1959-sound-pattern-rus, Ewen-hulst-2001-words, Hay-2003-causes}.  Both types of constraints  contribute to the complexity of words. In  particular, the intervention of morphology tends to trigger complex phonological sound sequences, which have been observed for many phonotactically rich languages \cite[for an overview, see][]{KDK-etal-2026-CHNL}.

The languages of the world display restrictions on the combinations of sounds, with the most severe constraints holding for sequences of consonants at syllable and word margins.  A universally unmarked monosyllable is composed of a single consonant (C) and a vowel (V) \citep{Greenberg-1966, Donohue-2013-phonotactic-database, Gordon-2016-typology}.  The cross-linguistic preference for CV is phonetically motivated: the most salient acoustic modulations are found at the C---to---V interface \citep{Ohala-1990}, which supports articulatory precision \citep{Ohala-kawasaki-1984-prosodic-phon}, and facilitates perceptual contrast \citep{KDK-2002-b&b, KDK-2014-explaining}.  CV violations can lead to phonologically marked clusters as several adjacent consonants are likely to violate universal principles of well-formedness.  To take English syllable onsets as an example, phonotactic constraints specify the ordering of plosives and liquids such that /tr dr/ are admissible in \textit{train} and \textit{drain}, while their reverse ordering */rt rd/ is prohibited.  Such constraints also limit the post-plosive position to the rhotic, eliminating plosive + lateral */dl tl/ onsets. 

In contrast to English, Polish has been mentioned among languages with notoriously complex phonotactics \citep{Maddieson-2013-WALS}, as it allows strings of multiple consonants whose ordering is subject to comparatively fewer constraints \citep[e.g.][]{Bargielowna-1950, Dobrogowska-1990-internal, Dobrogowska-1992-initial-final, Dunaj-1986-wyglosowe, Orzechowska-2019-complexity, Zydorowicz-etal-2016-morphonotactics}.  Phonotactic complexity in Polish has been explored in studies on \textit{inter alia} syllabification \citep[e.g.][]{Bertinetto-etal-2007-intersegmental, Bethin-1992-Polish-syllable, Cetnarowska-zygis-2004-syllabification, Cyran-gussmann-1999-clusters-GP, Gussmann-1992-resyllabification, Rubach-booij-1990-syllable-pl, Szpyra-2000-syllabification-pl}, gestural coordination and timing \citep[e.g.][]{Hermes-etal-2017-berber-pl, Mucke-2010-articulation-clusters-pl, Schwartz-2024-ema-nad-pl}, language acquisition \citep[e.g.][]{Jarosz-2017-defying-the-stimulus-pl, Jarosz-etal-2017-input-freq-pl, lukaszewicz-2010-book, Marecka-KDK-2014-evaluating, Milewski-2005-grupy, Schwartz-2023-sC-pl-en, Yavac-marecka-2014-acquisition, Zydorowicz-2019-pl}, and online processing \citep[e.g.][]{Orzechowska-2019-complexity, Wagner-etal-2012-cluster, Wiese-etal-2017-EEG-pl}. 

Greater phonotactic permissibility in Polish can be illustrated by the rich set of plosive + liquid onset sequences. In addition to /tr dr/ in \textit{trawa} `grass' and \textit{droga} `way', /t d/ may also precede the lateral, as in \textit{tlen} `oxygen' and \textit{dlaczego} `why'.  The reverse --- universally marked --- ordering is attested as well, though only in a handful of examples such as \textit{rtęć} `mercury' and \textit{rdest} `Polygonum' (a plant).  Even more strikingly, Polish allows hundreds of longer cluster types at syllable margins. This holds also for word-final position, which is prosodically weaker --- and thus more prone to phonotactic erosion and reduction ---- and which, in contrast to word-medial position, typically displays more stringent constraints. Among some more complex clusters in Polish are the initial CCCC /g\textipa{\textyogh}bj/ in \textit{grzbiet} '(horse) back' and final CCC /r\textipa{\textesh}\textipa{\textteshlig}/ in \textit{barszcz} `borscht'.  These examples show that the phonological universals constraining the number of consonants in a sequence  \citep{Gordon-2016-typology, Greenberg-1966} and their ordering \citep{Selkirk-1984-sonority} hold less rigorously in Polish than in English.\footnote{Note that such constraints have been primarily posited for the syllable. In the present work, we consider the word to be the least controversial domain for formulating phonotactic generalizations (cf. \cite{KDK-2002-b&b, Rubach-1996-nonsyllabic} for Polish; \cite{Steriade-1999-alternatives} for German), which makes it possible to investigate morphologically-motivated clusters.}   

One of the factors that contributes to cluster complexity in Polish is morphology.  There is an affinity between the phonological markedness of the clusters and their morphological composition.  In strongly inflecting languages, morphological operations can generate new, longer, and phonologically marked cluster types, which do not occur in morphologically simple words \citep[for comprehensive overviews, see][]{Dressler-KDK-2006, Dressler_KD_Pestal-2010-morphonotactics, KDK-etal-2026-CHNL}.  Polish provides an excellent testing ground for this observation.  For instance, the word initial cluster /skrf/ arises when the perfectivising prefix \{s-\} attaches to the verb \textit{krwawić} `to bleed'-imperf. $\rightarrow$ \textit{s+krwawić} `to bleed'-perf., and /sx/ emerges in the verb \textit{s+chodzić} `to go down(stairs)', where the prefix \{s-\} adds the semantics of directionality to the base verb \textit{chodzić} `to go, walk'. Regardless of its precise semantic contribution, the prefix increases cluster complexity in the same way. In both cases, the intervention of morphology overrides the phonological universals and principles by (1) increasing the number of adjacent consonants and (2) generating phonologically marked output.

Word-initial clusters containing this prefix have attracted the attention of phonotacticians, since /s/ creates marked clusters reminiscent of those in English words such as \textit{stay, scream, splash}. It is due to the fact that --- disregarding its perfectivising or derivational function --- the highly  productive \{s-\} exponent gives rise to more than 160 marked cluster types \citep[e.g.][]{Orzechowska-2019-complexity}.  Apart from \{s-\}, there are other non-syllabic consonantal prefixes such as \{z- w-\}, phonetically realized as /z f v/, and their combination \{ws-\}, phonetically realized as /fs vz/.  Generally, it has been argued that while \{s- z-\} can function as purely aspectual prefixes that do not contribute to the meaning of the derived verb \citep{Lazinski-2020-aspekt, Lazorczyk-2010-slavic-aspect} as in \textit{blogować} `to blog' $  \rightarrow $ \textit{z+blogować} `to put into a blog' \citep[see `pure perfectivizers' in][]{Lazorczyk-2010-slavic-aspect}, the \{w- wz-\}-prefixed forms can be semantically remote from the base form as in \textit{głębić} `to dig (a mine)' $ \rightarrow $ \textit{w+głębić} `to explore' and \textit{bić} `to beat' $\rightarrow$ \textit{wz+bić} `to rise'.  The latter prefixes can be also used in verbs with metaphorical meaning, frequently with a humorous or blunt tone \citep{Czarnecka-2014-czasownik} as in \textit{chrapać} `to snore' $ \rightarrow $ \textit{w+chrapać się} `to reach a goal with difficulty', \textit{szuflować} `to shovel' $ \rightarrow $ \textit{w+szuflować} `to have a big appetite' or in \textit{kopać} `to dig' $ \rightarrow $ \textit{w+kopać} `to frame sb'.  Irrespective of their grammatical function (for further details, see Section \ref{sec:aspect}), the \{s- z-\} and \{w- wz-\} prefixes give rise to increased structural complexity.  

In the present study, we do not consider the grammatical status of the prefixes but focus on their effects on word-initial phonotactics.  And their impact on the cluster structure is indisputable.  Comprehensive dictionary and corpus-based cluster lists in \cite{Orzechowska-2019-complexity} suggest that around 60\% of prefixed cluster types (also called morphonotactic types) occur exclusively in bimorphemic word onsets. For example, /fkr vgw/ are found only in prefixed words: \textit{kroić} `to cut'-imperf. $ \rightarrow $ \textit{w+kroić} `cut into (a bowl)', \textit{blednąć} `to turn pale'-imperf. $ \rightarrow $ \textit{z+blednąć} `to turn pale'-perf. A different class of cluster types consists of those with two realizations: /sp/ is purely phonotactic in \textit{spać} `to sleep' but morphonotactic \textit{s+padać} `to fall down'.  

Moreover, the prefixes can also contribute to certain grammatical and semantic complexities.  In addition to their function as perfectivizers in the aspectual system, \{s- z- w- wz-\} can also affect tense interpretation.  Perfective verbs in the present tense have future tense meaning as in \textit{robię} `I am doing' $ \rightarrow $ \textit{z+robię} `I will do', \textit{kradnie} `(s)he is stealing' $ \rightarrow $ \textit{s+kradnie} `(s)he will steal', \textit{piszecie} `you (pl.) are writing' $ \rightarrow $ \textit{w+piszecie} `you (pl.) will write down'.  Aspect is also marked on all verb types (including infinitive, imperative and impersonal) and deverbal parts of speech (participles and gerunds). 

The intervention of morphology can have an effect on the perception and processing of phonotactics depending on the morphological parsability of clusters.  Marked morphonotactic clusters, such as the ones listed above, can facilitate perception and processing because they are easily decomposable \citep{Hay-2000-causes-PhD, Hay-2002-affix-ordering, Hay-2003-causes, Korecky-etal-2014-morphonotactic-RT, Calderone-etal-2014-computational-morphonotactics} \citep[for counter-evidence, see][]{Celata-etal-2015-morphonotactics}.  Parsability is facilitated by (1) marked phonotactic patterns, as opposed to less parsable unmarked patterns \citep{Hay-2000-causes-PhD, Hay-2003-causes}, and (2) high frequency and productivity of affixes \citep{Hay-2003-causes, Hay-baayen-2003-phonotactics-productivity}.  In Polish, the prefixes are not only productive in deriving perfective verbs or verbs with modified meaning, but are --- by analogy --- carried to other parts of speech, as illustrated in \ref{list:prefixation productivity}:

\begin{enumerate}[resume]
  \item Productivity of prefixation in Polish\label{list:prefixation productivity}
    \begin{enumerate}
      \item  \textit{chodzić} `to walk' 
          
          $ \rightarrow $ \textit{ws+chodzić} `to rise', \textit{ws+chód} `sunrise', \textit{ws+chodni} `eastern',
      \item \textit{godzić} `to reconcile' 
          
          $ \rightarrow $ \textit{z+godzić się} `to agree', \textit{z+goda} `agreement', \textit{z+godny} `consensual',  \textit{z+godnie} `according to'
    \end{enumerate}
\end{enumerate}

In contrast, in lexicalised forms the processing of morphologically complex clusters can be less straightforward.  An example of such a form is \textit{lnu} `linen'-gen.sg., derived from \textit{len} `linen'-nom.sg. by stem vowel deletion.  Here, the omission of the vowel /e/ in the derived form (the so-called mobile /e/) is expected to hinder perception and increase processing costs \citep[e.g. as the violation of the biuniquess axiom in][]{Dressler-1985-morphonology-book,Dressler-1985-predictiveness-NP}.  

The ease or difficulty with which morphologically induced  clusters are processed is an empirical question that will not be pursued in the present work.  Instead, the analyses to follow focus on selected phonological characteristics of initial clusters, such as cluster size and markedness, and the morphonological aspect of cluster parsability \citep{Hay-2003-causes}.  The analyses to be reported in this paper add a new perspective to the accumulated knowledge on Polish by addressing the relation between (mor)phonology and meaning using distributional semantics and computational modeling.

\subsection{Polish (mor)phonotactics}\label{sec:(mor)phonotactics}

The complexity of consonant clusters in Polish has been studied from various perspectives.  The earliest accounts dating back to the 1950s \citep{Bargielowna-1950, Kurylowicz-1952-grupy} teased apart unmotivated and morphologically-motivated cluster types.  In \cite{Bargielowna-1950}, the unmotivated class comprises clusters found within a morpheme such as /sp/ in \textit{spać} `to sleep', while clusters in the morphologically-motivated class arise in affixation as in \textit{stać} `to stand' $ \rightarrow $ /fst/ in \textit{w+stać} `to get up'.  The latter type of clusters is further subdivided depending on the morphological productivity of a morpheme boundary.  Productive boundaries emerge in inflection (e.g. /pstf/ in \textit{hrab+stw} `county'-gen.pl.), while unproductive boundaries are lexicalised and thus no longer recognizable (e.g. /r\textipa{\textctc}\textipa{\texttctclig}/ in \textit{ga+rść} `hand(ful)'). 

As a follow up to the precursor work, \cite{Kurylowicz-1952-grupy} demonstrated the syntagmatic differences between motivated and unmotivated clusters by proposing a method of dividing CCC cluster types into shorter constituents.  In the motivated class, a bipartite structure of a cluster is marked by the presence of a productive morpheme such as in \textit{w+stać}, leading to the C|CC split.  In the unmotivated class, the splitting procedure involves identifying a CC that is attested in Polish.  For instance, the C|CC division represents a three-member cluster such as /m\textipa{\textctc}\textipa{\texttctclig}/ in \textit{mścić} `to avenge', in which the fricative + affricate combination /\textipa{\textctc}\textipa{\texttctclig}/ is an existent initial type (e.g. \textit{ścinać} `to cut', \textit{ścierka} `cloth'), while */m\textipa{\textctc}/ is not attested word-initially.  Another possibility of cluster division is CC|C represented by /krn/ in \textit{krnąbrny} `defiant', in which /kr/ is a legitimate word onset (e.g. \textit{krowa} `cow', \textit{kręcić} `turn'), in contrast to */rn/. 

A contemporary take on (un)motivated clusters has been proposed in \cite{Dressler-KDK-2006}, who introduced the term \textit{morphonotactics} to refer to the interaction between morphotactics and phonotactics.  Similarly to the previous approaches, a distinction is made between phonotactic clusters which occur morpheme-internally and morphonotactic clusters which are the product of morphology.  The classification is based on the degree of deviation of morphonotactic clusters from purely phonotactic ones, where the extremes of the continuum are occupied by exclusively morphonotactic and exclusively phonotactic clusters. 

In this approach, the identification of morpheme boundaries in languages such as English is relatively straightforward, as word-medial and word-final morphonotactic cluster types typically arise from prefixation and suffixation, respectively (e.g. /bd/ in English \textit{sub+discipline, grabb+ed}) or in compounding  (e.g. /stm/ in \textit{post+man}).  Howeverr, in Polish morphonotactic clusters can originate from two distinct sources: concatenative and non-concatenative, as outlined in~\ref{list:Morphological sources of initial clusters in Polish}.  The first type covers cases of overt marking (\ref{list:Concantenative morphology}), as discussed above.  The non-concatenative source, in turn, serves as an umbrella term for various morphological operations that involve vowel elision (\ref{list:Vowel-zero alternation}).

\begin{enumerate}[resume]
  \item Morphological sources of initial clusters in Polish\label{list:Morphological sources of initial clusters in Polish}
    \begin{enumerate}
      \item Concatenative morphology\label{list:Concantenative morphology}
        \begin{enumerate}
          \item /zgw/ results from prefixation:
          
          imperfective \textit{głosić} /gwo\textipa{\textctc}i\textipa{\texttctclig}/ `to preach' → perfective \textit{z+głosić} /zgwo\textipa{\textctc}i\textipa{\texttctclig}/ `to notify',
          
          where
          
           \{głos-\} is a stem
          
          \{z-\} is a prefix
          
          \{-i-\} is a theme vowel
          
          \{-ć\} is an infinitival suffix
          
          \item /pxn/ results from suffixation:
          
          imperfective \textit{pchać} /pxa\texttctclig/ `to push' → perfective \textit{pch+nąć} /pxno\textipa{\textltailn}\textipa{\texttctclig}/, 
          
          where

          \{pch-\} is a stem
          
          \{-a-\} is a theme vowel
          
          \{-ną-\} is a semelfactive suffix
          
          \{-ć\} is an infinitival suffix
          
        \end{enumerate}
      \item Vowel-zero alternation\label{list:Vowel-zero alternation}
    \begin{enumerate}
          \item /krfj/ results from vowel loss in word-formation:
      
      noun \textit{krew} /kref/ `blood' → adjective \textit{krwionośny} /krfjono\textipa{\textctc}ny/ `circulatory' 
      
          \item /mx/ results from vowel loss in inflection (case):
          
      nominative \textit{mech} /mex/ `moss' → genitive \textit{mchu} /mxu/
      
      \item /ps/ results from vowel loss in inflection (number):
          
     singular \textit{pies} /pjes/ `dog' → plural \textit{psy} /ps\textipa{\textbari}/
      
        \end{enumerate}
    \end{enumerate}
\end{enumerate}

\noindent
\ref{list:Concantenative morphology} lists the contexts that apply to word-onsets, although rich (non-)concatenative morphology is also found word-finally \cite[for full exposition, see][]{Dressler-KDK-2006, Orzechowska-2019-complexity, Zydorowicz-etal-2016-morphonotactics}.  Clusters in~\ref{list:Vowel-zero alternation} exemplify three typical non-concatenative sources, namely the derivation of adjectives from nominal bases (i) and noun declension (ii, iii), in which the stem-medial alternating vowel (the mobile /e/) is lost.  Similarly, a word starting with a single consonant, such as \textit{lew} `lion', gives rise to cluster-initial inflected forms such as /lv/ \textit{lwy} `lion'-nom.pl., and derivatives such as /lvj/ \textit{lwie} `lion-like'-nom/acc/voc.sg.  In the present paper, we focus on the first --- and the most productive --- context summarized in~\ref{list:Concantenative morphology}, which is discussed in detail in the following section.

\subsection{Consonantal prefixes as triggers of morphonotactics}\label{sec:aspect}

Polish realizes imperfective and perfective aspect on verbs, gerunds and participles.  The imperfective construction essentially expresses an ongoing action or state, while the perfective expresses their completion. In verbs, aspect can be signalled by different suffixes (e.g. \textit{pch+ać} `to push' $ \rightarrow $ \textit{pch+nąć} `to push once'), stem alternation (e.g. \textit{oddychać} `to breathe' $ \rightarrow $ \textit{odetchnąć} `to catch one's breath'), or suppletive stems (e.g. \textit{brać} $ \rightarrow $ \textit{wziąć} `to take', \textit{mówić} $ \rightarrow $ \textit{powiedzieć} `to say').  
However, most verbs form their perfective aspect through prefixation, as illustrated by the verb \textit{pisać} ‘to write’ in~\ref{list:Prefixes forming perfective forms of pisać}.  Out of 18 prefixal perfectivizers available in Polish, the verb takes 13 \citep[pp. 25ff]{Lazorczyk-2010-slavic-aspect}. 

 \begin{enumerate}[resume]\label{ex:3}
  \item Prefixes forming perfective verb forms of the verb \textit{pisać} `to write'\label{list:Prefixes forming perfective forms of pisać}
    \begin{enumerate}
      \item \textit{do+pisać} ‘to write something additional’ or `to be favourable' \label{list:dopisać}
      \item \textit{na+pisać} ‘to write down'
      \item \textit{od+pisać} `to copy (homework)'
      \item \textit{o+pisać} `to describe'
      \item \textit{po+pisać} ‘to write a little’ or `show off'
      \item \textit{pod+pisać} ‘to sign’
      \item \textit{prze+pisać} ‘to re-write’ or ‘to prescribe’
      \item \textit{przy+pisać} ‘to ascribe’
      \item \textit{roz+pisać} ‘to make a schedule’
      \item \textit{s+pisać} ‘to make a list of’ or `to copy down' \label{list:spisać}
      \item \textit{w+pisać} ‘to write in’ \label{list:wpisać}
      \item \textit{wy+pisać} `to write out' or `to release (from hospital)'
      \item \textit{za+pisać} `to prescribe' \label{list:zapisać}
    \end{enumerate}
\end{enumerate}

Apart from the main function of expressing temporal definiteness and completion \citep[compare with other Slavic languages in][]{Dickey-1997-aspect-slav}, consonant prefixes also lead to a change in word meaning. For example, \{od-\} `from', \{pod-\} `under' and \{w-\} `into' suggest the type or direction of the process of writing: writing a text from somebody (i.e. copying), writing one's name underneath a text (i.e. signing), or writing into a text (for further discussion on the semantics of the prefixes, see Section~\ref{sec:aspect meaning}). The only natural perfective form is \textit{na+pisać}, as the meaning of \{na-\} in combination with the meaning of the base \textit{pisać} is the closest to the original meaning of the base itself \citep{Dickey-janda-2015-aspect}.  

Traditionally, there has been a clear division of labour between prefixation and suffixation \citep[although see]{Divjak-etal-2024-aspect-pl}, such that grammatical aspect was expected to be supplied exclusively by suffixation, and word formation by prefixation \citep{Czochralski-1975-verbalaspekt, Grzegorczykowa-etal-1998-gramatyka}. In line with this view, \textit{pch+ać} $ \rightarrow $ \textit{pch+nąć} in~\ref{list:Concantenative morphology} constitutes a true aspectual pair, as the two forms are binary grammatical variants of the same lexeme, differing only in their perfective/imperfective value. In contrast, the verbs listed in~\ref{list:dopisać} through~\ref{list:zapisać} involve an additional semantic shift associated with the prefix and would thus be listed in a dictionary as 13 distinct entries derived from the base verb. A corollary to this view is that aspect cannot be unequivocally regarded as being a clearly derivational (prefixal) category or an inflectional (suffixal) category \citep{Aalstein-blackburn-2006-aspect-pl}.  

In this paper, we refrain from making claims about the grammatical status of the prefixes under scrutiny and the place of (aspectual) prefixes in Polish grammar. We adopt a prevalent view --- also followed in grammar coursebooks, dictionaries, and the corpus used in the analyses below --- that aspect is a grammatical phenomenon, whereby perfective and imperfective verb pairs have their own infinitive forms listed as separate lexical entries \citep[e.g.][]{Banko-2011-wyklady, Grzegorczykowa-etal-1998-gramatyka}. An aspectual prefix is selected by a verb via a somewhat haphazard prefix choice \citep[compare true aspectual pairs: \textit{wierzyć}  $ \rightarrow $ \textit{u+wierzyć}, \textit{robić}  $ \rightarrow $ \textit{z+robić}, \textit{czytać}  $ \rightarrow $ \textit{prze+czytać}, after][]{Mlynarczyk-2004-aspectual-pairs}, although /s z/ have been argued to be more common than other prefixes and are associated with the so-called natural perfectives \citep{Dickey-2005-s-z-aspect-slavic, Dickey-janda-2015-aspect, Janda-2007-aspect-rus}, or true aspectual pairs \citep{Mlynarczyk-2004-aspectual-pairs}. In the case of \textit{pisać}, the only true perfective aspectual twin is \textit{na+pisać}, as the prefix simply suggests the attainment of a result of the action of writing \citep{Czochralski-1975-verbalaspekt}. 

Additionally, to account for more complex forms, linguists have supplemented such true pairs with the so-called secondary imperfectives, i.e. forms that employ an infix to derive an imperfective verb from a prefixed verb with a perfective meaning, e.g. \textit{angażować} `to employ'-imperf $ \rightarrow $ \textit{za+angażować}-perf. $ \rightarrow $ \textit{za+angażo+wy+wać}-secondary imperfective \citep{Divjak-etal-2024-aspect-pl}. 

The discussion of the perfective/imperfective pairs cannot be abstracted from that of tense, especially in light of recent research. \citet{Divjak-etal-2024-aspect-pl} have shown that the majority of verbs (90\% of their dataset) exhibit a bias towards either the perfective form or the imperfective form.  This corpus study has further revealed, for example, that \textit{chcieć} 'to want' typically occurs in the imperfective form regardless of tense, while \textit{robić} 'to do' strongly gravitates towards the imperfective in the present tense, but the perfective in the past and future tenses.

In the present study, the division of word forms into perfective and imperfective is based on a tagging convention that treats prefixed verbs, gerunds, and participles as representing the same lexeme, disregarding whether aspect is conveyed by a suffix or not (see Section~\ref{sec:data}). Following corpus-based work of \cite{Lazinski-2020-aspekt}, we assume that prefixation plays a greater role in forming aspectual pairs in Polish than suffixation \citep[although no parallel pattern was observed in other Slavic languages such as Russian, see][]{Janda-lyashevskaya-2011-aspectual-pairs-rus}. 

In spite of a variety of aspectual exponents, our analyses focus on non-syllabic consonants \{s- z- w-\} and their more complex variant \{wz-\}, as they supply a wide range of morphonotactic word onsets. \footnote{Note that initial /s/ can also function as a non-morphemic segment, leading to the coincidence in semantically and etymologically unrelated words such as \textit{port} `port' – \textit{sport}, \textit{kok} `hair bun' – \textit{skok} `jump', \textit{lew} `lion' – \textit{zlew} `sink', \textit{wędzić} `to smoke' – \textit{swędzić} `to itch'.} The prefixes are phonetically realized as non-syllabic consonants /s z f v/ and bi-consonantal sequences/fs vz/. The variant depends on the voice characteristics of the stem-initial segment, whereby /v z vz/ attach to voiced consonants and /f s fs/ to voiceless ones.  As evidenced in~\ref{list:spisać} and~\ref{list:wpisać}, the verb \textit{pisać} can take two consonant-only prefixes, resulting in initial clusters /sp/ and /fp/, which are phonologically marked (for a detailed exposition, see Section~\ref{sec:aspect phonotactics}). The unattested form with a more complex prefix represents an incidental gap: Although the emergent cluster /fsp/ occurs in Polish in other words, the absence of *\textit{ws+pisać} may be due to semantic constraints \citep[see for defectiveness in Russian nominal paradigms][]{Chuang-etal-2023-defective-nouns-rus}. 

In Sections~\ref{sec:aspect meaning} and~\ref{sec:aspect phonotactics}, we discuss the semantics of \{s- z- w- wz-\}, and the phonotactic possibilities that the prefixes contribute to.

\subsection{The semantics of prefixes}\label{sec:aspect meaning}
 
Aspectual prefixes can function as prepositions and add the prepositional meaning to the perfective verbs.  The core meaning of \{w-\} is `in'. The prefix specifies horizontal movement `into' (e.g. \textit{sunąć} `to glide' $ \rightarrow $ \textit{w+sunąć} `to insert', \textit{prowadzić} `to lead' $ \rightarrow $ \textit{w+prowadzić} `lead in'), and vertical movement `upward, up to the top' (e.g. \textit{skoczyć} `to jump' $ \rightarrow $ \textit{w+skoczyć} `jump onto', \textit{tachać} `to carry with difficulty' $ \rightarrow $ \textit{w+tachać} `to carry up with difficulty'). A more general meaning of the prefix is defined as an activity directed towards oneself or a given point \citep{Mizerski-2000-tabele}, an act of approaching the center of an object or reaching this goal \citep{Czarnecka-2014-czasownik, Smiech-1986-derywacja-czasow, Wrobel-1998-czasownik} (e.g. \textit{ładować} `to load' $ \rightarrow $ \textit{w+ładować} `to load a lot of something', \textit{bić} `to beat' $ \rightarrow $ \textit{w+bić} `to place something inside by beating'). However, it must be noted that depending on the context, the same perfective verb can suggest different types of movement: \textit{w+jechać} from \textit{jechać} `to drive' can be used in \textit{wjechać do garażu} `to drive into the garage' as well as in \textit{wjechać na trzecie piętro} `take a lift to the third floor'.

The basic meaning of \{s- z-\} is `from, with', expressing the completion of the centripetal movement. The prefixes generally point to objects or their fragments approaching each other or joining together \citep{Czarnecka-2014-czasownik, Smiech-1986-derywacja-czasow, Wrobel-1998-czasownik} (e.g. \textit{tulić} `to hug' $ \rightarrow $ \textit{s+tulić} `to put together', \textit{szyć} `to sew' $ \rightarrow $ \textit{z+szyć} `to stitch, sew together'). In some contexts, the  prefixes can suggest moving something away from the surface of an object \citep{Smiech-1986-derywacja-czasow, Wrobel-1998-czasownik} (e.g. \textit{myć} `to wash' $ \rightarrow $ \textit{z+myć} `remove something by washing'). They also express vertical movement downward (e.g. \textit{z+jechać do garażu} `to drive into the garage (underground)', \textit{z+jechać na parter} `to take a lift to the ground floor'), or movement from the upper surface of an object (e.g. \textit{z+lecieć} `to fly down', \textit{s+paść} `to fall down') \citep{Lazinski-2020-aspekt, Mizerski-2000-tabele}. 

In contrast to \{z- w-\}, \{wz-\} cannot function synchronically as a preposition. This bi-consonantal prefix also conveys more complex meaning. It indicates upward movement (\ref{list:wschodzić} and~\ref{list:wznieść}), as well as an increase or intensification of an action (\ref{list:wzrosnąć} and~\ref{list:wzbogacić}) \citep{Lazorczyk-2010-slavic-aspect}. The verb \textit{wz+rosnąć} suggests an augmentative action from its beginning to its effective execution. \textit{Wz+bogacić} indicates an action, whose results exceeds the previous state \citep{Agrell-1918-czasownik}. Note that the semantic load of the complex prefix limits its use to a small subset of verbs.

\begin{enumerate}[resume]
  \item Semantics of complex prefixes \label{list:Semantics of complex prefixes}
    \begin{enumerate}
      \item \textit{chodzić} `to walk'
      
      /sx/ \textit{s+chodzić} `to go down'
      
      /fx/ \textit{w+chodzić} `to enter' or `to go up'
      
      /fsx/ \textit{ws+chodzić} `to rise' (referring to the sun, the stars or planted vegetables) \label{list:wschodzić}
      
      \item \textit{nieść} `to carry'
      
      /z\textltailn/ \textit{z+nieść} `to carry down'
      
      /v\textltailn/ \textit{w+nieść} `to carry into (a flat)' or `to carry up'
      
      /vz\textltailn/ \textit{wz+nieść się} `to rise up, to soar' \label{list:wznieść}
     
      \item \textit{rosnąć} `to grow'

      /vzr/ \textit{wz+rosnąć} `to shoot up' or `to escalate' \label{list:wzrosnąć}
    
     \item \textit{bogacić się} `to grow rich'

     /vzb/ \textit{wz+bogacić się} `to grow rich, to enrich' \label{list:wzbogacić}

    \end{enumerate}
\end{enumerate}

Despite the apparent complexity of the aspectual system in Polish, there is a tightly-knit relationship between between the semantics of a verb and the type of prefix it selects. The analyses presented in Section~\ref{sec:analyses} will demonstrate that a word's meaning can be predicted with high accuracy by computational models and that the predictions of these models largely capture the semantics of prefixed forms.  

At this point, let us inspect another level of complexity of prefixed words, namely their phonotactics. 

\subsection{The phonological markedness of prefixed words}\label{sec:aspect phonotactics}

Another aspect of complexity induced by prefixation is an increase in cluster size and --- as a consequence --- cluster markedness. In prefixed word forms, the number of consonants found in prefixed words exceeds the number of consonants found in bare stems, resulting in the violation of phonological universals and well-formedness principles. For example, just the very prefix \{wz-\} represents a relatively marked structure in phonological terms. The distinction between marked and unmarked \citep{Battistella-1990-markedness, Jakobson-1941-kindersprache, Lacy-2006-markedness, Trubetzkoy-1939} phonotactics has traditionally been determined by the adherence of clusters to sonority-based ordering. Consonants can be placed on a scale of relative loudness (sonority) depending on their manner of articulation. In the exemplary sonority scale of \cite{Goldsmith-1990-autosegmental}, the most sonorous end of the scale is occupied by vowels, followed by glides (e.g. /j w/), liquids (e.g. /r l/), nasals (e.g. /m n/), fricatives (e.g. /f z/), affricates (e.g. /\textipa{\textteshlig} \textipa{\textdzlig}/), and finally plosives (e.g. /p d/). The distribution of the consonants in an unmarked syllable is determined by the Sonority Sequencing Generalization \cite{Selkirk-1984-sonority} \citep[for overviews, see e.g.][]{Cairns-raimy-2011-handbook-syllable, Parker-2012-controversy, Parker-2017-compass}, which states that in an articulatorily felicitous syllable, sonority should decrease from the vowel outward. In other words, an onset cluster is required to display a sonority rise from the first consonant towards the vowel, as in plosive $+$ liquid /kl/, fricative $+$ nasal /sn/, and plosive $+$ glide /tw/.  Sequences violating this ordering, exemplified by the fricative $+$ stop ($+$ C) as in /sp str skw/, are considered marked. \footnote{The prefixed fricatives are classified as appendices in some theoretical approaches \citep[e.g.][]{Gussmann-2007-phonology, Rochon-2000-OT, Rubach-1997-extrasyllabic, Vaux-wolfe-2009-appendix}.}

The phonetic variants of the prefix \{wz-\}, /fs vz/, represent yet another class of clusters, as the constituent consonants display an identical sonority level. Such plateaus are universally considered marked due to a lack in sonority differential. However, in Polish obstruent onsets do not require a difference in sonority \citep{Rochon-2000-OT, Rubach-booij-1990-syllable-pl}, which explains why some phonologists consider /pt ss/ in \textit{ptak} `bird' and \textit{ssak} `mammal' as unmarked. Since we make no claims concerning the markedness status of these cluster types, in the present study we make use of the general categories, namely rise, fall, and plateau. This classification is also important for another reason. Evaluating the markedness of longer clusters is problematic when constituent sub-clusters display various sonority slopes. The most extreme example of such a long cluster is /fskr/ composed of a plateau /fs/, a fall /sk/ and a rise /kr/. Thus, sequences such as /sxr fst zm\textltailn/ are here conveniently classified as plateau $+$ rise, plateau $+$ fall, and rise $+$ plateau, respectively.

Table~\ref{tab:Sonority complexity of prefixed verbs} provides an overview of phonological characteristics of initial morphonotactic clusters by cluster size\footnote{Note that initial clusters are also created when the suffix \{-nąć\} is attached to the stem. /px/ in \textit{pchać} `to push' changes to /pxn/ in the perfective form \textit{pch+nąć}, and to /px\textltailn/ in the nominalized form \textit{pch+nięcie} `a push', resulting in morphologically-motivated non-/s/-initial clusters.}. The presentation is limited to initial CC and CCC clusters, as the description of longer sequences becomes increasingly complex. Column 'rise' lists examples of morphonotactic clusters displaying a homogeneous rise in sonority (a positive unmarked slope, 'fall' --- a homogeneous fall (a negative marked slope), and 'plateau' --- a constant value in sonority. In Table~\ref{tab:Sonority complexity of prefixed verbs}, hyphens indicate cluster types that are unattested in Polish verbs. These gaps arise due to the voice agreement constraint that operates between the prefix and the stem-initial consonant. For example, to form a sonority-compliant cluster beginning with /s f/, the fricatives should be followed by a (voiced) sonorant.

\begin{sidewaystable}
\centering
\begin{tabular}{ |c||c|c|c|c| } 
\hline
cluster &prefix  &rise                      &plateau                            &fall\\
length  &        &&                        &(+plateau)\\\hline
CC      &/s/     &-                         &/s\textesh/ z+szyć `to sew'        &/sp/ s+pić `to drink'\\
CC      &/z/     &/zr/ z+robić `to do'      &/zv/ z+ważyć `to weigh'            &/zdz/ z+dzwonić (się) `to call'\\
CC      &/f/     &-                         &/fs/ w+sypać `to pour into'           &/fp/ w+padać `to fall into'\\
CC      &/v/     &/vn/ w+nosić 'to carry inside' &/v\textyogh/ w+rzucać 'to throw in' &/vb/ w+bić `to beat into'\\\hline
CCC     &/s/     &-                         &/sxf/ s+chwytać `to capture'       & - \\
CCC     &/z/     &/zmr/ z+mrozić `to freeze' & - & - \\
CCC     &/f/     &-                         &/fsx/ ws+chodzić `to rise'& - \\
CCC     &/v/     &/wmj/ w+mieszać `to mix into' & /vzv/ wzwyczaić się `to get used to' (archaic)                              & - \\\hline
\end{tabular}
\caption{\label{tab:Sonority complexity of prefixed verbs}Examples of verb-initial clusters representing cases of a clear sonority rise, plateau and fall.}
\end{sidewaystable}

In the computational experiments reported below, it will be demonstrated that the phonological properties of clusters such as cluster size and markedness can be predicated far above the chance level from words' semantics.  

Although all the clusters to be analysed are the product of prefixation, the presence of a morpheme boundary is not always straightforward. In the examples in~\ref{list:Concantenative morphology}, we have observed morphonotactic clusters with a clearly identifiable morpheme boundary. However, in some cases, the morphological decomposition of a cluster is less transparent, leading to a continuum of morphological parsability. 

Section~\ref{sec:transparency} presents a three-way division of morphonotactic clusters based on the separability of the prefix from the stem, and the ensuing strength of the morpheme boundary. 

\subsection{Parsability of morpheme boundaries}\label{sec:transparency}

The parsability of morpheme boundaries refers to the ease with which all sorts of affixes can be identified and separated from the stem.  This concept is grounded in the Parsability Hypothesis \citep{Bergsland-Vogt-1962-validity-comparative-method}, which emphasised the role of morpheme-boundary transparency (morphotactic transparency) in predicting diachronic change. \cite{Bergsland-Vogt-1962-validity-comparative-method} have stated that morphemes that are easily decomposable (i.e. parsable or transparent) are more likely to resist phonological reduction and loss over time. Subsequent linguists have adopted this idea to model the cognitive aspects of morphological processing. 

Ever since \cite{Taft-1979-prefix-stripping}'s Prefix Stripping model, morphological parsing has been widely regarded as an obligatory, critical initial stage in early word recognition. \cite{Hay-2000-causes-PhD, Hay-2002-affix-ordering, Hay-2003-causes, Hay-baayen-2003-phonotactics-productivity} proposed that morphological decomposition is a more gradient phenomenon, with complex words being more easily parsable when their base frequency is substantially higher than the whole-word frequency, on the one hand, and when the transitional probability at the morpheme boundary is lower, on the other hand. Building on this line of research, it has been further argued that easily parsable morphemes (1) facilitate morphological processing \citep[e.g.][]{Baayen-1992-transparency-frequency, Bybee-1985-morphology, Hay-2003-causes}, (2) reinforce patterns in the mental lexicon, contributing to higher productivity \citep[e.g.][]{Baayen-2009-morph-productivity, Bybee-2003-phonology-use}, and (3) are processed through generalizing rules, leading to more productive morphological processes \citep[cf. the dual-route model;][]{Prinz-1996-dual-route, Schreuder-baayen-1995-morph-processing}.

Drawing on previous work and following the morphological parsing convention adopted in \textit{The Grammar of Contemporary Polish} \citep{Grzegorczykowa-etal-1998-gramatyka} and \textit{The Word-Formation System of Contemporary Polish} \citep{Jadacka-2001-slowotworstwo}, \cite{Zydorowicz-etal-2016-morphonotactics} proposed a formal method to determine degrees of prefix parsability in Polish morphonotactic clusters. Their approach relies on several diagnostic rules. Rules 1 through 3, which correspond to the cases of “clear parsing,” “active stem” and “part-of-speech boundary transfer”, respectively, help to establish the morphological status of a cluster and assess the separability of the prefixes /f v s z fs vz/ from the stem. These rules can be understood as reflecting the cognitive steps involved in identifying a morpheme boundary within a complex onset. Table~\ref{tab:degrees of parsability} summarizes the cognitive cost associated with accessing a word, where (1) constitutes the context in which the prefix is most easily parsed, and (3) the context in which the prefix and the root are least separable.

\begin{table}[H]
\centering
\begin{tabular}{ |c|c|c| } 
\hline
    1. most parsable     &2. less parsable    &3. least parsable\\
     clear parsing       &active stem         &part-of-speech transfer\\\hline
\hline
         prefix +       &prefix +          &prefix + \\
         free base      &active bound root &bound root from another part of speech\\\hline
         s+kłamać `to lie' & w+płacać 'to pay in' & s+padochron 'parachute'  \\\hline
\end{tabular}
\caption{\label{tab:degrees of parsability}Degrees of parsability of a prefix.  Categories (1) through (3) indicate an increasing number of steps necessary to determine the presence of a morpheme boundary in an initial cluster.}
\end{table}

Rule 1 refers to cases of clear parsing, whereby the prefixes /f v s z fs vz/ attach to a free base. For instance, \textit{rodzić} `to give birth', \textit{kłamać} `to lie' and \textit{rosnąć} `to grow' exist on their own in the stock of Polish words, and --- when prefixed --- become perfective verbs with word onsets such as /zr/ \textit{z+rodzić}, /skw/ \textit{s+kłamać} and /wzr/ \textit{wz+rosnąć}.  The same simple separation rule holds for English words such as \textit{un+do, over+do, re+do}.  

A less parsable morpheme boundary is found at the juncture of a prefix and a bound root, similarly to English words containing the bound morpheme \textit{-mit}, e.g. \textit{commit, permit, submit}.  The term `active stem' in \cite{Zydorowicz-etal-2016-morphonotactics} refers to a rule used to identify roots that are productive in word formation. For example, \textit{*płacać} functions as an active stem: Although it cannot occur independently, it gives rise to a range of prefixed derivatives. In most cases, the meanings of these derivatives are strictly related to the root, e.g. \textit{w+płacać} `pay in', \textit{wy+płacać} `pay out', \textit{o+płacać} `pay (the bills)'.  However, such synchronic semantic proximity cannot always be easily established. The bound root \textit{*przątać}, for example, generates derivatives related both to the physical act of cleaning in \textit{s+przątać, wy+przątać} (= to free from dirt or pollution) and to mental activity in \textit{za+przątać} (`to be preoccupied with'). 

The least parsable context is represented by words in which the morpheme boundary can be determined neither by rule 1 nor by rule 2. In such morphotactically opaque examples, a morpheme boundary can be identified only when the target word is compared to its equivalents in other parts of speech.  That is, a morphological boundary identified in one part of speech is carried to another part of speech.  Let us take /sp/ in \textit{spadochron} `parachute' as an example.  The simple /s/-stripping rule of \cite{Taft-foster-1975-lexical-storage} and \cite{Taft-1979-prefix-stripping} does not work here as the root *\textit{padochron} does not exist in isolation. Neither does *\textit{padochron} represent an active root, from which numerous forms can be derived. The morphological status of /sp/ in this particular word can be determined only by analogy to the verb \textit{s+padać} `to fall down', which is straightforwardly parsable.  

As shown above, morphonotactic clusters in Polish display different degrees of parsability. There is evidence that some degrees of parsability of consonant combinations are reflected in phonetic realization. An ultrasound study in \cite{Orzechowska-etal-2022-usg} has shown that with an increase in the strength of morpheme boundary the stronger the morphological (or word) boundary, the weaker the regressive assimilation of consonants, with the strongest assimilation of consonants observed in the intramorphemic context. The question to be addressed below is whether computational models that do not work with stems and exponents, and hence make no attempt to parse words into their constituents, nor build up words from their constituents, can account for these findings. The only circumstance in which a model such as the Discriminative Lexicon might be able to capture these boundary effects is one in which parsability is reflected in semantic space (see Section~\ref{sec:analysis_parsability}). 

\subsection{Polish inflectional morphology}\label{sec:morphology}

Polish inflectional morphology is characterized by high degree of syncretism, and a wide range of irregular stem alternations. The challenge that such a system poses for computational modelling has been reported in \cite{Lubaszewski-2000-inflection-pl}. The proposed model codes morphological exponents and stem alternation rules, making it possible to successfully generate 47 verb forms, 44 adjective forms, up to 49 numeral forms, 14 nominal and pronominal forms, and three adverbial forms. Yet the generated paradigm omits to list numerous (de)verbal forms as it is unable to associate aspect with the orthographic form of a verb. As we shall see below, Polish features approximately 50 imperfective verb forms, 40 perfective verb forms and 80 forms of different participles. 

In this section, we offer a brief overview of Polish inflectional morphology, describing the morphological exponents of the various morphosyntactic categories, and document the language's complexity in terms of vast syncretism and richness of inflectional exponents in the paradigm of nouns, adjectives, adverbs, verbs and deverbal categories.

Nouns in Polish are inflected for number (singular and plural), case (nominative, genitive, dative, accusative, locative, instrumental and vocative) and gender (masculine, feminine and neuter), with individual exponents often simultaneously realizing multiple grammatical meanings. The gender system is particularly complex \citep[for an overview, see][]{Wierzbicka-2014-rodzaj}, as masculine nouns can be further subcategorized into three subgenders based on animacy: animate personal denoting human males, animate impersonal referring to male non-humans, and inanimate nouns. In the plural, the system becomes even more intricate, as masculine personal nouns form a distinct category opposed to all other masculine nouns, which are grouped together under a single non-personal plural gender class.

Case exponence in Polish nominal inflection is conditioned by gender and declension class, with animacy playing a key role in the masculine domain. Depending on a theoretical account, masculine nouns come in several declension patterns \citep[see e.g.][]{Grzegorczykowa-etal-1998-gramatyka, Jadacka-2001-slowotworstwo, Kowalik-2025-fleksja,Medak-2014-slownik-odm-rzecz} that correlate with the aforementioned animacy and with the phonological structure of the stem-final consonant (e.g. \textit{plecak} `rucksack', \textit{tort} `layercake', \textit{chrząszcz}'beetle'). Feminine nouns follow two major declension patterns, depending on whether the noun ends in a vowel (typically {-a} or {-i/-y}) or a (soft) consonant (e.g. \textit{droga} `path', \textit{pani} `lady', \textit{mysz} `mouse'). Neuter nouns likewise fall into distinct patterns: those ending in vowels {-o -e} decline differently compared to nouns ending in {-ę} or the borrowed suffix {-um} (e.g. \textit{lato} `summer', \textit{zdrowie} `health', \textit{cielę} `calf', \textit{millenium}). Table~\ref{tab:casenum_paradigm_ex} illustrates examples from these declension types, including animate \textit{matka} `mother', \textit{dziecko} `child', \textit{pan} `Mr', \textit{wilk} `wolf' and inanimate nouns \textit{radość} `joy', \textit{liceum} `secondary school' and \textit{miecz} `sword'.

\begin{landscape}
\begin{table}[H]
\centering
\begin{tabular}{ |c|c||c|c|c|c|c|} 
\hline
case & number & fem & neut & masc.pers. & masc.anim. & masc.inanim.\\\hline
\hline
Nom. & sg. & matka, radość & dziecko, liceum & pan & wilk & miecz\\\hline
Gen. & sg. & matki, radości & dziecka, liceum & pana  & wilka & miecza\\\hline
Dat. & sg. & matce, radości & dziecku, liceum & panu & wilkowi & mieczowi\\\hline
Acc. & sg. & matkę, radość & dziecko, liceum & pana & wilka & miecz\\\hline
Instr. & sg. & matką, radością & dzieckiem, liceum & panem & wilkiem & mieczem\\\hline
Loc. & sg. & matce, radości & dziecku, liceum & panu & wilku & mieczu\\\hline
Voc. & sg. & matko! radości! & dziecko! liceum! & panie! & wilku! & mieczu!\\\hline
\hline
case & number & fem & neut & masc.pers. & masc.non-pers. & masc.non-pers.\\\hline
\hline
Nom. & pl. & matki, radości & dzieci, licea & panowie & wilki & miecze\\\hline
Gen. & pl. & matek, radości & dzieci, liceów & panów  & wilków & mieczy\\\hline
Dat. & pl. & matkom, radościom & dzieciom, liceom & panom & wilkom & mieczom\\\hline
Acc. & pl. & matki, radości & dzieci, licea & panów & wilki & miecze\\\hline
Instr. & pl. & matkami, radościami & dziećmi, liceami & panami & wilkami & mieczami\\\hline
Loc. & pl. & matkach, radościach & dzieciach, liceach & panach & wilkach & mieczach\\\hline
Voc. & pl. & matki! radości! & dzieci! licea! & panowie! & wilki! & miecze!\\\hline
\end{tabular}
\caption{\label{tab:casenum_paradigm_ex} Case by number paradigm for three genders; feminine, neuter and masculine.  The latter gender is further divided into three animacy subgenders; masculine personal, masculine animate and masculine inanimate.}
\end{table}
\end{landscape}

Case and number syncretism leads to formally identical word forms across the paradigm of nouns, adjectives and verbs. This arises from suffixal exponents that are shared across declensional and conjugational patterns, producing extensive homophony in the inflectional system, as discussed in \cite{Banko-2011-wyklady}, \cite{Grzegorczykowa-etal-1998-gramatyka}, \cite{Mizerski-2000-tabele} and \cite{Nagorko-2006-gramatyka}. For instance, the word \textit{stali} is a genitive/dative/locative/vocative singular feminine form of the noun `steal' (\textit{stal}-nom.sg.), or a genitive plural form of the noun; a nominative/vocative plural adjective form meaning `stable' (\textit{stały}-nom.sg.). Moreover, \textit{stali} can also serve as a 3rd person plural masculine personal past verb `they stood' (\textit{stać}-inf.). This morphological syncretism highlights how Polish inflection can generate substantial cross-categorial homophony.

Similarly to the nominal paradigm, verbs also feature numerous homophonous forms.  Verbs are inflected for person (1st, 2nd and 3rd), number (singular and plural), gender (female, neuter and masculine), aspect (perfective and imperfective), tense (past, present and future), mood (indicative, imperative and conditional) and voice (active and passive).  The grammatical functions are primarily realised by suffixal exponents \citep{Dryer-2013-wals-suffixation}, as illustrated in Table~\ref{tab:chować_basic_forms}.  Only perfective forms are realized with prefixes.  

Perfective and imperfective verbs in Polish share the same inflectional morphology, with one major exception: perfective verbs lack true present-tense forms and therefore occur only in the past and future. Thus, verbs such as \textit{schować} form past such as \textit{schowałam} `I hid', and simple future siuch as \textit{schowam} `I will hide', but no present tense. Moreover, imperfective verbs can be alternatively expressed periphrastically by adding the verb \textit{być} `to be', conjugated in the future tense, e.g. \textit{będę} `I will', \textit{będziesz} `you will', \textit{będą} `they will'.  These properties of the aspectual system are expected to influence the distribution of perfective/imperfective forms in semantic space.  Indeed, in Section~\ref{sec:analyses}, we will show that t-SNE representations capture the interaction between tense and aspect. For example, for Russian, \cite{Janda-lyashevskaya-2011-aspectual-pairs-rus} observed that perfective verbs tend to be used in the past tense rather than in non-past tenses, while imperfective verbs preferentially occur in the present tense.  Similarly, \cite{Divjak-etal-2024-aspect-pl} found that 90\% of Polish verbs in corpora exhibit a strong bias toward one aspect over the other \citep{Divjak-etal-2021-exposure}. Such idiosyncratic properties of verbs pose challenges for computational modelling of Polish inflection. If aspect is lexeme-specific, and given the many stem alternations, the question remains whether the Discriminative Lexicon can learn the system. 

The number of unique inflected verb forms of a single lexeme can reach several dozen \citep[see also][]{Saloni-2000-koniugacja, Lubaszewski-2000-inflection-pl}. Table~\ref{tab:chować_basic_forms} lists the basic inflectional forms of the lexeme \textit{chować} `to hide' and shows that the paradigm of such forms spans close to one hundred paradigm cells for the imperfective alone. This count reflects the full range of categories expressed in Polish verbal morphology, including mood, tense, person, number, gender and aspect. We explicitly distinguish three genders in the singular, i.e. masculine, feminine, neuter, and two genders in the plural, i.e. masculine personal and masculine non-personal, the latter of which comprises all feminine and neuter plural forms.  

\begin{landscape}
\begin{table}[H]
\centering
\begin{tabular}{ |c||c|c|c|c|c|c| } 
\hline
verb form & mood & 	person &  number &  gender & tense&aspect\\\hline
chowa-m &  indicative & 	1 & sg & & pres & - \\\hline
chowa-sz	& indicative & 2	& sg & & pres & - \\\hline
chowa	& indicative &  3	& sg & & pres & - \\\hline
chowa-my	& indicative & 	1	& pl & & pres & - \\\hline
chowa-cie & indicative	&  2 & pl &	& pres & - \\\hline
chowa-ją & indicative & 3 & pl & & pres & - \\\hline
będę chowa-ł/ła & indicative & 1 & sg & m/f & fut & schowam\\\hline
będziesz chowa-ł/ła & indicative & 2 & sg	&  m/f & fut & schowasz\\\hline
będzie chowa-ł/ła/ło	&  indicative	&  3	&  sg & m/f/n & fut & schowa\\\hline
będziemy chowa-li/ły & 	indicative	&  1 & 	pl & m/f &  fut & schowamy\\\hline
będziecie chowa-li/ły & indicative	&  2 & pl &  m/f, n	&  fut & schowacie\\\hline
będą chowa-li/ły & 	indicative & 3	&  pl &  m/f, n	&  fut & schowają\\\hline
chowa-łem/łam & 	indicative	&  1 &  sg	&  m/f &  past & schowa-łem/łam\\\hline
chowa-łeś/łaś &  indicative & 2	&  sg & m/f	&  past & schowa-łeś/łaś\\\hline
chowa-ł/ła/ło	&  indicative & 3 & sg & m/f/n	&  past & schowa-ł/ła/ło\\\hline
chowa-liśmy/łyśmy &  indicative	&  1 & 	pl	&  m/f, n&  past & schowa-liśmy/łyśmy\\\hline
chowa-liście/łyście	&  indicative &  2 &  pl & 	m/f, n & past & schowa-liście/łyście\\\hline
chowa-li/ły & 	indicative	&  3 &  pl	&  m/f, n &  past & schowa-li/ł\\\hline
\hline
chowa-j & imperative & 2 & sg & &  & schowa-j\\\hline
niech chowa	&  imperative	&  3 & 	sg	& &  & niech schowa\\\hline
chowa-jmy & 	imperative	&  1 & 	pl	& &	 & schowa-jmy\\\hline
chowa-jcie	&  imperative	&  2 & pl &  &  & schowa-jcie\\\hline
niech chowa-ją & imperative	&  3 & 	pl	& &  & niech schowa-ją\\\hline
\hline
chowa-łbym/łabym	&  conditional	&  1 & 	sg	&  m/f & & schowa-łbym/łabym \\\hline
chowa-łbyś/łabyś	&  conditional	&  2 &	sg	& m/f & & schowa-łbyś/łabyś\\\hline
chowa-łby/łaby/łoby & conditional	&  3 & sg &	m/f/n & & schowa-łby/łaby/łoby\\\hline	
chowa-libyśmy/łybyśmy &  conditional & 	1 &	pl	& m/f & & schowa-libyśmy/łybyśmy\\\hline
chowa-libyscie/łybyście & 	conditional	& 2	& pl & m/f, n & & schowa-libyscie/łybyście\\\hline
chowa-liby/łyby	&  conditional	&  3 & 	pl & m/f, n & & schowa-liby/łyby\\\hline
\hline
chowano	&  impersonal & & & & & schowano\\\hline		
\end{tabular}
\caption{\label{tab:chować_basic_forms} Basic forms of the verb \textit{chować} `to hide' (n=81).  The imperfective paradigm features 51 unique forms; the perfective paradigm - 38 unique forms.}
\end{table}
\end{landscape}

Participle forms of the verb \textit{chować} are listed in Table~\ref{tab:chować_participles}, grouped by case, number, and gender.  The available perfective forms are indicated with the prefix \{s-\}. The upper part of the table presents passive adjectival participle forms of \textit{chowany} `being hidden' and \textit{schowany} `having been hidden', the central part shows active adjectival participles of \textit{chowający} `hiding' (no perfective forms exist), and the bottom part provides perfect adverbial participle forms \textit{chowawszy} `when hiding' and \textit{schowawszy} `after having hidden'. Notably, the syncretic forms \textit{(s)chowane} and \textit{chowające} represent the nominative, accusative and vocative cases in both singular and plural feminine, illustrating that a suffix or a sequence of suffixes can realize multiple grammatical functions. 

\begin{table}[H]
\centering
\begin{tabular}{ |c||c|c||c|c| } 
\hline
 case & singular & gender & plural & gender \\\hline
\hline		
\multicolumn{5}{|c|}{passive adjectival participle} \\ \hline
Nom.    & (s)chowa-ny/na/ne	    & m/f/n	    &(s)chowa-ni/ne     & m.p/m.np\\\hline
Gen.    & (s)chowa-nego/nej	    & m,n/f     &(s)chowa-nych      & m.p,m.np\\\hline
Dat.    & (s)chowa-nemu/nej	    & m,n/f     &(s)chowa-nym       & m.p,m.np\\\hline
Acc.    & (s)chowa-nego/ną/ne   & m/f/n	    &(s)chowa-nych/ne   & m.p/m.np\\\hline
Instr.  & (s)chowa-nym/ną	    & m,n/f	    &(s)chowa-nymi      & m.p,m.np\\\hline
Loc.    & (s)chowa-nym/nej	    & m,n/f	    &(s)chowa-nych      & m.p,m.np\\\hline
Voc.    & (s)chowa-ny/na/ne     & m/f/n	    &(s)chowa-ni/ne     & m.p/m.np\\\hline
\multicolumn{5}{|c|}{active adjectival participle} \\ \hline
Nom.    &chowa-jący/jąca/jące   & m/f/n	    &chowa-jący/jące    & m.p/m.np\\\hline
Gen.    &chowa-jącego/jąca	    & m,n/f     &chowa-jacych       & m.p,m.np\\\hline
Dat.    &chowa-jącemu/jącej	    & m,n/f     &chowa-jącym        & m.p,m.np\\\hline
Acc.    &chowa-jącego/jącej/jące& m/f/n	    &chowa-jących/jące  & m.p/m.np\\\hline
Instr.  &chowa-jącym/jącą	    & m,n/f	    &chowa-jącymi       & m.p,m.np\\\hline
Loc.    &chowa-jącym/jącą	    & m,n/f	    &chowa-jących       & m.p,m.np\\\hline
Voc.    &chowa-jący/jąca/jące   & m/f/n	    &chowa-jący/jące    & m.p/m.np\\\hline
\multicolumn{5}{|c|}{perfect adverbial participle} \\ \hline
\multicolumn{5}{|c|}{(s)chowawszy} \\ \hline
\end{tabular}
\caption{\label{tab:chować_participles}Participle forms of the verb \textit{chować} `to hide' (n=83). Out of 27 available passive adjectival participle imperfective forms, 16 (=60\%) are syncretic.  The same proportion holds for the perfective counterparts.  Out of 27 available active adjectival participle imperfective forms, 17 (=63\%) are syncretic.}
\end{table}

Finally, Table~\ref{tab:chować_gerunds} presents the gerund forms of \textit{chować}, grouped by case and number. As in other parts of the paradigm, many of these forms exhibit syncretism.

\begin{table}[H]
\centering
\begin{tabular}{ |c||c|c| } 
\hline
case    & singular            & plural \\\hline
\hline				
Nom.    &    (s)chowa-nie     & (s)chowa-nia\\\hline
Gen.    &	(s)chowa-nia      & (s)chowa-ń\\\hline
Dat.    &	(s)chowa-niu      & (s)chowa-niom\\\hline
Acc.    &	(s)chowa-nie      & (s)chowa-nia\\\hline
Instr.  &   (s)chowa-niem     & (s)chowa-niami\\\hline
Loc.    &	(s)chowa-niu      & (s)chowa-niach\\\hline
Voc.    &	(s)chowa-nie      & (s)chowa-nia\\\hline
\end{tabular}
\caption{\label{tab:chować_gerunds}Gerund forms of the verb \textit{chować} `to hide' (n=28). Out of 14 available imperfective gerund forms, six (=43\%) are syncretic.  The same proportion holds for the perfective counterparts.}
\end{table}

Overall, the full paradigm of \textit{chować} comprises 192 forms (81 verbal, 111 deverbal) including aspectual pairs, but roughly half of the forms are syncretic. This inflectional complexity, together with a wide range of consonantal and vocalic alternations as well as numerous inflectional classes of nouns, poses a substantial challenge for any computational approach. Also note that the three or five gender categories and seven cases that we consider in the analysis are possibly just the tip of the iceberg.  

First, numerous declension groups can be distinguished based on the stem-final segments. \cite{Tokarski-1973-fleksja} lists 13 declensions that include five for the masculine gender, five for the neuter gender, and six for the feminine gender. Second, pluralization can simultaneously entail other types of plural-forming processes such as vowel alternation, and/or deletion of mobile vowels /e o/ (e.g. \textit{miesiąc}-nom.sg. `month' $ \rightarrow $ \textit{miesięcy}-gen.pl., \textit{karzeł}-non.sg. `dwarf' $ \rightarrow $ \textit{karły}-nom.pl.). Additionally, some proper nouns have their own inflectional paradigm. This can lead to a total of 144 unique noun inflection types \citep[as listed in][]{Jagodzinski-2008-gramatyka}. In the nominal paradigm, pluralization is expressed by 21 different inflectional endings across cases and genders (i.e. \{-i, -y, -e, -a, -o, -u, -ach, -ami, -ech, -em, -ich, -im, -imi, -mi, -om, -owie, -ów, -ych, -ym, -ymi, and -Ø\}), compared to 17 possible endings available for the singular forms (i.e. \{-i, -y, -e, -ę, -a, -ą, -o, -u, -ego, -ej, -em, -emu, -im, -owi, -um, -ym, -Ø\}). Moreover, there are lexemes with missing cells or suppletion in the plural forms, in particular in \textit{plura tantum} or uncountable nouns. 

These complexities challenge not only decompositional models but also non-decompositional models such as the DLM.  Morphological irregularities and gaps have been successfully coped with in various models. For example, the DATR model \citep{Evans-Gazdar-1996-DATR} treats irregular morphology systematically, allowing it to capture both default patterns and exceptions, including suppletion and stem alternations. Its formal implementation in Network Morphology \citep{Corbett-fraser-1993-network-morph} and the subsequent extension by \cite{Baerman-brown-corbett-2010-defective-paradigms} have proven effective in dealing with paradigm gaps in the nominal system of Russian. In computational terms, such case/number gaps seem to cluster in semantic space with other defective or low-frequency nouns, and have been shown to be semantically distinct from non-defective lexemes \citep{Chuang-etal-2023-defective-nouns-rus}. For Polish, the question remains whether computational models can accurately predict Polish word forms despite inflectional irregularities and heavy syncretism. 



\section{Computational modelling with the DLM}\label{sec:modelling_dlm}

\subsection{Data}\label{sec:data}

Since one of the goals of the study is to investigate the morphonotactic complexity in Polish and its possible parallels in semantic space, the data comprised a list of words featuring initial morphonotactic clusters.\footnote{Although the present analysis rests upon the orthographic form of a word, given a nearly one-to-one correspondence between spelling and pronunciation in Polish, the results of the study can be generalized to phonotactics.}  The dataset was extracted from The Basic Dictionary of Polish for Foreigners (\textit{Słownik Podstawowy Języka Polskiego dla Cudzoziemców}; \cite{Bartnicka-Sinielnikoff-1999-slownik-podst}, which contains 8000 lemmas.  This resource has been explored in \cite{Zydorowicz-etal-2016-morphonotactics}, who reported 2532 words starting with initial clusters. 456 of these words were morphologically complex and represented different degrees of parsability, as outlined in Section~\ref{sec:transparency}.  The words studied by \cite{Zydorowicz-2019-pl} contained 70 unique morphonotactic cluster types, as listed in Table~\ref{tab:Word-initial morphonotactic clusters used in the study.}.  The table also provides a three-fold classification of the clusters according to their size and sonority profile. 

\begin{table}[H]
\centering
\begin{tabular}{ |c||c|c|c| } 
\hline
size & rise (unmarked) & plateau & fall (marked) \\\hline
\hline
CC & vj vl vm vr vw zj & f\textipa{\textctc} fs fx s\textipa{\textctc} sf ss & \textipa{\textctc}\textipa{\texttctclig} fp f\textipa{\texttctclig} sk sp st zb zd z\textipa{\textdctzlig} zg\\
 & z\textipa{\textObardotlessj} zl zm zn z\textipa{\textltailn} zr zw & s\textipa{\textesh} sx vz v\textipa{\textyogh} zv z\textipa{\textyogh} &\\\hline
CCC & - & fsx & fkw fpr fpw fsk fsp fst skj skl skr sk\textipa{\textesh}\\
  &  & & skw spj spl spr sp\textipa{\textesh} spw stf str stw vzb\\
 & & & vzd vzg zbj zbl zbr zgj zg\textipa{\textltailn} zgr zgw\\\hline
CCCC & - & - & fstr fst\textipa{\textesh} stfj vzdw vzgl\\\hline
\end{tabular}
\caption{\label{tab:Word-initial morphonotactic clusters used in the study.}Word-initial morphonotactic clusters found in the data.}
\end{table}

To ensure that the analyses to follow are based on a representative list of lexical items, for each of the 456 words we extracted their inflectional paradigm members from a balanced 300-million subcorpus of The National Corpus of Polish (\textit{Narodowy Korpus Języka Polskiego, henceforth NKJP300M}; \cite{Przepiorkowski-etal-2012-NKJP}, http://www.nkjp.pl).  The resulting  paradigm list of 21417 cluster-initial words was extracted along with their morphosynatactic description originally developed for a morphosyntactic analyser and generator for Polish, \textit{Morfeusz} \citep{Kieras-wolinski-2017-morfeusz, Wolinski-2003-korpus-znaczniki}, \url{http://morfeusz.sgjp.pl}). 

From this list, we excluded abbreviations (e.g. \textit{wsch} referring to 'the East') as well as word forms that do not start with a cluster. Specifically, we omitted superlative adjectives formed with the \{naj-\} prefix (e.g. \textit{ciemny} `dark' $ \rightarrow $ \textit{najciemniejszy}, \textit{mądry} `wise' $ \rightarrow $ \textit{najmądrzejszy}), and inflected forms of \textit{zgiąć} `to bend' and \textit{ściąć} `to cut' that start with a single consonant (e.g. imperative singular \textit{zegnij} and \textit{zetnij}). Moreover, because the corpus-derived dataset does not include conditional forms created with the suffix \{-by\}, the analyses omit grammatical mood.

The final list of cluster-initial words represents different parts of speech for which the following morphosyntactic properties are specified in the corpus.

\begin{enumerate}[resume]
  \item Morphosyntactic categories for parts of speech\label{list:Grammatical categories for parts of speech}
    \begin{enumerate}
      \item adjective: number, case, gender (n=441)
       \item adverb (n=19)
       \item gerund: number, case, gender, aspect (n=1044)
       \item noun: number, case, gender (n=694)
       \item participle (n=2513)
        \begin{enumerate}
          \item active adjectival: number, case, gender, aspect (n=636)
          \item passive adjectival: number, case, gender, aspect (n=1730)
          \item present adverbial: aspect (n=92)
          \item perfect adverbial: aspect (n=55)
        \end{enumerate}
      \item verb (n=3286)
        \begin{enumerate}
          \item inflected: person, number, aspect, tense  (n=2481) 
          \item impersonal: aspect (n=166)
          \item imperative: person, number, aspect (n=372)
          \item infinitive: aspect (n=267)
        \end{enumerate}
    \end{enumerate}
\end{enumerate}

Apart from the morphosyntactic categories, word-initial clusters were tagged for the following phonological properties: cluster length and sonority profile (see Section~\ref{sec:aspect phonotactics}) as well as a three-degree parsability of prefixes (see Section~\ref{sec:transparency}). 

The next step consisted in constructing a list of unique word forms which can be coupled with word embeddings.  Thus, the list of 21417 inflected forms was curtailed by eliminating homographs.  That is, for each set of syncretic forms, only the most frequent form (according to the NKJP300M corpus) was retained in the dataset, together with the inflectional features of that specific syncretic form.  As a consequence, the original inventory with 21417 forms (including syncretic forms) was reduced to a dataset with 8015 forms. 

For this set of unique words, \textit{fastText} embeddings were available for 7997 word forms, representing 418 lemmas. The average paradigm size for the lemmas is 19 (SD = 11, range=1--42). The largest paradigm is observed for the word \textit{wprowadzać} `to introduce', which has 42 paradigm members. \textit{Word2vec} embeddings were identified for 5936 unique word forms, corresponding to 412 lemmas. The average paradigm size across lemmas is 14 (SD = 9, range=1--36). The largest paradigm is associated with the verb \textit{znajdować} `to find', which comprises 36 distinct word members.

\subsection{Word embeddings}\label{{Extracting word embeddings}}

To represent the meanings of the words in our dataset, we used embeddings. Embeddings are high-dimensional numeric vectors derived from text corpora that represent word meanings. Similarities between embeddings, measured with the Pearson correlation or the cosine similarity measures, dovetail well with perceived similarities as perceived by human raters \citep{Landauer-dumais-1997-plato,Bruni-tran-baroni-2014-multimodal-distr-sem}. Embeddings have been widely used in NLP, AI, and also in psychology and linguistics. They have been successfully used to predict, among others, parts of speech \citep{Westbury-hollis-2018-POS-identification}, grammatical gender \citep{Veeman-2020-gender-embeddings}, word formation \citep{Marelli-baroni-2015-affixation-semantic-space, kisselew-etal-2015-better-understanding-of-distributional-models}, semantic transparency \citep{Marelli-etal-2015-transp, Tian-baayen-2022-transp}, and the semantic structure of nominal pluralization \citep{Chuang-etal-2023-defective-nouns-rus, Nikolaev-etal-2022-nouns-fin, Shafaei-etal-2024-plural-clusters-en}.  

There are many ways in which word meaning can be represented numerically \citep[see, e.g.][]{Shaoul-westbury-2010-HiDEx, Burgess-lund-1995-HAL,Landauer-dumais-1997-plato, Pennington-etal-2014-glove}.  In the present study, we made use of two types of pre-compiled embeddings, one set created with the \textit{Word2vec} algorithm \citep{Mikolov-etal-2013-distributed} and one set created with the \textit{fastText} algorithm \citep{Bojanowski-etal-2017-fasttext}. The former algorithm creates embeddings using prediction methods that have words as indivisible input units. The latter algorithm builds up embeddings for words from the embeddings for multiple sub-strings (with minimally three letters) of words. The \textit{fastText} algorithm is known to perform better than \textit{Word2vec}, especially for languages with complex morphology. By working with sub-strings of words, the problem of data sparsity is considerably reduced.

Previous studies using distributional semantics have mainly made use of \textit{Word2vec}.  A skip-gram model of Polish \textit{Word2vec} was proposed in \cite{Kedzia-etal-2016-vector-pl}, and the application of the original algorithm of \cite{Mikolov-etal-2013-vector-space} to Polish Wikipedia was reported in \cite{Rogalski-szczepaniak-2016-emb-pl}.  Analyses on synonymy, analogy recognition and lexical variation were reported in \cite{Mykowiecka-etal-2017-testing-emb} and \cite{Tatjewski-etal-2017-word-similarity}.  However, it has been observed that  by leveraging the n-grams \textit{fastText} tends to generate embeddings that improve substantially on those of \textit{Word2vec} \citep[see work on Russian and Finnish in][]{Chuang-etal-2023-defective-nouns-rus, Nikolaev-etal-2022-nouns-fin}.

Since \textit{fastText} vectors work with co-occurrences of sub-lexical strings, it might be argued that they absorb aspects of morphology and do not represent ``pure'' lexical meaning. However, it can also be argued that the morphological structure makes systematic contributions to lexical meaning.  Assessment of what exactly is gained by including co-occurrence information about subword n-grams is complicated by the fact that many morphological exponents occur as part of high-frequency simple words. For instance, the English prefix \{be-\} is found in \textit{bear, beer, bend}, and \textit{bet}. As a consequence, its validity as a cue for the semantics of \textit{be-} is severely compromised in token-based evaluations \citep[see e.g.][]{Schreuder-baayen-1994-prefix}. We therefore report results using both \textit{fastText} embeddings (detailed in the main text) and \textit{Word2vec} embeddings (details of which are provided in the Appendix).

\textit{FastText} embeddings were downloaded from \url{https://fastText.cc}. They are based on 387 million tokens representing 1,3 million words from the Polish Wikipedia, and 22 billion word tokens representing 10,2 million words from the Common Crawl corpus \citep[see][for details]{Grave-etal-2018-vectors-157-lg}. \textit{Word2vec} vectors (downloaded from  \url{https://wikipedia2vec.github.io/wikipedia2vec/pretrained/}) are calculated from approximately 1,5 billion tokens in the Polish Wikipedia, books and articles. 

\subsection{The Discriminative Lexicon Model}\label{sec:dlm}

Connectionist models of lexical processing \citep{Rumelhart-mcclelland-1986-distributed-processing, Seidenberg-1987-visual-word-recognition} make use of neural network models to model and explain experimental findings about morphological processing. These models make use of low-level `distributed' representations to represent words' forms and meanings. Effects of phonology and morphology are emergent properties of the distributional statistics of a given language. 

The discriminative lexicon model \citep{Baayen-etal-2019-DLM, Heitmeneier-etal-book} builds on the connectionist tradition, and has focused mainly on using the simplest possible networks to model the relation between form and meaning. This model was originally developed as a computational implementation of Word and Paradigm Morphology \citep{Matthews-1965-wpm, Matthews-1991-morphology, Blevins-2006-wmp, Blevins-2016-wpm}, a theory that relies on analogical generalizations between word forms. Perception, production, language acquisition and development in Word and Paradigm Morphology are driven by analogy, and require no access to the internal structure of words. \cite{Matthews-1991-morphology} explains analogical generalizations on the example of the inflection of the Latin words \textit{dominus} ‘master’ and \textit{servus} ‘slave’. Given the inflectional forms of \textit{dominus}, it is possible ---  by analogy --- to deduce the forms of \textit{servus} using proportional analogy, e.g. of the canonical format A:B = C:D. The exemplary pattern \textit{dominus}--nom.sg. --- \textit{domini}--gen.sg. suggests  \textit{servi} as a genitive singular form of the latter noun.  In distributional semantics, these kind of analogies are straightforward to express at the level of semantics. Representing vectors using bold font, we can write 
$$
\bm{v}_{\text{dominus}, \textsc{pl}} = \bm{v}_{\text{dominus}, \textsc{sg}} + \bar{\bm{v}}_{\text{plural}} 
$$
$$
\bm{v}_{\text{servus}, \textsc{pl}} = \bm{v}_{\text{servus}, \textsc{sg}} + \bar{\bm{v}}_{\text{plural}},
$$
from which it follows that
$$
\bm{v}_{\text{dominus}, \textsc{pl}} - \bm{v}_{\text{dominus}, \textsc{sg}} = 
\bm{v}_{\text{servus}, \textsc{pl}} - \bm{v}_{\text{servus}, \textsc{sg}}.
$$
In other words, shifting from the meaning of the singular to the meaning of the plural in semantic space is identical for the lexemes \textit{dominus} and \textit{servus}: \textit{dominus : domini = servus : servi}.  

The DLM makes use of embeddings to represent words' meanings.  Although one could derive the meaning of a plural from the meaning of its singular by vector addition, it has recently been shown that actual shift vectors for noun plurals can be different for different subsets of words. \cite{Shafaei-etal-2024-plural-clusters-en} observed for English that plural shift vectors vary by semantic class, while \cite{Nikolaev-etal-2022-nouns-fin} and \cite{Chuang-etal-2023-defective-nouns-rus} found for Finnish and Russian that shift vectors vary with case. Furthermore, \cite{Nikolaev-etal-2022-nouns-fin} clarifies that the empirical embeddings of inflected words have their own semantic idiosyncracies that cannot be captured completely by shift vectors or even interactions between shift vectors. Fortunately, the DLM does not depend on how the semantic representations of words are obtained: it can work with simulated embeddings, with embeddings imputed from empirical embeddings, or with actual empirical embeddings.  The embeddings for \textit{dominus} and \textit{servus}, brought together as row vectors of a semantic matrix $\bm{S}$, could look like this:

\begin{equation*}
\renewcommand{\kbldelim}{(}
\renewcommand{\kbrdelim}{)}
\bm{S} =  
\kbordermatrix{
                &  S1 & S2 & S3 & S4 & S5 & S6  & \ldots  \\
\text{dominus}  &  0.1483 &  2.8810 &   1.4648 &  -0.8663 &  -0.4064 &  -2.0027 & \ldots \\
\text{servus}   &  0.8104 &  0.1501 &   1.1981 &  -1.7612 &   0.1230 &  -0.768 & \ldots \\
}.
\end{equation*}

At the level of words' forms, the DLM does not work with stems and exponents.  One kind of form representation that has been found useful works with sub-lexical n-grams, its `form cues'.  The orthographic representation of a word is given by a numeric vector with ones and zeroes, with a '1' representing that an n-gram is present in a given word. For \textit{dominus} and \textit{servus}, form vectors using letter trigrams, brought together in a cue matrix $\bm{C}$, are as follows:

\begin{equation*}
\renewcommand{\kbldelim}{(}
\renewcommand{\kbrdelim}{)}
\bm{C} =   \kbordermatrix{
             &  \#do & dom & omi & min & inu & nus  & us\# & \#se & ser & erv & rvu & vus  & \ldots \\
\text{dominus}  &    1  &  1  &  1  & 1   & 1   & 1    &  1   &  0   &  0  &  0  &  0  & 0  & \ldots  \\
\text{servus}   &    0  &  0  &  0  & 0   & 0   & 0    &  1   &  1   &  1  &  1  &  1  & 1  & \ldots   \\
}.
\end{equation*}

\noindent
The number of columns of the cue matrix is equal to the number of unique trigrams in the dataset.  Note that the exponent for the nominative singular \textit{-us} is represented, mostly in part, by five trigrams: \textit{inu, nus, us\#, rvu,} and \textit{vus}.  Of these, the word final trigram \textit{us\#} is expected to be the most reliable cue to nominative singular.  

Given the $\bm{C}$ and $\bm{S}$ vectors, LDL sets up mappings between the form vectors in $\bm{C}$ and the semantic vectors in $\bm{S}$.  
The comprehension mapping matrix $\bm{F}$ takes form vectors as input, and produces semantic vectors, and the production mapping $\bm{G}$ takes embeddings as input and produces form vectors:

\begin{eqnarray}
    \bm{C}\bm{F} &=& \bm{S} \\ \nonumber
    \bm{S}\bm{G} &=& \bm{C}.
\end{eqnarray}

In this study, we make use of these equations  to estimate the mappings between form and meaning.  This way of learning these mappings is referred to as ``end-state learning'' because the solutions can be obtained in principle by infinite iterative learning with the Widrow-Hoff learning rule \citep{Widrow-hoff-1960-larning-algorithm} to the dataset. It is also possible to use linear mappings that take into account words' frequencies of use, using so-called frequency-informed learning \cite{Heitmeier-etal-2023-frequency-effects-LDL}.  This method is well suited for the modelling of frequency effects.  In the present study, however, we are primarily interested in what can in principle be learned, abstracting away from usage, and hence make use of end-state learning. 

The rows of the $\bm{F}$ matrix specify the contribution that a trigram makes to the (predicted) embedding representing a word's meaning. The row vector for \textit{us\#} is somewhat similar to a nominative singular shift vector. However, the task of realising nominative and singular is shared with other row vectors of $\bm{F}$, such as the one for \textit{nus}.

\begin{equation*}
\renewcommand{\kbldelim}{(}
\renewcommand{\kbrdelim}{)}
\bm{F} =   \kbordermatrix{
             & S1       &  S2       &   S3      &  S4       &  S5       &   S6   & \ldots \\
\text{\#do}  & 10.107	&  7.785	&   8.018   &  -9.277	& -4.782	& -7.148 & \ldots \\
\text{dom}	 & 18.066	&  6.224	&  14.100	&  -7.397	& -5.072	& -5.621 & \ldots \\
\text{nus}	 &  7.416	&  3.604	&   7.625	&  10.396	&  1.664	&  7.678 & \ldots \\
\text{us\#}	 & -3.914	& -1.249	&  -3.554	&  -1.325	&  0.745	&  1.058 & \ldots \\
}.
\end{equation*}

\noindent
The column vectors of the production matrix $\bm{G}$ specify the support that the semantic dimensions provide for the trigrams.

\begin{equation*}
\renewcommand{\kbldelim}{(}
\renewcommand{\kbrdelim}{)}
\bm{G} = \kbordermatrix{
            & \#do	    & dom	    & min       & ser 	    & vus        	& us\#    & \ldots    \\ 
\text{S1}	& -0.002	& -0.002	& -0.001	& -0.001	& -0.001	    & -0.007  & \ldots \\
\text{S2}	&  0.001	& -0.000	& -0.001	& -0.000	& -0.000	    & -0.000  & \ldots \\
\text{S3}	& -0.001	& -0.001	& -0.001	& -0.000	& -0.000	    & -0.000  & \ldots \\
\text{S4}	&  0.001	&  0.001	& -0.001	& -0.001	& -0.001	    & -0.000  & \ldots \\
\text{S5}	& -0.002	& -0.003	& -0.001	&  0.000	&  0.000 	    &  0.000  & \ldots \\
\text{S6}	& -0.001	& -0.002	& -0.001	&  0.000  	&  0.000	    &  0.000  & \ldots \\
\ldots      &           &           &           &           &               &         & \ldots \\
}.
\end{equation*}

The predicted form vectors specify the amount of support that words' trigrams receive from their semantics. They do not specify the order in which the trigrams have to be arranged for production. Furthermore, trigrams that are irrelevant for a given word may nevertheless receive some support from the embeddings. 
In the DLM, the selection and ordering of the trigrams (or triphones) is handled by a separate algorithm which is explained in detail in \cite{Baayen-chuang-blevins-2018-inflectional-morphology-with-linear-mappings} and \cite{Heitmeneier-etal-book}. This algorithm exploits the fact that trigrams provide partial information about order: \textit{\#do} can be followed immediately by \textit{dom} (due to the shared bigram \textit{do}), but not by \textit{min}. The algorithm produces for a given word the most likely pronunciation and its closest competing pronunciations. \citet{Heitmeneier-etal-book} provide an overview of model performance for a range of languages with widely different morphological systems. The ordering algorithm requires the user to specify a threshold parameter \textipa{\texttheta}. The algorithm takes as input the predicted triphone vector, but only works with those trigrams that have a value greater than the threshold parameter.  For the models reported below, the threshold was set to 0.01.

The modelling study to follow applies the DLM to Polish, and investigates model performance with respect to a comprehensive set of morphosyntactic characteristics of Polish words along with their phonological, morphological and morphonological properties. 

The modelling of comprehension and production was performed using the \textit{JudiLing} package \citep{Heitmeneier-etal-book} designed for \textit{Julia} (v. 1.9.0; \cite{Bezanson-etal-2017-julia}). We constructed a cue matrix using trigrams, resulting in a 7997 $\times$ 1922 matrix $\bm{C}$ with words on rows and letter trigrams on the columns. For these words, the corresponding \textit{fastText} vectors were brought together in a 7997 $\times$ 300 matrix $\bm{S}$. The $\bm{S}$ matrix for \textit{Word2vec} embeddings had the same dimension.

\subsection{Results}

We ran a series of analyses to calculate form-to-meaning (comprehension) and meaning-to-form (production) mappings, using different datasets.  The first dataset comprised all words. Models built for this dataset are useful for evaluating how well the model can learn with maximal input.  The second dataset comprised a subset of 7597 words that we used for training. The remaining 400 words constituted held-out data that we used for assessing the extent to which a model generalizes to unseen data.  The held-out dataset was chosen in such a way that the stems and inflectional features of the words were attested in the training set.  In other words, the held-out data did not contain words with previously unseen stems, or previously unseen cases or numbers.  

Model accuracy for comprehension was evaluated by comparing predicted semantic vectors with the gold standard \textit{fastText} vectors. For a given word, recognition was judged to be accurate if the predicted semantic vector was closer to its targeted gold standard counterpart than to any other gold standard \textit{fastText} vector. For production accuracy, we compared the word form predicted by the model with the targeted word form.  Production was evaluated as accurate when the predicted and targeted word forms were identical, and as inaccurate otherwise. 

For all the datasets, we observed high accuracies for the form-to-meaning mapping (cf. Table~\ref{tab:LDL_accuracy}). That is, comprehension was successful in more than \SI{97.2}{\percent} for the full dataset, \SI{97.5}{\percent} for the training dataset, and \SI{98.3}{\percent} for the held-out dataset.  These results clarify that the comprehension model is not overfitting on the training data. For production, accuracies were similar for the full dataset (97.3\%) and the training data (97.6\%), but lower for the held-out data (77.8\%).\footnote{The production algorithm makes use of a feedback loop from form to meaning. For this feedback loop, we made use of a mapping matrix $\bm F$ that was trained on both the training and test data. This set-up implies that we are testing production accuracies for words that have not been spoken before, but that have been heard before. Accuracy for the scenario for which held-out words also have not been heard before is 76.0\%.}. As human production typically lags behind comprehension, the drop in accuracy on held-out data fits well with human learning. For a comparison with \textit{Word2vec} embeddings, see Table~\ref{tab:LDL_accuracy_2vec} in the Appendix. 

\begin{table}[H]
\centering
\begin{tabular}{ |c||c|c|c|} 
 \hline
  & full data & training data & test data\\\hline
   \hline
comprehension & 97.2 & 97.5 & 98.3 \\\hline
production & 97.3 &  97.6 & 77.8 \\\hline
\end{tabular}
\caption{\label{tab:LDL_accuracy} Accuracy (in percentages) for comprehension and production for the full dataset, a training dataset and a held-out test data set.}
\end{table}

We inspected the 104 errors made by the production model for the held-out data. The analysis shows that within this set, 82\% (n=85) of forms constitute fully-fledged members of the Polish vocabulary. In the majority of cases, the new and target forms represent the same lexeme, but differ in terms of one or two grammatical features.  An overview of the types of production errors conditional on the stem being produced correctly is listed in Table~\ref{tab:production_errors}. 

\begin{table}[H]
\centering
\begin{tabular}{ |c|c||c|c|} 
 \hline
 change in one feature & no of forms & change in two features & no. of forms\\\hline
   \hline
aspect & 11 & aspect+tense & 3 \\\hline
case & 8 & number+person & 2 \\\hline
number & 7 & number+case & 2 \\\hline
tense & 6 & tense+person & 1 \\\hline
gender & 3 & case+gender & 1 \\\hline
person & 2 &  &  \\\hline
\end{tabular}
\caption{Counts of errors cross-classified by inflectional features for words that are produced by the model with the correct stem.}
\label{tab:production_errors}
\end{table}

Among the predicted forms that are part of the Polish vocabulary, there are 15 cases where the wrong lexeme has been selected but where the inflections are otherwise correct. \textit{zdziwi} `(s)he will surprise' is the best form predicted for the target \textit{spyta} `(s)he will ask', and \textit{zmniejszyły} `they diminished' is the closest form for the target \textit{zwiększyły} `they increased'.  The other cases are \textit{wstępie} `introduction' > \textit{wspomnieniu} `memory', \textit{wschodziło} `rose (the sun)' > \textit{wstawało} `it got up', \textit{znienawidzić} `to hate' > \textit{wzruszyć} `to move (emotionally)', \textit{stworzę} `I  will create' > \textit{zrobię} `I will do', \textit{scharakteryzuję} `I will characterise > \textit{wskazuję} `I will point something out', \textit{scharakteryzowanego} `characterised' > \textit{sformulowanego} `formulated', \textit{względniejsza} `more relative' > \textit{wprawiana} `set (into motion)', \textit{skomplikowanego} `complicated' > \textit{sprawnego} `efficient', \textit{względami} `considerations' > \textit{zmartwieniami} `worries'. Several of these changes involve semantically somewhat related stems. Finally, \textit{zwiększyły} `they increased' > \textit{zmniejszyły} `they decreased' illustrates a semantic `speech error' where antonyms are exchanged. A possible reason is that word embeddings tend to be challenged when it comes to teasing antonyms apart, as antonyms appear in very similar contexts.

Interestingly, only in 19 cases, LDL produced words that are non-existent in Polish, although they are phonotactically legal and exhibit phonetic and morphological similarity to existing  forms. These production errors involve, among others, the deletion or substitution of a consonant.  For \textit{zmarszczyć} `to frown' and \textit{zniszczcie} `destroy (imp.pl.)' , the model simplifies the medial cluster and produces \textit{zmarszyć} and \textit{zniszcie}. These examples illustrate that given the embeddings of the target word, phonologically simplified alternative pronunciations are just as good as expressing their meanings, and hence suggest potential language change, or alternative reduced forms suitable for spontaneous informal speech.  In some cases, a full syllable is deleted, as in \textit{zmieszajcie} `mix (imp.)' > \textit{zmieszcie}).  Occasionally, the model adds a meaningless syllable, as for \textit{zjecie} `you will eat' > \textit{zjedzicie}, perhaps garden pathed by \textit{zjedz} (imperative sg.) or \textit{zjedzcie} (imperative pl.). Likewise, \textit{zda} `(s)he will pass' > \textit{zdadza}, perhaps by analogy to \textit{zdadzą} `they will pass'.

\section{LDA and t-SNE analyses of semantic space}\label{sec:analyses}

The question that arises at this point is how it is possible that simple linear mappings perform so well, not only for training data but also for held-out test-data. Our hypothesis is that substantial similarities between the form space and the semantic space are at issue. The more random the relation between form and meaning is, the more a model will have to depend on memorisation, and the less it will be able to generalize. Since the DLM is able to generalize, it is likely that variation in form is reflected in variation in semantics.  Of course, it is well-known that morphology breaks the arbitrary relation of the linguistic sign, introducing systematic form-meaning correspondences.  We therefore expect that it should be possible to predict words' morphosyntactic features (aspect, tense, number, person, gender, case) from their embeddings.  However, we will show that the phonotactic properties of word onsets (cluster size, markedness and the phonological characteristics of the prefix) and the morphonological properties of such onsets (the prefix's separability from the stem) can also be predicted with surprising accuracy from their embeddings.  In what follows, we use linear discriminant analysis and t-distributed stochastic neighbour embedding to document the considerable isomorphies between the form space and the semantic space.

\subsection{Hypotheses}

Consider three more specific hypotheses on why it is possible for the DLM to predict comprehension and production so well. 

\begin{description}
    \item[Hypothesis 1] \textit{There is considerable isomorphy between the semantic space of Polish and the phonological and morphonological properties of Polish word forms.} 
    \item[Hypothesis 2] \textit{Phonological properties of clusters such as cluster length, cluster markedness and the phonological characteristics of a prefix are mirrored in semantic space.}
    \item[Hypothesis 3] \textit{Degrees of prefix-stem parsability are reflected in semantic space.}
    \item[Hypothesis 4] \textit{The grammatical functions realized by exponents for tense, aspect, person, number, gender, and case are more strongly reflected in semantic space than phonotactics and morphonotactics.}
\end{description}

Hypothesis 1 (H1) is general, and argues that there must be considerable similarity in the structure of the form space and the structure of the semantic space for the linear mappings of the DLM to be able to achieve high accuracies. Hypothesis 2 (H2) narrows this hypothesis down to phonotactics (see Sections~\ref{sec:(mor)phonotactics} and~\ref{sec:aspect phonotactics}), hypothesis 3 (H3) to morphonotactics (see Section~\ref{sec:transparency}), and hypothesis 4 (H4) proposes that morphological form should have the strongest isomorphies with meaning (see Section~\ref{sec:morphology}). Drawing on De Saussure’s view on the linguistic signs \citep{DeSaussure-1966-course}, phonemes and their distinctive features must be more arbitrary than morphemes, as morphemes carry lexical and grammatical meaning. Morphology implements form-meaning correspondences that are more systematic and pervasive than correspondences provided by phonesthemes and sound symbolism for simple stems.  In consequence, the relationship between morphemes and meaning should be more predictable than between the phonological properties of consonant clusters and meaning.  Thus, for the hypotheses adumbrated above, we expect clear evidence for semantic clusters by grammatical functions (H4), and a less clear reflection of phonotactic patterns in patterns of meaning (H2). 

Perpendicular to these hypothesis, we also examine the question to what extent \textit{Word2vec} embeddings are informative about the above mentioned form properties. Given that \textit{fastText} embeddings work with sublexical n-grams, it is expected that they will outperform \textit{Word2vec} embeddings. However, if the \textit{Word2vec} embeddings also prove informative, this would imply that the distributional patterns of whole-word forms in text likewise reflect underlying form-based constraints.

\subsection{Method}

We made use of two statistical methods to explore the isomorphies between the form space and the semantic space of Polish.  The first method is Linear Discriminant Analysis (henceforth LDA, \cite{Fischer-1936-LDA}; \cite{Rao-1948-LDA}), which is a supervised classification algorithm.  Given a matrix with row vectors as words' embeddings, and given a vector specifying for each word the class it belongs to, LDA predicts, to the extent this is possible, class membership from the embeddings.  We made use of the LDA implementation in the \textit{MASS} package for R, which also provides an efficient procedure for leave-one-out cross-validation.  We employed this option as a safeguard against overfitting.  It should be kept in mind that LDA is an excellent classifier, but one that is limited by its linearity assumptions.  More accurate results could be obtained by using, for instance, support vector machines \citep{Vapnik-chervonenkis-1964-note}.  In the present study, the LDA results therefore provide a lower bound on accuracy.

We complemented the LDA analyses with visualization using t-SNE \citep{Maaten-2008-visualizing-tSNE}.  T-SNE is an unsupervised technique for dimensionality reduction that makes it possible to visualize high-dimensional data (here, the embeddings of complex words in semantic space) in a 2-dimensional plane.  This visualization method is optimized for finding groupings of data points (or clusters), if groupings exist.  Words with similar semantic vectors are likely to be clustered together, while words with dissimilar vectors are likely to be wide apart.  In a t-SNE plot, colour coding by class can be used to clarify whether words from the same class indeed are grouped together. 

The following analyses make use of \textit{fastText} embeddings, as these proved to be superior to \textit{Word2vec} embeddings also for our dataset. The results obtained with \textit{Word2vec} embeddings are reported in detail in the Appendix.  In section~\ref{sec:analysis_summary}, we compare the results obtained with the two types of embeddings.

In what follows, we report the extent to which the semantic space reflects (and is predictive for), successively, parts of speech, tense, person, case, gender, number, aspect, prefix type, morphotactic parsability of a cluster, cluster markedness, and cluster length.  
This order of presentation reflects the degree to which clusters are separable in t-SNE maps.  That is, t-SNEs for the morphosyntactic categories presented first give rise to tighter and well-defined clusters, while t-SNEs for the (mor)phonological categories presented last display greater overlap. 

\subsection{Results}

\subsubsection{Parts of speech}\label{sec:analysis_POS}

We carried out two analyses: one using six parts of speech, and a more fine-grained analysis with 14 parts of speech.  The first set of parts of speech includes nouns, adjectives, adverbs, verbs, participles and gerunds.  The second set is based on a more detailed classification from the NKJP300M corpus, which subdivides verbs into infinitive, impersonal, imperative, past, present, future, and participles into active adjectival, passive adjectival, present adverbial and perfective adverbial. 

\begin{figure}[htbp]
\centering
\includegraphics[width=0.5\textwidth, height=6cm]{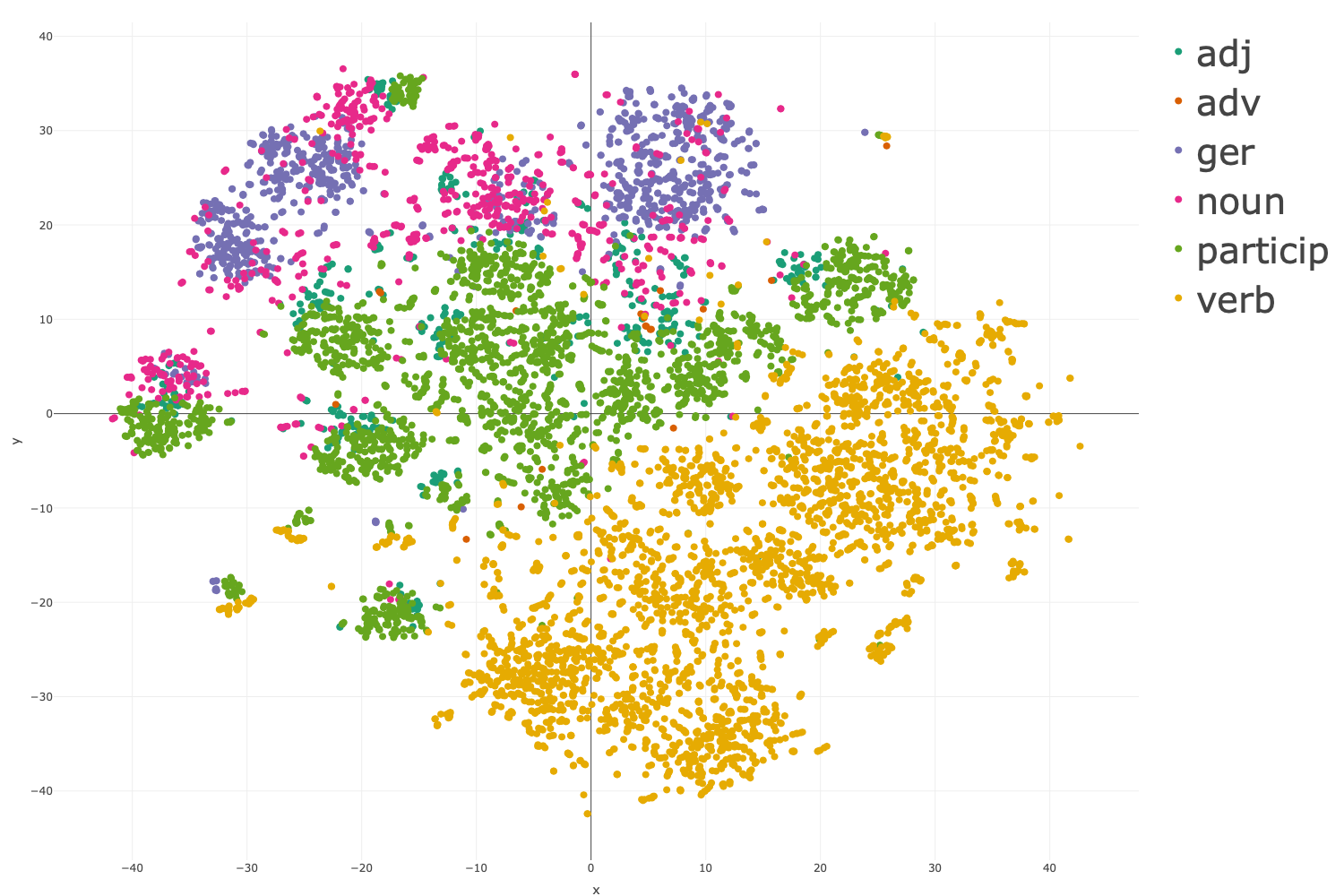}

\ \\

\hspace*{4.8em}\includegraphics[width=0.6\textwidth, height=6cm]{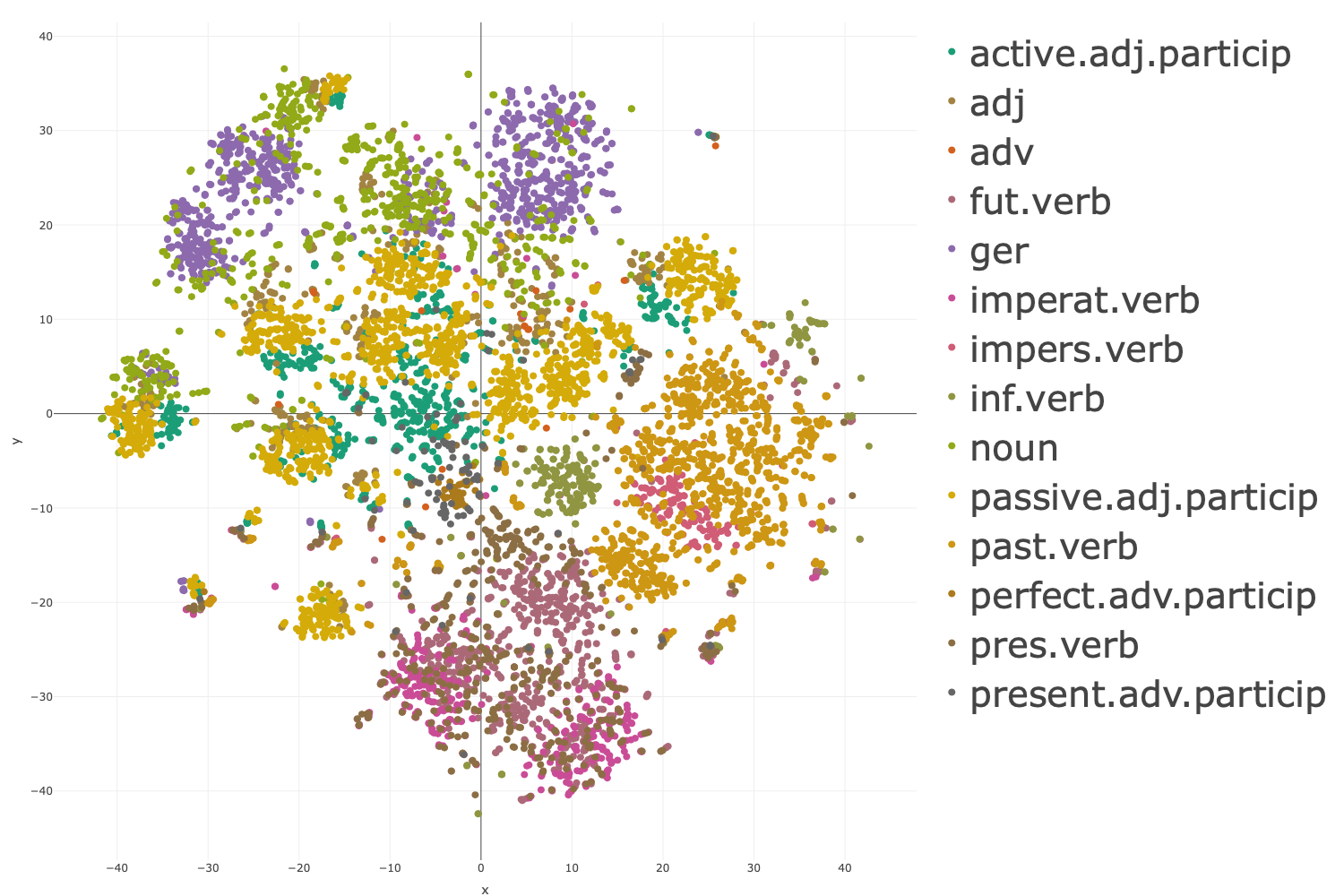}
\caption{Scatterplots of words in a 2-dimensional t-SNE space, colour-coded for six parts of speech (top) and 14 parts of speech (bottom). Further details can be traced in the corresponding interactive maps for the [\href{run:POS_tsne.html}{6 parts of speech}] and [\href{run:POS_detailed_tsne.html}{14 parts of speech}].}
\label{fig:POS}
\end{figure}

In Figure~\ref{fig:POS}, the t-SNE map in the upper panel shows relatively distinct clusters for the six high-level parts of speech. The horizontal axis separates nominal categories from verbs. Nouns (pink) and gerunds (purple) are represented more to the upper left, and verbs (yellow) more to the lower right. Some parts of speech show overlap, notably gerunds and nouns. The lower panel presents the same map, now colour-coded for 14 parts of speech, revealing more distinct clusters for verbs and participles. While some of the fine-grained parts of speech are well differentiated and form coherent clusters of their own, e.g. infinitive verbs (olive green), there is still some overlap, notably within the verb categories: impersonal verbs (light pink) overlap with past verbs (gold yellow), and imperative verbs (pink) overlap with future/present verbs (lilac and brown). Past tense verbs are positioned close to future/present tense verbs. Similar patterns are observed in the class of participles: perfective/present adverbial participles overlap, while passive/active adverbial participle clusters occur in close proximity, complementing each other. Additionally, adjectives tend to cluster around the latter groups.

It should be noted that, for some parts of speech, individual clusters represent further subdivisions based on other categories, which will be detailed in the following sections. For example, among the two clusters for imperative verbs in the central-bottom quadrant of the map, plural forms are distributed to the left of the vertical axis, while singular forms appear to the right. A similar pattern holds holds for future/present verbs, although the clusters are less distinct. In other cases, individual groupings are determined by gender or case. For instance, in the class of past verbs, feminine forms are located above the horizontal axis, whereas instrumental and genitive gerunds occupy opposite ends of the horizontal axis. These word properties are discussed in detail in the sections that follow. 

Given the excellent clustering in the t-SNE plots,  the high overall accuracy of the corresponding LDA analysis for six parts of speech (96\%, majority baseline = 41\%, $p < 0.0001$, proportions test) is unsurprising.  The confusion matrix for the six higher-level parts of speech, shown as Table~\ref{tab:POS}, dovetails well with the corresponding t-SNE map.  Values for the predicted parts of speech are displayed in columns, and values for the actual parts of speech in rows. In brackets, we provide proportions of correctly classified words for each category, i.e. calculated over the total number of words in a row.   

As expected, the highest values are found on the diagonal. The highest values generally correspond to clear groupings by part-of-speech in the t-SNE map, particularly for verbs and participles, for which the classification accuracy exceeds 98\%. Among the misclassified cases, adjectives are confusable with participles (n=34), gerunds with nouns (n=61), nouns with both adjectives (n=43) and gerunds (n=55). These patterns are mirrored in the t-SNE plot, where gerunds, nouns, adjectives, and some participles are not clearly distinguishable. The high proportion of correct classifications for verbs and participles is also evident in the t-SNE map in Figure~\ref{fig:POS}. When misclassifications occur, verbs tend to cluster with nouns or participles, while participles are often grouped with adjectives. The first pattern is visible in both t-SNE maps, where verbs are located within or near noun and participle clusters. Interestingly, the LDA classifier is 100\% accurate for adverbs and gerunds. In the t-SNE map, the small number of adverbs (n=19) is barely visible, although they are relatively well --- but not perfectly --- separated. However, in the original high-dimensional space, these two word classes are well-separated. 

The corresponding LDA analysis using \textit{Word2vec} embeddings is summarized in the Appendix in Table~\ref{tab:POS_2vec}. While the accuracies are lower than for \textit{fastText} embeddings, they remain well above the majority baselines.

\begin{table}[H]
\centering
\begin{tabular}{ |c||c|c|c|c|c|c| } 
 \hline
            &adjective      &adverb &gerund &noun  &participle &verb   \\\hline
  \hline
adjective   &381 (86\%)     &0          &1          &19         &34         &6      \\\hline
adverb      &2              &14 (74\%)  &0          &1          &2          &0      \\\hline
gerund      &3              &0          &977 (94\%) &61         &3          &0      \\\hline
noun        &43             &2          &55         &574 (83\%) &11         &9      \\\hline
participle  &18             &1          &1          &0          &2487 (99\%) &6      \\\hline
verb        &10             &6          &9          &18         &14         &3229 (98\%)   \\\hline
\end{tabular}
\caption{\label{tab:POS}A confusion matrix for six parts of speech; model accuracy = \SI{96}{\percent}.}
\end{table}

The overall accuracy of the model for 14 parts of speech achieves 95\% (majority baseline = 22\%, $p < 0.0001$), as shown in the confusion matrix in Table~\ref{tab:POS_detailed}.  For some parts of speech, prediction accuracy exceeds 99\%, e.g. present adverbial participles, present verbs and active adjectival participles. In the corresponding t-SNE plane shown in Figure~\ref{fig:POS}, many subclasses form distinct clusters, although there is also considerable overlap. For example, some imperative verbs are scattered in the top portion of the plot, overlapping with adjectives, gerunds, and nouns. The confusability of imperatives with other categories is reflected in Table~\ref{tab:POS_detailed}, where the accuracy of the LDA for this part of speech is among the lowest. Similarly, nouns are often misclassified as gerunds (n=54). This misclassification is also visible in the t-SNE plot, where two clusters of nouns found at the top of the plot overlap or are adjacent to gerund forms.

The results for \textit{Word2vec} vectors are summarized in Table~\ref{tab:POS_detailed_2vec} in the Appendix . As before, accuracies tend to be lower than those for \textit{fastText}, but by a wide margin exceed majority baselines. 

\begin{landscape}
\begin{table}[H]
\centering
\begin{tabular}{ |c||c|c|c|c|c|c|c|c|c|c|c|c|c|c| } 
 \hline
                    & active.adj.P & adj  &adv &fut.V  &ger  &imperatV  &impersV &infV &noun   &passive.adj.P &past.V &perf.adv.P &presV &pres.adv.P \\\hline
  \hline
active.adj.particip & 633  (99,5\%)&0&1&0&0&0&0    &0   &0&2&0&0   &0&0\\\hline
adj                 & 4&387  (88\%)&0  &0&1&0&0    &0   &22&22&3&2   &0&0 \\\hline
adv                 &0&3&13  (68\%)&0   &0&0&0    &0   &1&1&0     &0   &0&1\\\hline
fut.verb            &4     &1&1&735  (97\%)&5&0&0    &1   &2&1&1&1&5&1\\\hline
ger                 &0&2&0  &0&980  (94\%)&0&0    &0   &61&1&0&0&0     &0 \\\hline
imperat.verb        &4&2&0  &8   &5&331  (89\%)&3&4&5&3&3&2&1&1\\\hline
impers.verb         &0&0   &0  &0&0&0&162  (98\%)&0   &1&2&1&0   &0     &0 \\\hline
inf.verb            &2&2&0  &0&0&0   &0    &263  (99\%)&0    &0&0&0   &0     &0 \\\hline
noun                &5&36  &2  &1   &54&2&1&3&579  (83\%)&4&0&1   &2&4\\\hline
passive.adj.particip &5&25&0  &1&2&0&0&0   &0&1697 (98\%)&0&0&0     &0\\\hline
past.verb           &0&3&1&0&0&0&1&0   &0&2&1212 (99\%)&2&0&1\\\hline
perf.adv.particip   &1&0   &0  &0&0&0&0    &0   &1&0&0&53  (96\%)&0     &0\\\hline
pres.verb           &0&0&0  &0&0&0&0&0   &0&0&1&0   &500  (99,8\%)&0\\\hline
pres.adv.particip   &0&0   &0  &0&0&0&0    &0   &0    &0&0&0&0&92  (100\%)\\\hline
\end{tabular}
\caption{\label{tab:POS_detailed}A confusion matrix for 14 parts of speech; model accuracy = \SI{95}{\percent}.}
\end{table}
\end{landscape}

\subsubsection{Tense}\label{sec:analysis_tense}

\begin{figure}[H]
\centering
\includegraphics[width=0.6\textwidth]{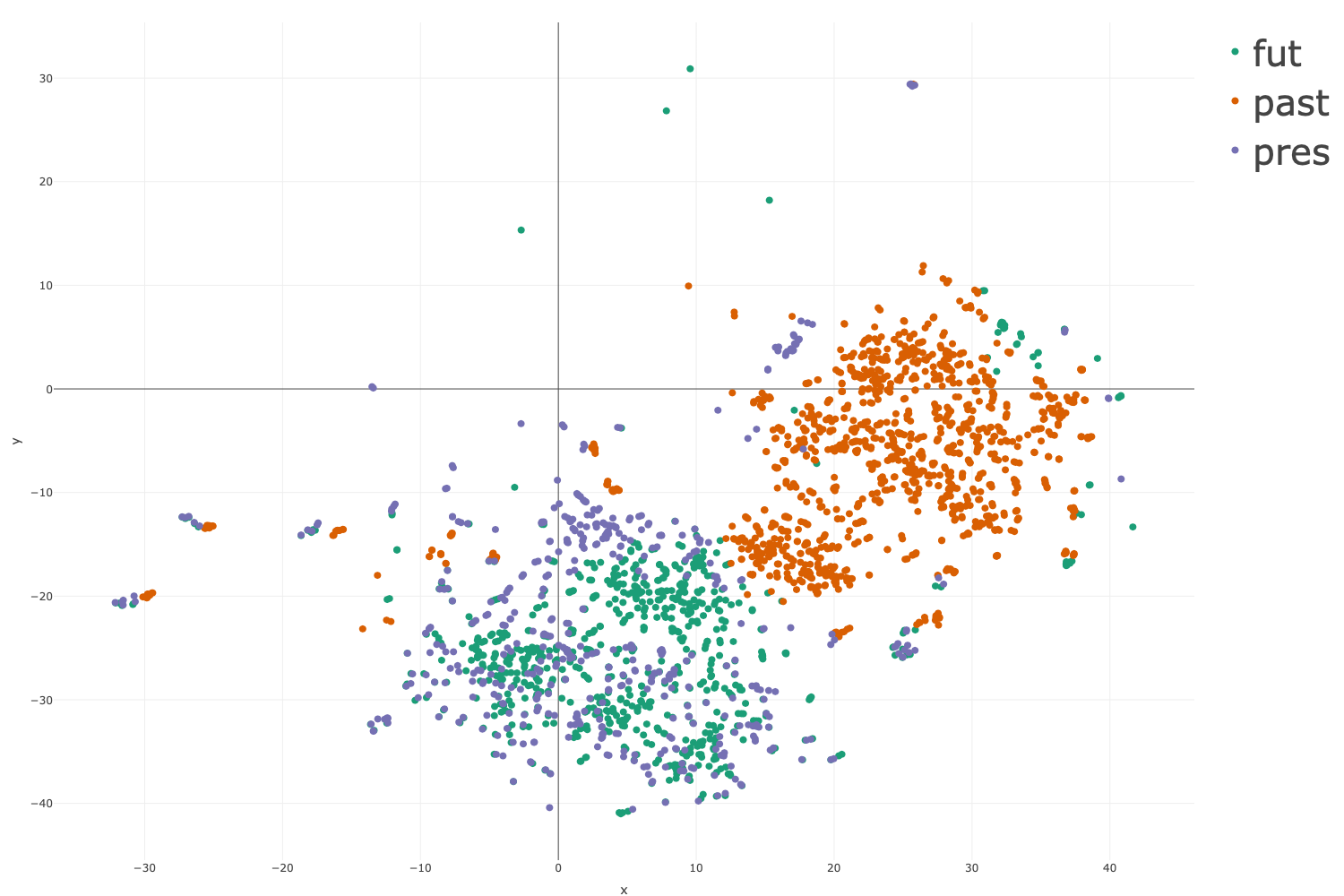}
\caption{\label{fig:tense}Scatterplot of words in a 2-dimensional t-SNE space, colour-coded for three tenses. An interactive map is found [\href{run:Tense_tsne.html}{here}].}
\end{figure}

Clear groupings are also observed for the verbal category of tense.  The t-SNE plane in Figure~\ref{fig:tense} shows a distinct separation between past (orange) and non-past tenses.  Present (purple) and future (green) tenses are more intermixed and dispersed across the map.  The overlap between them reflects their semantic proximity.  The horizontal axis further teases apart past verbs in the 3rd person singular feminine forms from all other forms.

Similarly, LDA achieves very high prediction accuracy for the three tenses, reaching 99\% (majority baseline = 49\%, $p < 0.0001$).  Each tense category is predicted above the 99\% accuracy level, particularly for the present and past tenses.  The results are summarized in the confusion matrix in Table~\ref{tab:tense} (for \textit{Word2vec} results, see Table~\ref{tab:tense_2vec}).

Each tense category is predicted with more than 99\% accuracy, particularly for the present and past tenses. The results are summarized in the confusion matrix in Table~\ref{tab:tense} (for \textit{Word2vec} results, see Table~\ref{tab:tense_2vec}).

\begin{table}[H]
\centering
\begin{tabular}{ |c||c|c|c| } 
 \hline
            &future     &past           &present        \\\hline
   \hline
future      &750 (99\%) &4              &4              \\\hline
past        &0          &1221(99,9\%)   &1              \\\hline
present     &3          &2              &496 (99,9\%)   \\\hline
\end{tabular}
\caption{\label{tab:tense}A confusion matrix for tense; model accuracy = \SI{99}{\percent}.}
\end{table}

\subsubsection{Person}\label{sec:analysis_person}

\begin{figure}[H]
\centering
\includegraphics[width=0.6\textwidth]{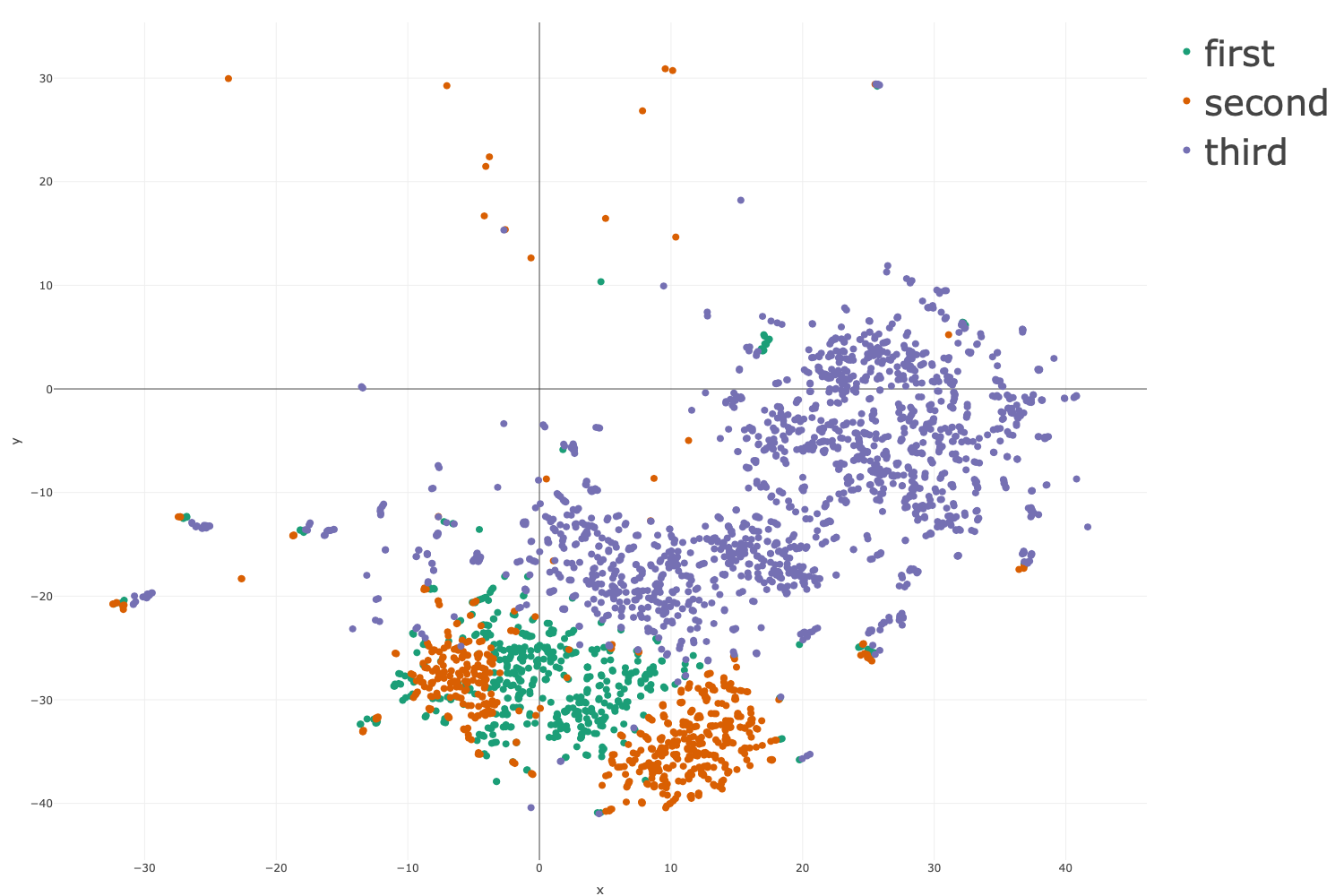}
\caption{\label{fig:person}Scatterplot of words in a 2-dimensional t-SNE space, colour-coded for three persons. An interactive map is found [\href{run:Person_tsne.html}{here}].}
\end{figure}

The grammatical category of person, marked on verbs, forms well-defined clusters in the t-SNE map.  As shown in Figure~\ref{fig:person}, there is a clear separation between the third person (purple) and the first/second persons. First (green) and second (orange) persons are also well distinguished: the (joint) cluster to the left of the vertical axis represents plurals, while the more separated clusters to the right represent singulars. Further subdivisions of the third person are present based on number, gender, and tense, although these groupings are less distinct. For example, the upper part of the map separates singular feminine past verbs from the remaining forms, which are located below the horizontal axis.

These clear splits are mirrored in the LDA analysis. The classifier predicts the respective groupings with an accuracy of 99\% (majority baseline = 60\%, $p < 0.0001$). High accuracy is achieved for all the categories, with performance for the first and third persons being particularly robust. The confusion matrix  summarizing the results is given in Table~\ref{tab:person} (for \textit{Word2vec} results, see Table~\ref{tab:person_2vec}). 

\begin{table}[H]
\centering
\begin{tabular}{ |c||c|c|c| } 
 \hline
            &first      &second         &third          \\\hline
   \hline
first       &494 (99\%) &2              &5              \\\hline
second      &15         &611 (97\%)     &4              \\\hline
third       &1          &8              &1713 (99\%)    \\\hline
\end{tabular}
\caption{\label{tab:person}A confusion matrix for person; model accuracy = \SI{99}{\percent}.}
\end{table}

\subsubsection{Case}\label{sec:analysis_case}

\begin{figure}[H]
\centering
\includegraphics[width=0.6\textwidth]{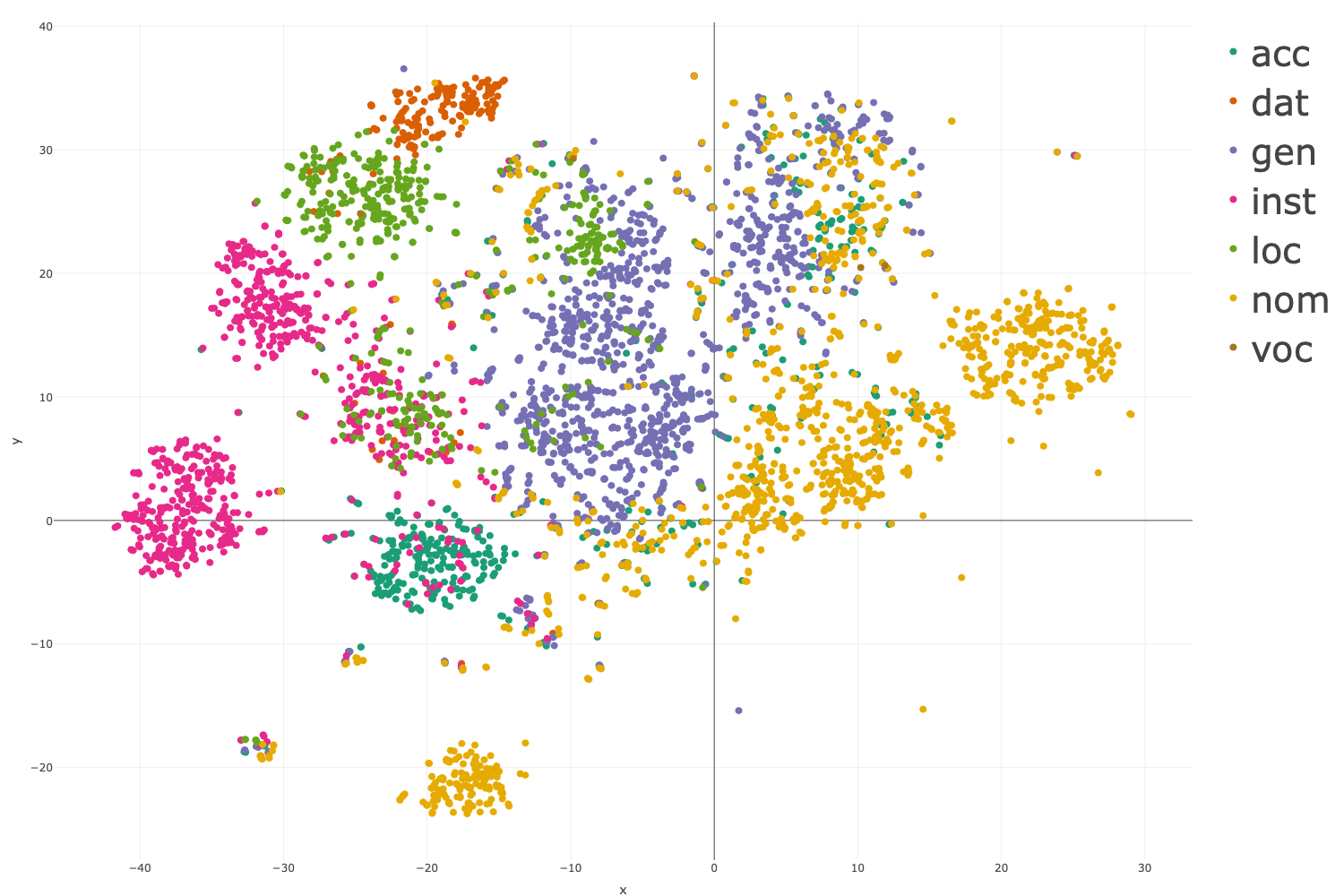}
\caption{\label{fig:case}Scatterplot of words in a 2-dimensional t-SNE space, colour-coded for seven cases. An interactive map is found [\href{run:Case_tsne.html}{here}].}
\end{figure}

Case in Polish is expressed on nouns, adjectives, gerunds and selected participles, which occupy three quadrants of the map.  Case constitutes another grammatical category for which well-defined groupings can be identified in the t-SNE representation in Figure~\ref{fig:case}, although some overlap arises due to syncretism (see Section~\ref{sec:morphology}). As shown in the figure, datives (orange) are well separated from accusatives (dark green) and locatives (light green) along the vertical axis. However, many isolated accusatives overlap with genitives (purple) and nominatives (yellow). Most nominatives are separated diagonally from the remaining cases, although they can be confusable with accusative forms --- a pattern that reflects the nominative-accusative syncretism present in nouns (i.e. masculine inanimate \textit{dom} 'house', neuter \textit{dziecko} 'child'), participles (i.e. active adjectival \textit{piszący} 'writing' for masculine inanimate singular referents, passive adjectival \textit{zrobione} 'made' for neuter singular referents) and gerunds (i.e. \textit{czytanie} 'reading'). Instrumentals (pink) form four distinct clusters, with two being well-separated on the far-left, and two showing considerable overlap with locatives and accusatives. This distribution likewise reflects case syncretism: instrumental and locative forms are identical for masculine and neuter singular adjectives and participles, and instrumental and accusative forms are identical for feminine singular adjective and participles (see Table~\ref{tab:chować_participles}). Beyond the joint locative–instrumental cluster, the locative case is spread out across two additional clusters. One isolated grouping in the upper left contains singular nouns. Another grouping, which overlaps with the genitive case, represents syncretic forms in plural participles and adjectives across all genders.

This clustering in the t-SNE plane is supported by the confusion matrix of a linear discriminant analysis, presented as Table~\ref{tab:case} (for \textit{Word2vec} results, see Table~\ref{tab:case_2vec}). The overall accuracy of the model reaches \SI{89}{\percent} (majority baseline = 28\%, $p < 0.0001$). The most confusable case is the accusative: its confusability with the nominative case reflects the pervasive syncretism between these forms in Polish. Moreover, a somewhat lower prediction accuracy for the locative arises from its frequent confusion with the instrumental.  We do not interpret the result for the vocative, given the scarcity of vocative forms in the dataset (n=4). 

\begin{landscape}
\begin{table}[H]
\centering
\begin{tabular}{ |c||c|c|c|c|c|c|c| } 
 \hline
            &accusative &dative    &genitive   &instrumental &locative &nominative  &vocative   \\\hline
  \hline
accusative  &321 (64\%) &1          &21         &1          &1          &158        &1          \\\hline
dative      &0          &170 (81\%) &0          &22         &18         &0          &0          \\\hline
genitive    &5          &0          &1244 (98\%) &2         &2          &6          &5          \\\hline
instrumental&32         &0          &1          &752 (94\%) &12         &2          &0          \\\hline
locative    &1          &2          &27         &65         &392 (80\%) &1          &1          \\\hline
nominative  &69         &4          &12         &4          &2          &1176 (93\%) &2         \\\hline
vocative    &0          &1          &0          &0          &0          &1          &2 (50\%)   \\\hline
\end{tabular}
\caption{\label{tab:case}A confusion matrix for case; model accuracy = \SI{89}{\percent}.}
\end{table}
\end{landscape}

\subsubsection{Gender}\label{sec:analysis_gender}

\begin{figure}[htbp]
\centering
\includegraphics[width=0.5\textwidth,height=6cm]{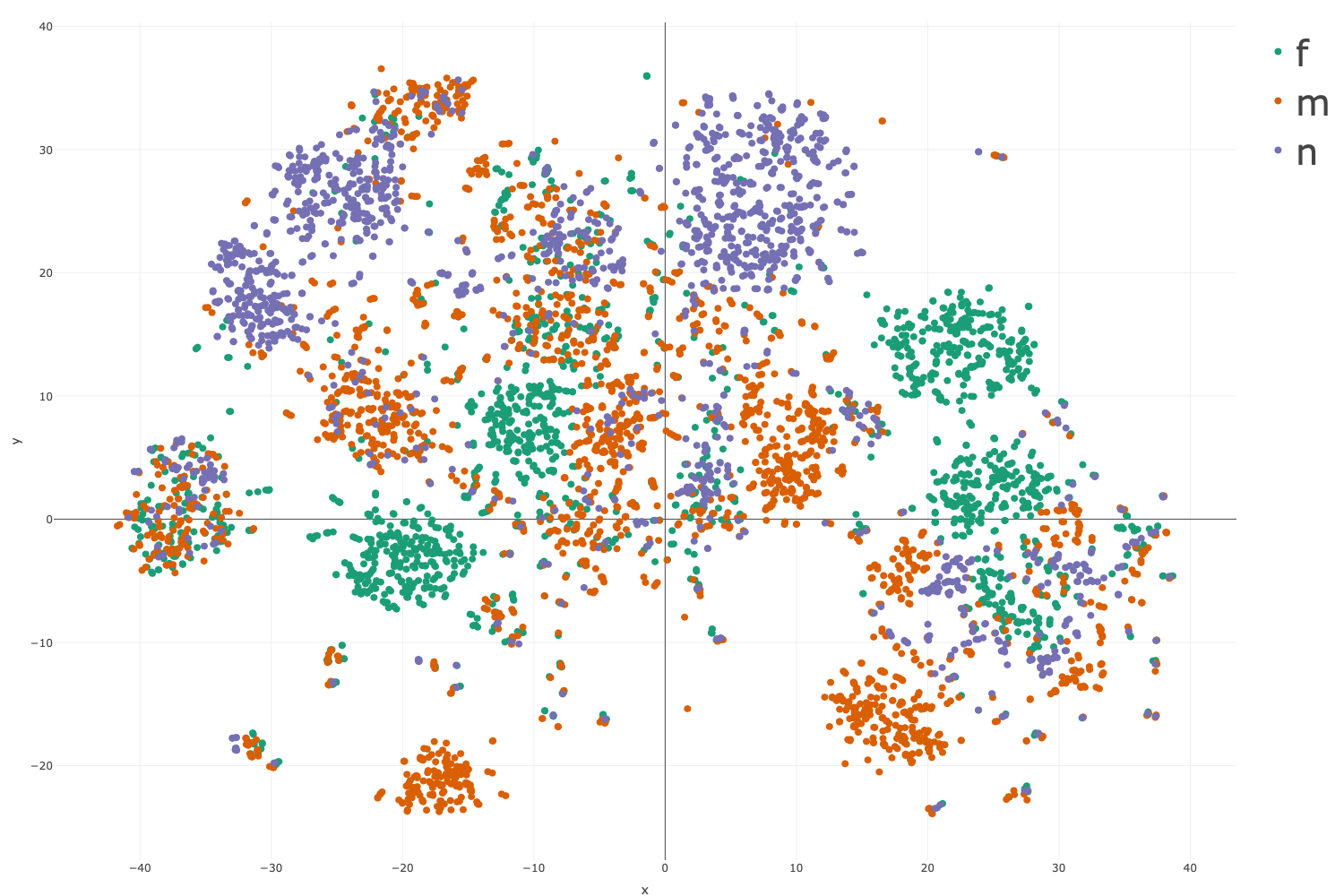}

\ \\

\hspace*{3em}\includegraphics[width=0.6\textwidth, height=6cm]{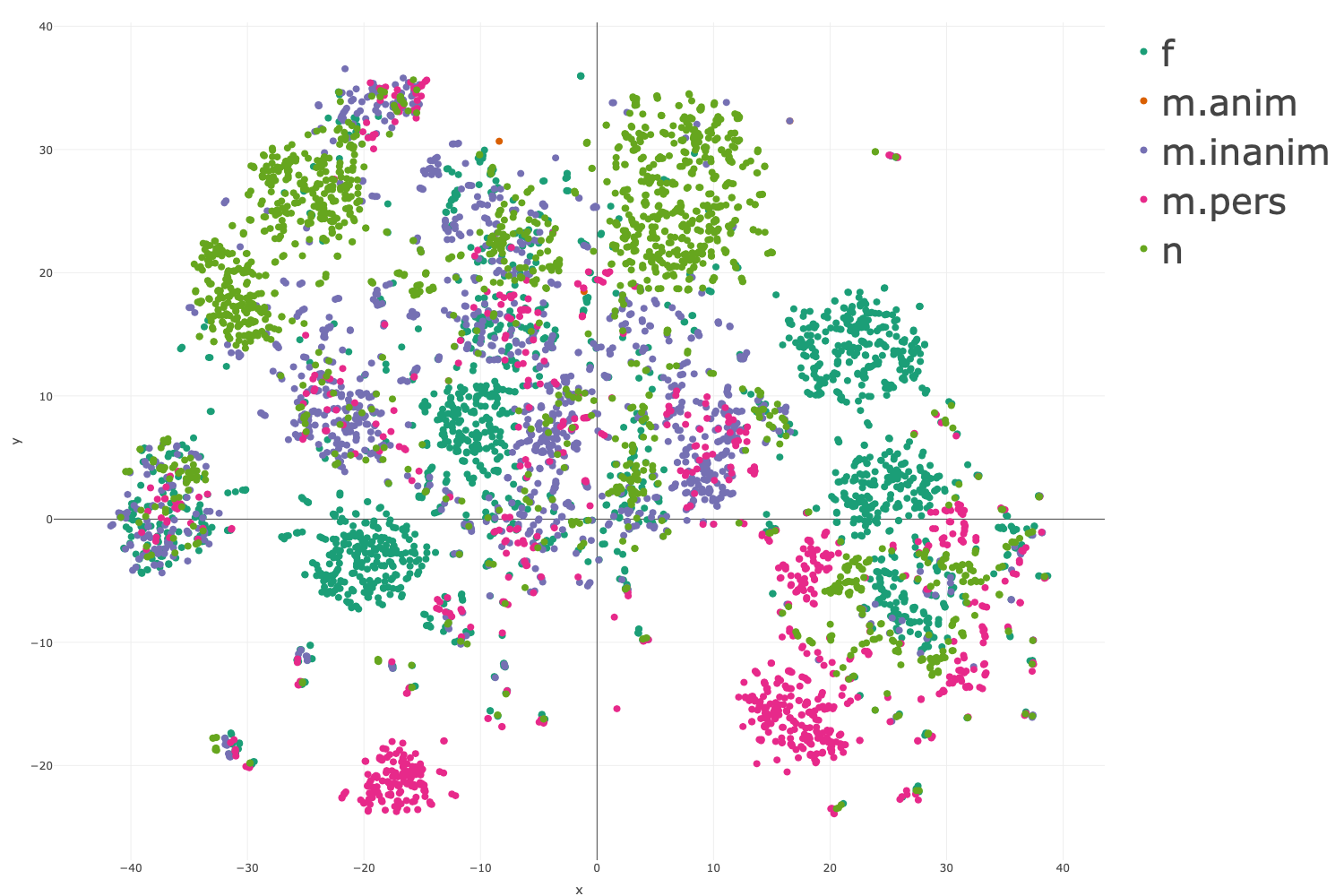}
\caption{Scatterplots of words in a 2-dimensional t-SNE space, colour-coded for three genders (top) and five genders (bottom). An interactive map for three genders is found [\href{run:Gender_tsne.html}{here}]; for five genders [\href{run:Gender_spec_tsne.html}{here}].}
\label{fig:gender}
\end{figure}

Gender in Polish is expressed on nouns, adjectives, gerunds and participles. For this category, many small clusters emerge in the t-SNE plane. The top panel of Figure~\ref{fig:gender} colour-codes the three main genders. Feminine forms (green) create several pure clusters, but also show overlap with masculine (orange) and neuter (purple) forms in several areas. There are also some pure clusters of masculine words found at the bottom of the map, but --- as with feminine forms --- masculine items are also found in clusters shared with other genders. The neuter gender predominates in the upper part of the plane, though it also appears in the lower right. The lower panel of Figure~\ref{fig:gender} colour-codes five gender classes: feminine, masculine animate and inanimate, masculine personal, and neuter forms. Here, too, instances of overlap can be observed. For example, the cluster of masculine inanimate and masculine personal forms in the upper-right region includes adjectives in the nominative/accusative/vocative case. Similarly, the left-most cluster of masculine and neuter forms comprises plural instrumental nouns (above the horizontal axis) and plural instrumental adjectives (below it). The distribution of the neuter gender in the form of distinct clusters in the upper quadrants of both plots mirrors the distribution of gerunds in Figure~\ref{fig:POS} (purple dots). This correspondence reflects the grammatical fact that Polish gerunds are neuter nouns and therefore pattern with neuter forms in their inflectional behaviour.

The confusion matrix for three genders is summarized in Table~\ref{tab:gender_gen}. The general accuracy reaches 87\%, with more confusions compared to the previous grammatical categories, but staying  well above the majority baseline (majority baseline = 38\%, $p < 0.0001$).  The masculine gender is mainly confusable with the feminine (n=165), while the feminine and neuter genders are confusable with the masculine (n=211 and n=163, respectively).  

The five-class division of genders in Table~\ref{tab:gender_spec} leads to a higher number of errors compared to the three-way classification, as evidenced by the lower performance accuracy of the model (83\%, majority baseline = 32\%, $p < 0.0001$), and the increased proportion of misclassified words (error rates = 0.17\% compared to 0.13\% for three genders). Misclassification patterns show that masculine inanimate forms tend to be confused with feminine forms (n=133), and masculine personal forms with inanimate forms (n=155).  As noted above, the overlap of these genders can be observed in the upper-right panel of Figure~\ref{fig:gender}.  Additionally, the particularly poor performance on masculine animate nouns can most likely be attributed to their very low frequency in the dataset (n=5). The corresponding LDA results for \textit{Word2vec} embeddings are summarized in Table~\ref{tab:gender_gen_2vec} and Table~\ref{tab:gender_spec_2vec}.

\begin{table}[H]
\centering
\begin{tabular}{ |c||c|c|c| } 
 \hline
            &feminine   &masculine  &neuter     \\\hline
 \hline
feminine    &1607 (85\%)&211        &70         \\\hline
masculine   &165        &1938 (88\%)&101        \\\hline
neuter      &60         &163        &1446 (87\%)\\\hline
\end{tabular}
\caption{\label{tab:gender_gen}A confusion matrix for three genders; model accuracy = \SI{87}{\percent}.}
\end{table}

\begin{table}[H]
\centering
\begin{tabular}{ |c||c|c|c|c|c| } 
 \hline
            &feminine &m.animate &m.inanimate &m.personal &neuter   \\\hline
  \hline
feminine    &1621(86\%)&0      &192          &4            &71      \\\hline
m.animate   &1         &1(20\%)&2            &1            &0       \\\hline
m.inanimate &133       &5      &1007 (79\%)  &52           &84      \\\hline
m.personal  &42        &2      &155          &705 (77\%)   &14      \\\hline
neuter      &63        &0      &152          &4            &1450 (87\%)\\\hline
\end{tabular}
\caption{\label{tab:gender_spec}A confusion matrix for five genders; model accuracy = \SI{83}{\percent}.}
\end{table}

\subsubsection{Number}\label{sec:analysis_number}

\begin{figure}[H]
\centering
\includegraphics[width=0.6\textwidth]{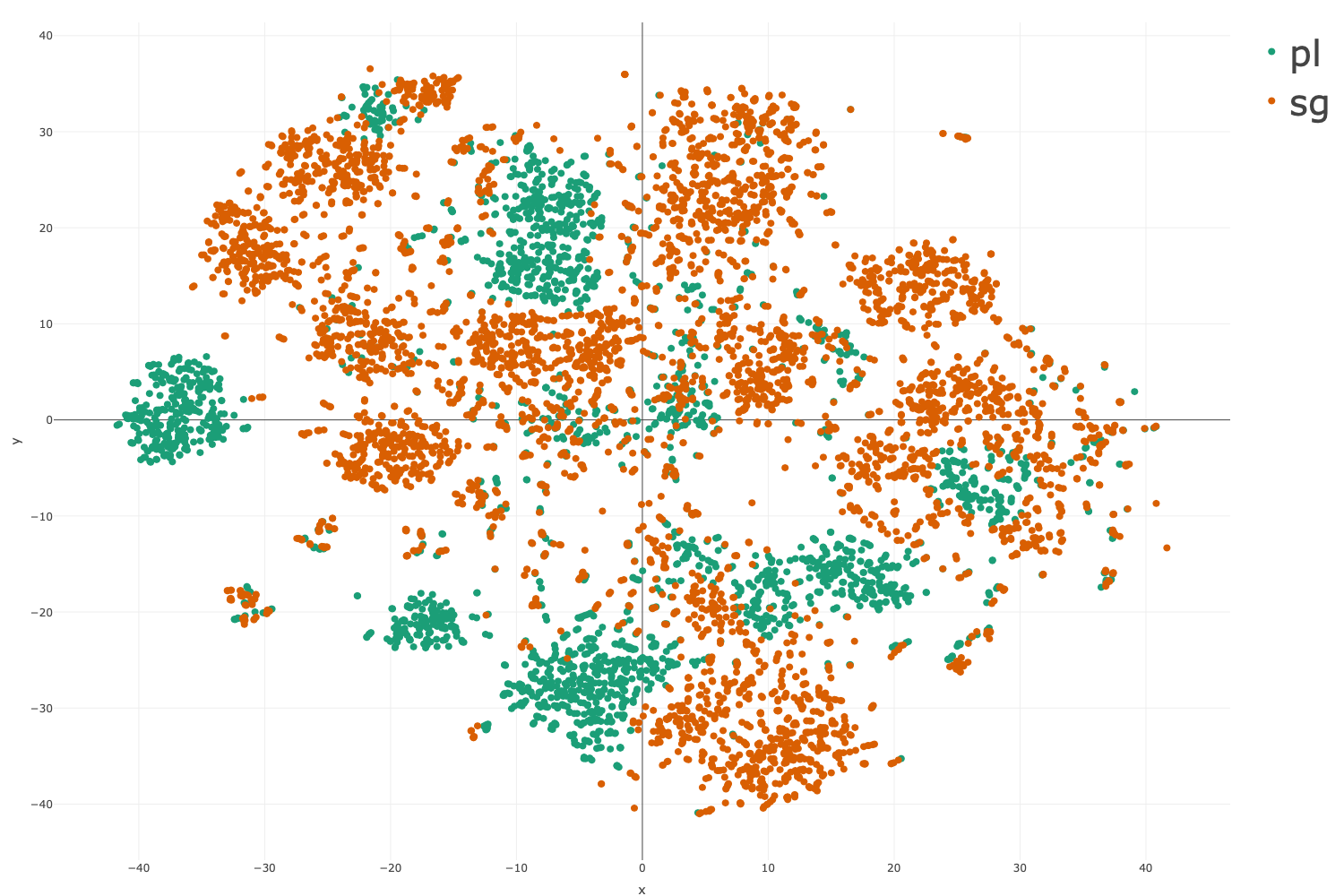}
\caption{\label{fig:number}Scatterplot of words in a 2-dimensional t-SNE space, colour-coded for two numbers. An interactive map is found [\href{run:Number_tsne.html}{here}].}
\end{figure}

The category of number applies to all parts of speech, with the exception of adverbs.  In Figure~\ref{fig:number}, some clear-cut singular (orange) and plural (green) clusters can be seen.  Coloured dots representing both categories occur in a complementary distribution, with few cases of overlap.  Within each number, further smaller subgroupings are determined, \textit{inter alia}, by case and gender.  

In spite of some overlap found in the visual representation of the semantic space, LDA achieves remarkably high classification accuracy for the category of number, above 96\% (majority baseline = 65\%, $p < 0.0001$) (for \textit{Word2vec} results, see Table~\ref{tab:number_2vec} in the Appendix).  Cases of misassignment can be attributed to inflectional syncretism, whereby identical forms represent singular and plural numbers in different cases and genders.  This holds for nouns (e.g. \textit{schroniska} `shelter'-gen.sg./nom.acc.voc.pl), adjectives (e.g. \textit{spostrzegawczy} `observant'-nom.instr.loc.masc.sg./nom.voc.masc-anim.pl.), gerunds and participles (e.g. \textit{wznoszącym} `ascending'-loc.instr.masc.neut.sg./dat.pl).  Yet, in spite of pervasive inflectional syncretism in Polish, the classifier performs well for all the categories that can display syncretic forms, i.e. number, gender, and case. 

\begin{table}[H]
\centering
\begin{tabular}{ |c||c|c| } 
 \hline
            &singular  &plural       \\\hline
   \hline
singular    &4692 (97\%)&134         \\\hline
plural      &157        &2409 (94\%) \\\hline
\end{tabular}
\caption{\label{tab:number}Confusion matrix for number; model accuracy = \SI{96}{\percent}.}
\end{table}

\subsubsection{Intermezzo: number differentiates by case}\label{sec:shift_vectors}

Studies of nominal semantics in Finnish \citep{Nikolaev-etal-2022-nouns-fin} and Russian \citep{Chuang-etal-2023-defective-nouns-rus} suggest that the semantics of plurality differ systematically by case. This was shown by considering the shift vectors between singulars and their corresponding plurals.

Shift vectors for number are obtained by subtracting the embedding of a singular from the embedding of its corresponding plural, while keeping other lexical properties such as case, gender, and part of speech constant, as in the example \textit{kot} `cat'-nom.sg. $ \rightarrow $ \textit{koty}-nom.pl. (compare with the computation of shift vectors for \textit{domini} and \textit{servi} in Section~\ref{sec:dlm}).

\[
\xrightarrow[\text{KOTY}]{plural} \quad = \quad \xrightarrow[\text{KOT}]{singular} \quad + \quad \xrightarrow[\text{Y}]{plural\ shift}
\]

Shift vectors specify how to move in semantic space from a singular inflected form to its corresponding plural.  In standard realizational morphology, number is associated with an abstract feature such as [+plural], a notation that implies that irrespective of case (or gender, or word category), the semantics of plurality are exactly the same. It follows that a linear discriminant analysis should not be able to predict a word's case from its shift vector, and that in a t-SNE plot, all shift vectors for number form one undifferentiated cluster. 

However, \cite{Chuang-etal-2023-defective-nouns-rus} observed that in Russian, the shift vector from a singular to a plural noun embedding varies by each of the six cases. \cite{Nikolaev-etal-2022-nouns-fin} demonstrated the same pattern for Finnish, with plural shift vectors differing for each of the 14 cases in this language.  The morphological similarity of Russian to Polish leads to the question of whether in Polish we can also observe systematic differences between singulars and plurals that vary with respect to case.  

\begin{figure}[H]
\centering
\includegraphics[width=1.0\textwidth]{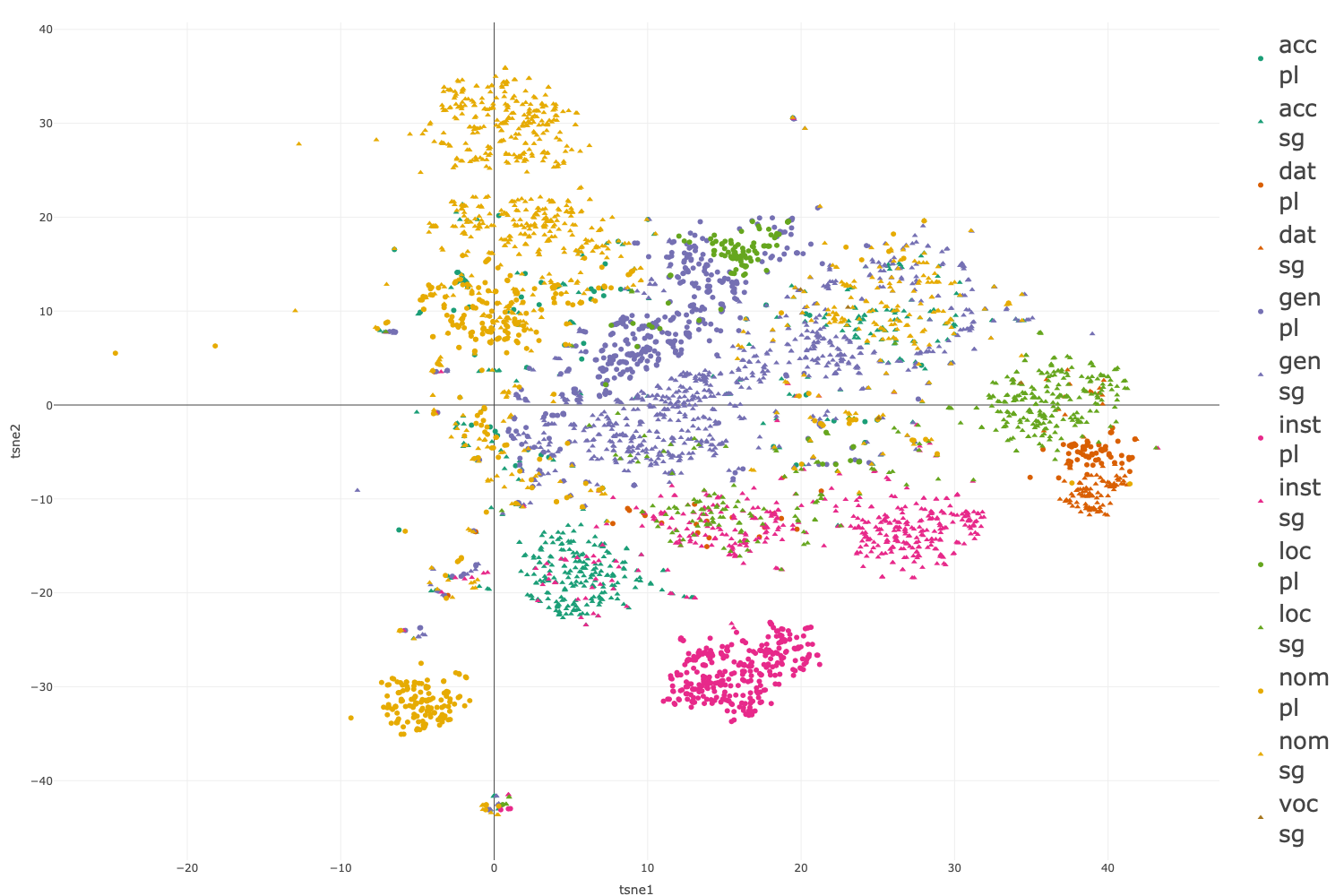}
\caption{\label{fig:casenum}Scatterplot of words in a 2-dimensional t-SNE space, colour-coded for case, and symbol-coded for number. Location of plural and singular forms differs by case.}
\end{figure}

Figure~\ref{fig:casenum} presents a t-SNE cluster analysis with words colour-coded for case, where dots represent plural forms, and triangles singular forms. Importantly, the relative positioning of plural clusters varies within the broader case-based groupings.  This distribution demonstrates that how plurality is represented in semantic space varies by case. For instance, consider the nominative forms (yellow). The top-most clusters predominantly contain singulars (triangles), while the lower clusters comprise plurals (dots). A comparable pattern holds for the instrumental case (pink), where two singular clusters are situated above the plural cluster. However, the dative case (orange) exhibits the reverse relation: singular forms cluster below plural forms. Yet, a different pattern is observed for the genitive (purple) and locative (light green) cases, where singulars are dispersed toward the bottom-right of the plural groupings. Taken together, these observations suggest informally (informally because t-SNE does not necessarily respect the geometry of the original semantic space) that the transition from singular to plural in semantic space is case-specific, rather than uniform across the nominal system.

To follow up on this possibility, we calculated the shift vectors from singular to plural words for the different cases.  For each case and gender combination, we selected all pairs of singulars and plurals for which both numbers are available, resulting in a dataset of 755 singular-plural word pairs.  Each pair represents the same part of speech.  For these pairs, we computed the shift vectors, and used t-SNE to visualize potential clustering, with colour-coding by case.  

\begin{figure}
\centering
\includegraphics[width=1.0\textwidth]{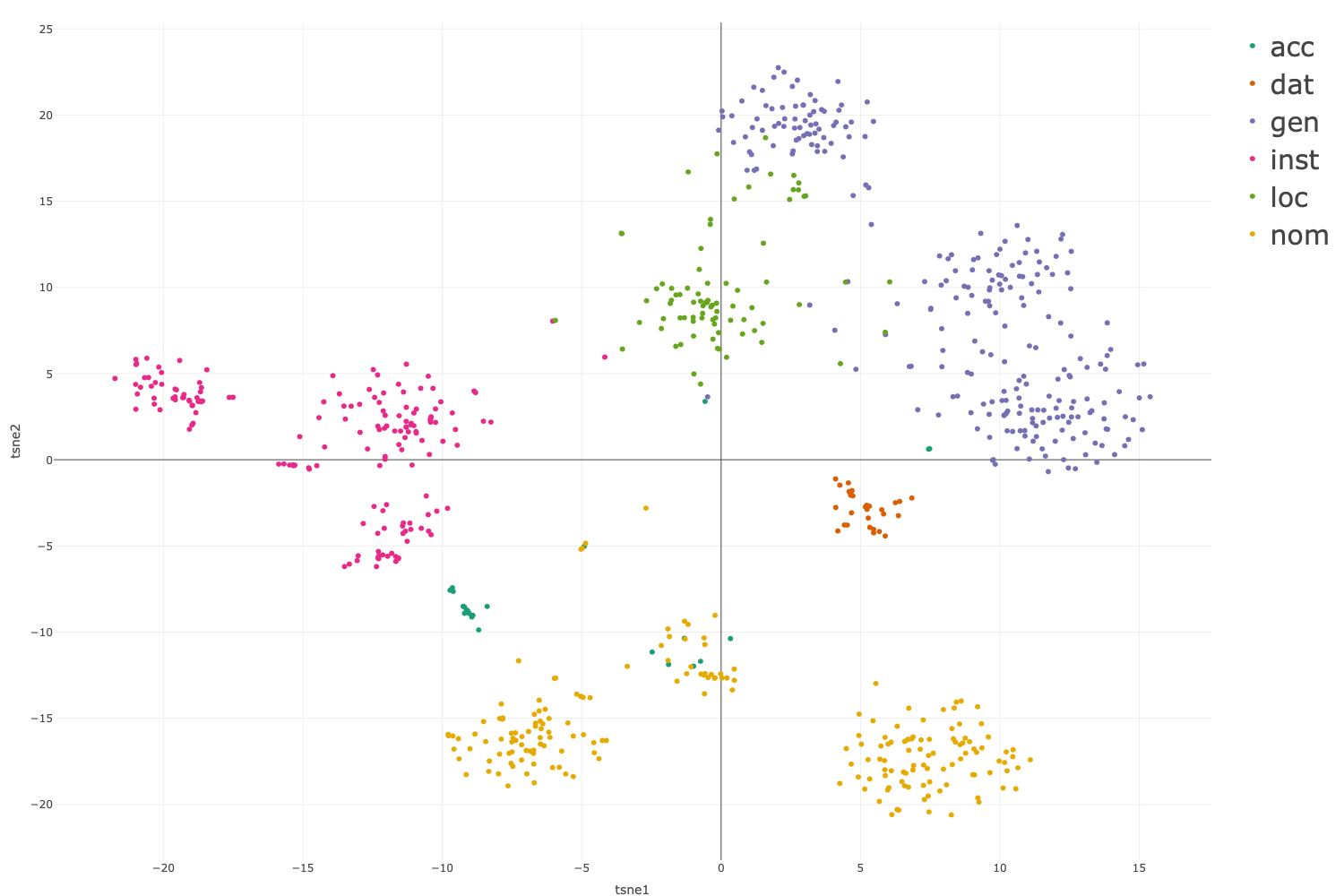}
\caption{\label{fig:shiftcectors_casenum}Scatterplot of the plural shift vectors of words in t-SNE space, colour-coded for case. Data points clearly cluster by case. Sub-clusters within each case comprise different genders and parts of speech.}
\end{figure}

Figure~\ref{fig:shiftcectors_casenum} shows remarkably well-separated clusters for the different cases, suggesting that the details of the semantics of pluralization vary with case.  This well-defined distribution is confirmed by a linear discriminant analysis that, using leave-one-out cross-validation, achieves an accuracy equal to 91\%, which compares favourably to a majority baseline of 34\%.  

Figure~\ref{fig:shiftcectors_casenum} indicates that within the clusters for the different cases, there are sub-clusters. Within the larger groupings of nominatives (yellow), genitives (purple) and instrumentals (pink), smaller clusters are present. For the nominative case (yellow), the cluster to the left of the vertical axis contains the feminine participles; the cluster that straddles the vertical axis contains the masculine nouns and participles; the cluster in the bottom-right quadrant primarily represents the masculine participles and adjectives. The three groupings for the genitive case (purple) contain, starting from the top of the plot, feminine participles and adjectives, followed by neuter nouns and gerunds, and finally masculine participles and adjectives positioned along the horizontal axis. For the instrumental case (pink), two clusters in the upper left quadrant of the plot feature masculine nouns grouped with neuter nouns and gerunds, as well as masculine adjectives grouped with masculine participles. By contrast, a cluster containing feminine nouns and participles is located in the lower-left quadrant.  

For the accusative (the small dark green cluster in the lower-left quadrant), only a few singular–plural pairs of words are available. This sparsity results from the extensive syncretism between the accusative, nominative, and vocative cases (see Section~\ref{sec:morphology}), combined with our decision to retain, for any set of syncretic forms, only the most frequent word in the dataset. We further note that in the case of the vocative, no singular–plural pairs were attested in our dataset.

Overall, the distribution of shift vectors indicates that the conceptualization of plurality varies systematically across cases and that within a given case it may further differ by gender and/or part of speech.

A parallel case-by-number analysis for \textit{Word2vec} embeddings presented as Figure~\ref{fig:casenum_2vec} in the Appendix shows clear and well-separated plural clusters for the nominative and instrumental case, and an overlap between singular and plural forms for other cases.  The analysis of shift vectors presented as Figure~\ref{fig:shiftcectors_casenum_2vec} suggests that the nominative, genitive, and instrumental cases are almost perfectly separated by the vertical and horizontal axes.  Shift vectors for the locative case are found in the lower-right quadrant of the plane, whereas for the accusative case between the nominatives and instrumentals.  This distribution is further supported by a linear discriminant analysis, which reaches an accuracy of 82\%, compared to a majority baseline of 37\% (compare with 91\% when using \textit{fastText} embeddings, with a majority baseline of 34\%).  

\subsubsection{Aspect}\label{sec:analysis_aspect}

Following the tagging scheme used in the NKJP300M corpus, we examined how aspect is realized on verbs, gerunds, and participles.  Many of the perfective forms are prefixed with one of the aspectual markers \{s- z- w- ws-\} discussed in Section~\ref{sec:aspect phonotactics}. 

\begin{figure}[H]
\centering
\includegraphics[width=0.6\textwidth]{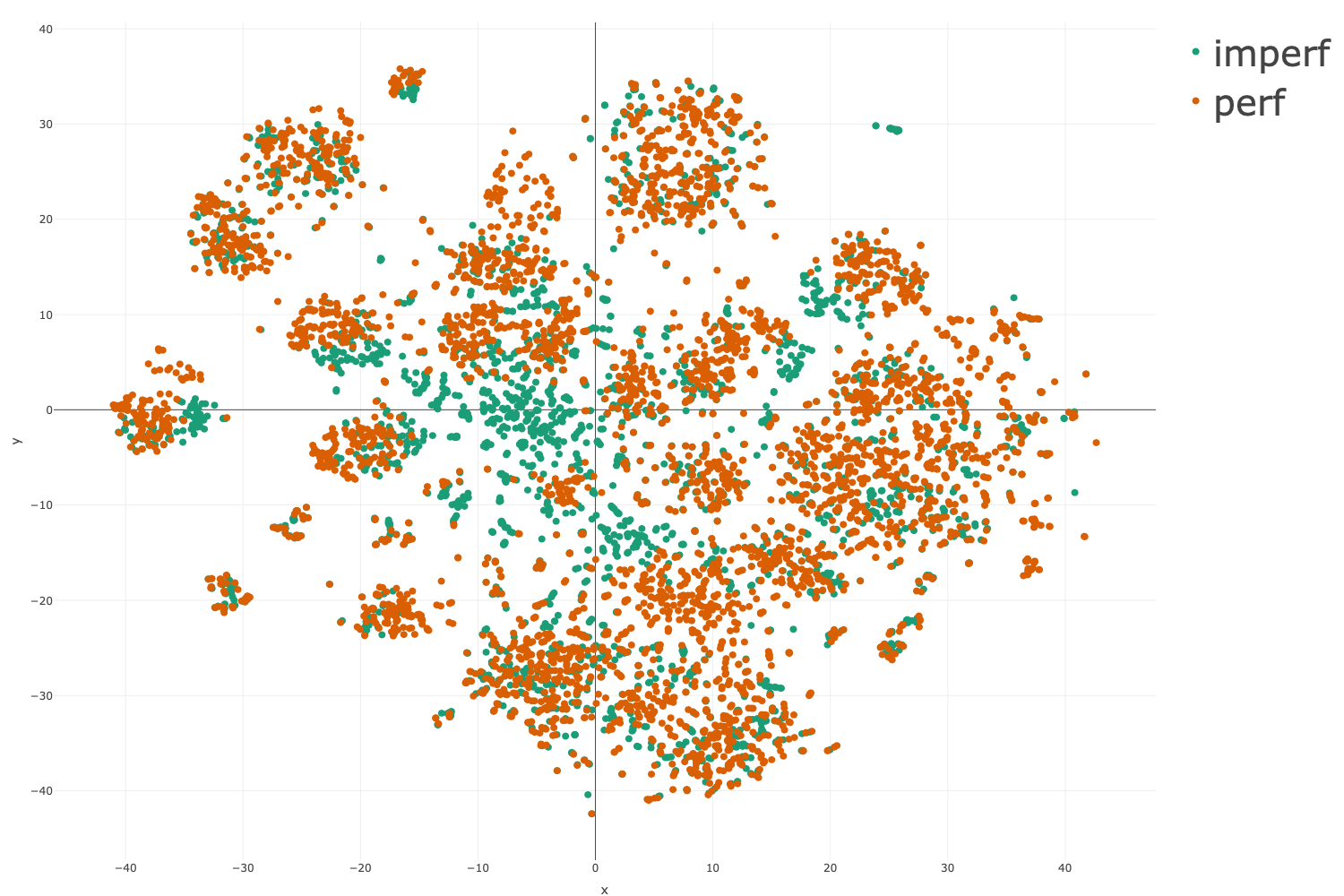}
\caption{Scatterplots of words in a 2-dimensional t-SNE space, colour-coded for two aspects. An interactive map is found [\href{run:Aspect_tsne.html}{here}].}\label{fig:aspect}
\end{figure}

A t-SNE map presented in Figure~\ref{fig:aspect} reveals a less clear picture compared to the previous categories.  The perfective and imperfective forms are scattered across the map.  There is substantial overlap between the two classes,  although the central area of the map is dominated by imperfectives (present verbs and active adjectival participles).  The most pronounced grouping of perfective forms, found in the bottom-right quadrant of the map, is represented by future verbs.  

Apparently, aspectual information is spread out across many dimensions of the embeddings, making it impossible for the t-SNE algorithm to project the high-dimensional data points into a two-dimensional plane.  A similar finding was reported for German particle verbs in \cite{Stupak-baayen-2022-particle-verbs}. Nevertheless, a simple LDA classifier reaches a very high prediction accuracy of 97\% (majority baseline = 60\%, $p < 0.0001$), and a very low error rate of 0.03\%. The results are summarized in Table~\ref{tab:aspect} (for a comparison with \textit{Word2vec} embeddings, see Table~\ref{tab:aspect_2vec} in the Appendix).  

\begin{table}
\centering
\begin{tabular}{ |c||c|c| } 
 \hline
                &imperfective  &perfective      \\\hline
   \hline
imperfective    &2643 (96\%)    &105            \\\hline
perfective      &112            &3982 (97\%)    \\\hline
\end{tabular}
\caption{\label{tab:aspect}A confusion matrix for aspect; model accuracy = \SI{97}{\percent}.}
\end{table}

The analyses presented so far show that a range of morpho-syntactic properties can be predicted reliably from word embeddings, although for one category (aspect), this predictability is only partly reflected in the corresponding t-SNE plot. The question to which we turn now is whether phonological and morphonological properties of words are likewise encoded in semantic space. To this end, the next sections examine four properties pertaining to the (morpho)phonological composition of word-initial clusters. Given the patterns observed for aspect, we anticipate that these properties will yield less clearly defined clusters in the t-SNE space. The key question is whether a linear discriminant analysis can nonetheless achieve robust prediction accuracy.

\subsubsection{Prefix type}\label{sec:analysis_prefix}

\begin{figure}[H]
\centering
\includegraphics[width=0.6\textwidth]{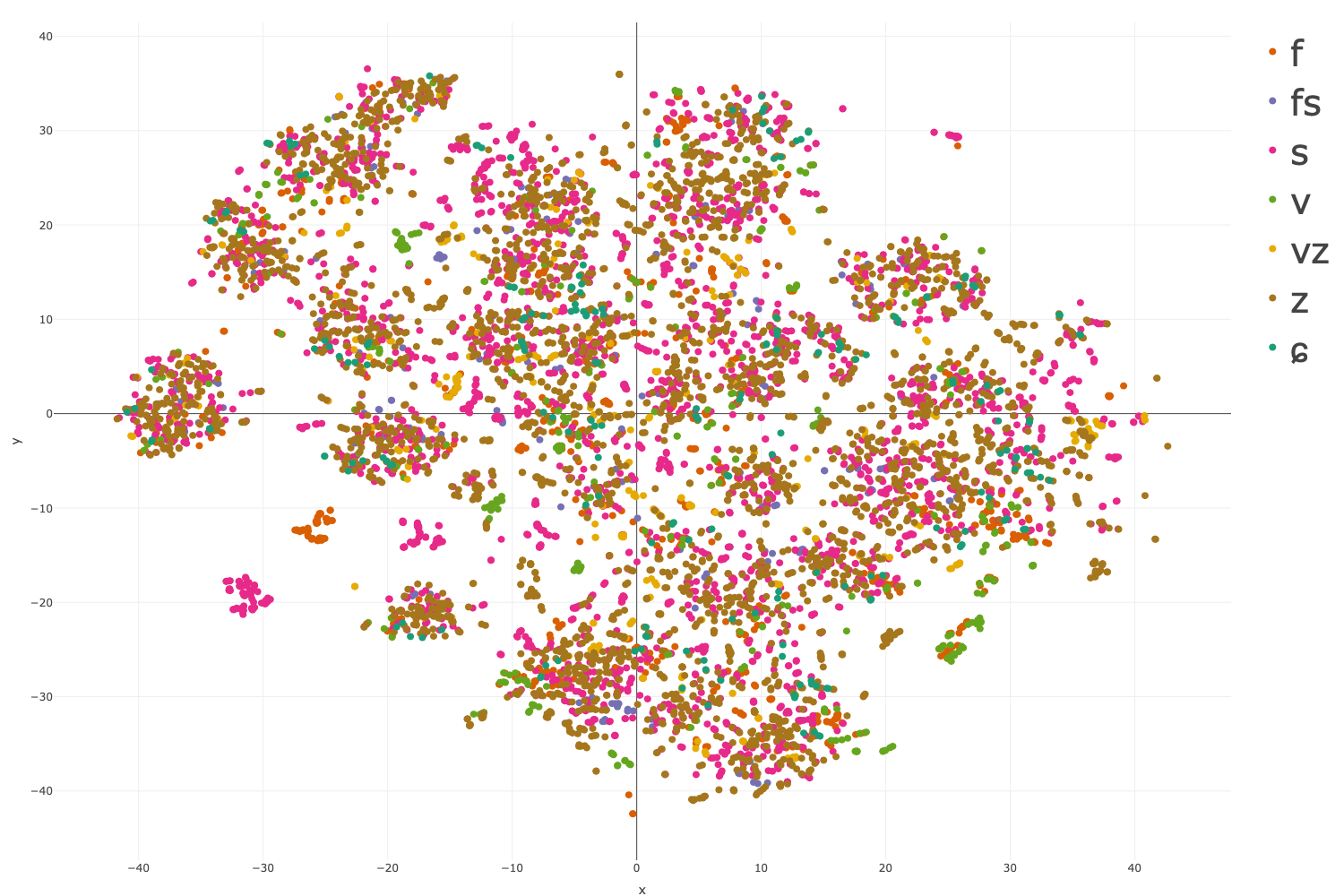}
\caption{\label{fig:prefix}Scatterplot of words in a 2-dimensional t-SNE space, colour-coded for seven prefixes.  An interactive map is found [\href{run:Prefix_tsne.html}{here}].}
\end{figure}

Let us first inspect the classification of the consonant prefixes, which are phonetically realized as /s z f v fs vz/ and /\textipa{\textctc}/ (which is a realization of \{s-\} in the context before /\texttctclig/, as in \textit{ś+ciskać} ‘to squeeze’). As can be observed in Figure~\ref{fig:prefix}, the voiceless and voiced variants as well as the simplex and complex variants exhibit substantial overlap in semantic space.\footnote{Exploration of the interactive version of this figure clarifies that the bi-consonantal prefixes /fs/ and /vz/ display a somewhat more complementary distribution with respect to one another, however, they largely overlap with prefixes composed of one consonant.} Overall, the t-SNE map would suggest that the phonological properties of the prefixes do not particularly serve word semantics. 

However, surprisingly, the LDA analysis presented in Table~\ref{tab:prefix} shows that the accuracy of model predictions reaches 97\%, under leave-one-out cross-validation (majority baseline = 39\%, $p < 0.0001$).  Misclassified  forms are found for each prefix under consideration, yet the overall error rate is very small (0,03\%).  This result aligns with the well-defined meaning of the prefixes, as discussed in Section~\ref{sec:aspect meaning}.  The semantics of \{s- z- w- ws-\} thus seems to be reflected in the following factors: (1) their mono- or bi-consonantal realization (e.g. /v/ vs. /vz/), (2) their place articulation characteristics (e.g. labio-dental /f/ vs. dental /s/ vs. alveolo-palatal /\texttctclig/), and (3) their voicing (e.g. voiceless /s/ vs. voiced /z/).  A comparison with \textit{Word2vec} embeddings are presented in Table~\ref{tab:prefix_2vec} in the Appendix. Here, too, a linear discriminant analysis performs far better than the majority baseline.  In section~\ref{sec:analysis_summary}, we address the implications of this divergence between unsupervised clustering with t-SNE and supervised classification with LDA. 

\begin{table} [H]
\centering
\begin{tabular}{ |c||c|c|c|c|c|c|c| } 
 \hline  
        &/\textipa{\textctc}/ &/f/    &/fs/   &/s/    &/v/    &/vz/   &/z/    \\\hline
      \hline
/\textipa{\textctc}/  &207 (96\%)&0      &0      &8      &0      &0      &0      \\\hline
/f/                 &0      &728 (96\%)&10     &5      &14     &0      &4      \\\hline
/fs/                &0      &0      &218 (90\%)&4      &1      &4      &15     \\\hline
/s/                 &5      &6      &7      &2799 (97\%)&5      &4      &55     \\\hline
/v/                 &0      &20     &3      &9      &348 (88\%)&1      &3      \\\hline
/vz/                &0      &0      &7      &0      &0      &336 (95\%)&12     \\\hline
/z/                 &6      &4      &17     &9      &5      &27     &3091 (98\%)\\\hline
\end{tabular}
\caption{\label{tab:prefix}A confusion matrix for prefix type; model accuracy = \SI{97}{\percent}.}
\end{table}

\subsubsection{Morphotactic parsability of clusters}\label{sec:analysis_parsability}

\begin{figure}[H]
\centering
\includegraphics[width=0.6\textwidth]{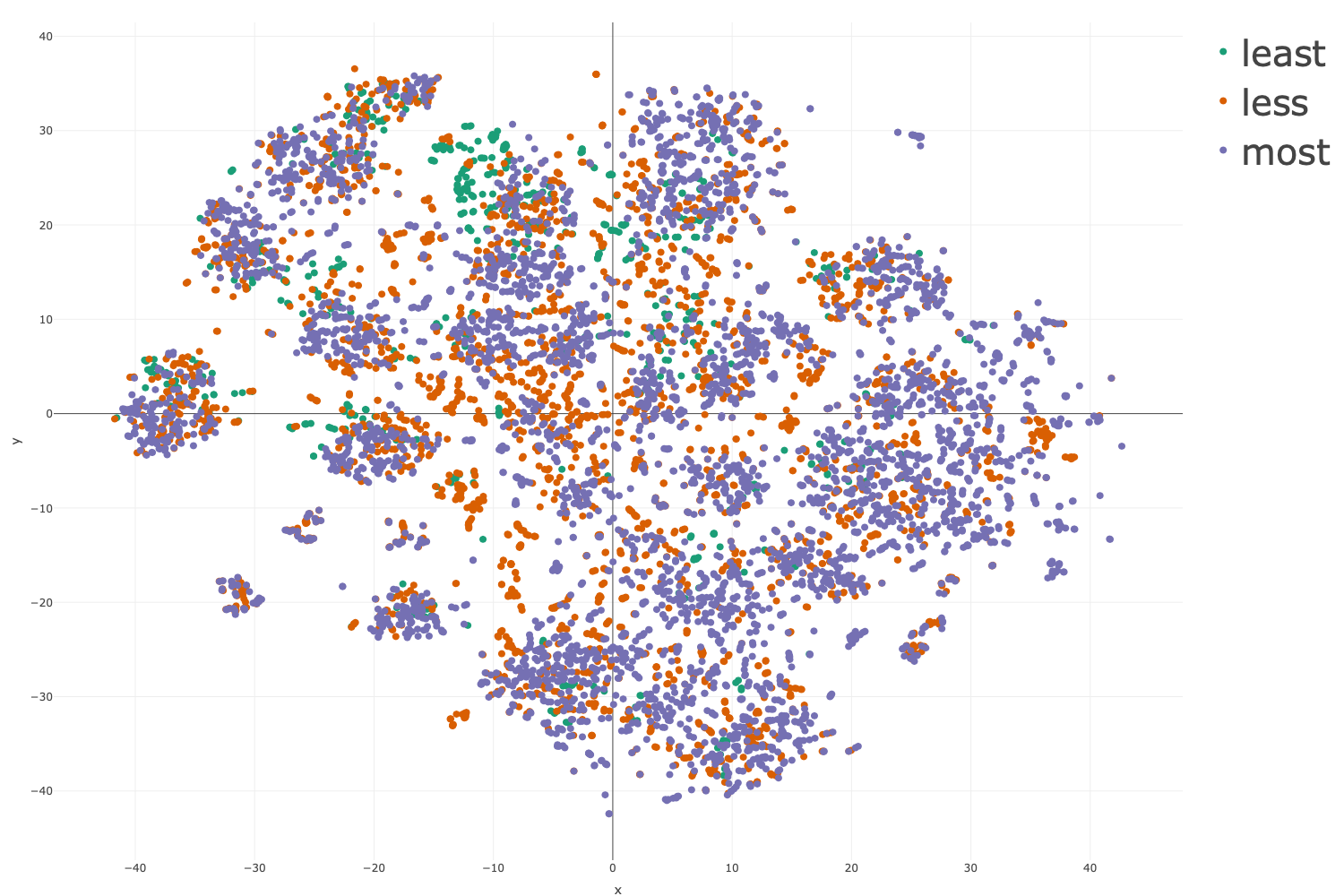}
\caption{\label{fig:parsability}Scatterplot of words in a 2-dimensional t-SNE space, colour-coded for three degrees of morphotactic parsability.  An interactive map is found [\href{run:Transparency_tsne.html}{here}].}
\end{figure}

The next criterion specifying the morphological properties of consonant clusters is the degree to which the consonantal prefix is separable from a consonant-initial stem.  In this section of the paper, we therefore ask to what extent morphotactic parsability is reflected in Polish semantic space.  The t-SNE plot in Figure~\ref{fig:parsability} shows that the three degrees of morphotactic segmentability cannot be clearly distinguished.  Although certain regions of the map are dominated by a single parsability type, the three categories nonetheless display substantial overlap. For example, several groupings of less parsable consonant clusters (orange) appear in the central portion of the plane, while small groupings of the most parsable consonant clusters (purple) lie toward the periphery of the plot. The question then is to what extent supervised classification with LDA can differentiate between these morphotactic classes.

As observed above for aspect and consonantal prefixes, LDA (under leave-one-out cross-validation) performs with much higher accuracy than the t-SNE map would leave one to believe.  For the three parsability classes, however, prediction accuracy is now `only' at 82\%.  Nevertheless, this is still substantially above a simple majority baseline (56\%, $p < 0.0001$, proportions test). As shown in Table~\ref{tab:parsability_2vec} in the Appendix, the same pattern is observed for \textit{Word2vec} embeddings.

The classifier performs relatively well at recognizing the most decomposable context, i.e. the one in which the prefix attaches to a free stem.  Prediction accuracy is greatest for the most parsable class of clusters, intermediate for the "less" parsable class, and lowest for the least parsable class. In other words, the less parsable an initial cluster is on the form side, the less distinctly its semantics are represented in semantic space.

\begin{table}[H]
\centering
\begin{tabular}{ |c||c|c|c| } 
 \hline
   observed/predicted &most parsable    &less parsable  &least parsable\\\hline
 \hline
most parsable       &3955 (89\%)        &439            &70             \\\hline
less parsable       &608                &2154 (76\%)    &88             \\\hline
least parsable      &132                &83             &468 (67\%)            \\\hline
\end{tabular}
\caption{\label{tab:parsability}Confusion matrix for the degrees of morphotactic parsability; model accuracy: \SI{82}{\percent}.}
\end{table}

\subsubsection{Cluster markedness}\label{sec:analysis_markedness}

\begin{figure}[H]
\centering
\includegraphics[width=0.6\textwidth]{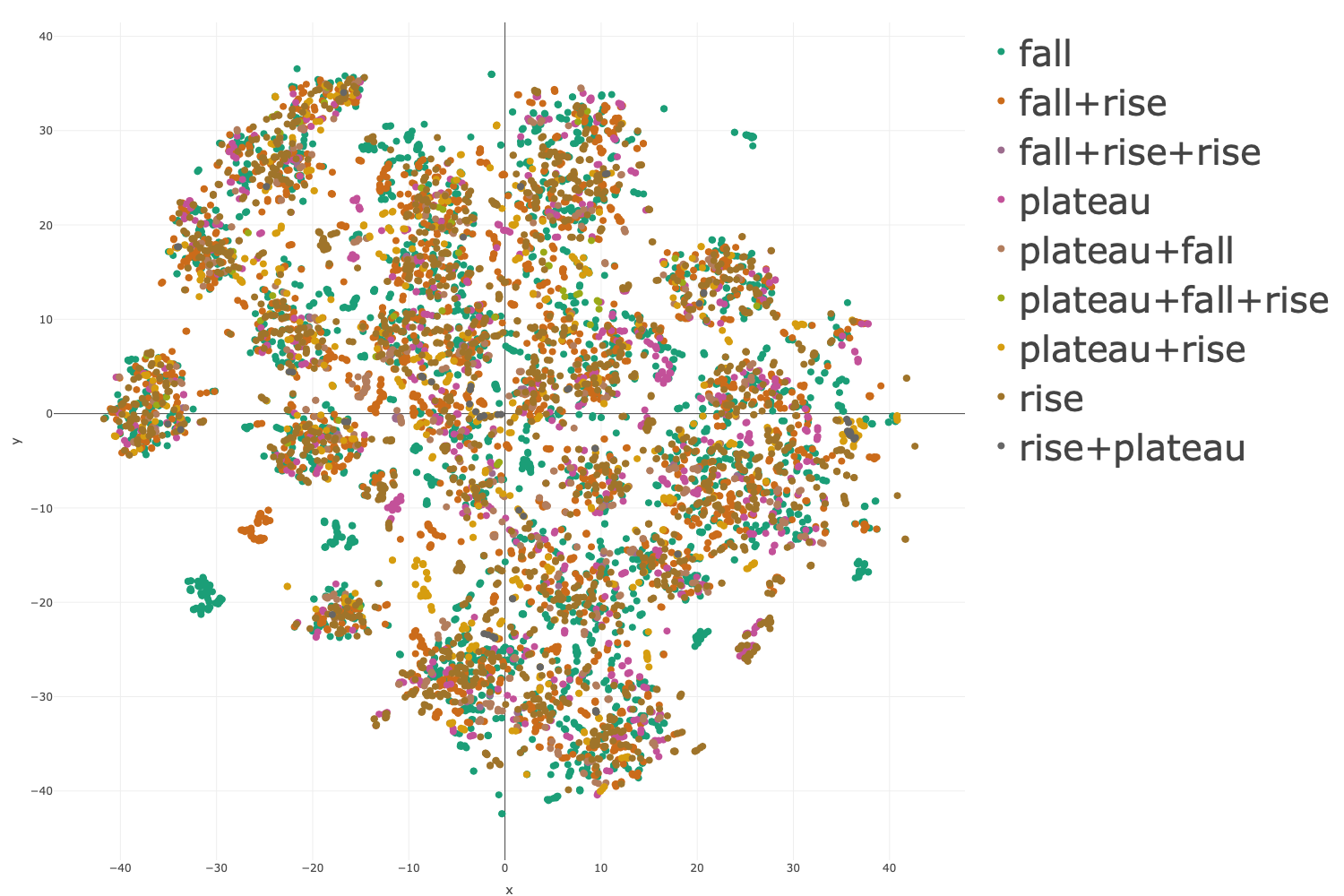}
\caption{\label{fig:markedness}Scatterplot of words in a 2-dimensional t-SNE space, colour-coded for nine sonority profile types. An interactive map is found [\href{run:Markedness_tsne.html}{here}].}
\end{figure}

Cluster markedness shows to be even a weaker predictor of word semantics.  As shown in the t-SNE plot, presented as Figure~\ref{fig:markedness}, there is a great overlap between groups representing all types of sonority slopes. Several very small groupings can be found on the outer edge of the plot, particularly for forms with a sonority fall (green) represented by fricative $+$ plosive clusters. For example, /sp/-initial words are found in the bottom-left and top-left part of the map; the groupings for /zg zd/ appear in the bottom-right; and a grouping for /sk/ is located in the central part of the plane. Isolated cases of overlap can be reported for sonority rise (brown) and plateau (pink) forms, as well as for plateau $+$ rise (yellow) and rise $+$ plateau (gray).

The overlap between the categories is reflected in the LDA analysis in Table~\ref{tab:markedness} (for a comparison with \textit{Word2vec} embeddings, see Table~\ref{tab:markedness_2vec} in the Appendix). Similarly to morphotactic parsability, the accuracy of the classifier (under leave-one-out cross-validation) is relatively low, at 82\%, but substantially exceeds the majority baseline (30\%, $p < 0.0001$, proportions test). There is substantial confusability for some markedness categories. For example, clusters representing a mixed fall $+$ rise profile are often misclassified as pure fall (16\%), and plateau $+$ rise clusters as rise (7\%). Also note that plateaus are confusable with several types of sonority slopes: fall (8\%), rise (8\%) and fall $+$ rise cases (9\%).  

\begin{landscape}
\begin{table} [H]
\centering
\begin{tabular}{ |c||c|c|c|c|c|c|c|c|c| } 
 \hline
         &F     &F+R    &F+R+R  &plat   &plat+F &plat+F+R   &plat+R &R      &R+plat \\\hline
  \hline
F        &1906 (91\%)  &235    &3      &55     &18     &5          &27     &121    &9      \\\hline
F+R      &317   &1583 (79\%)   &1      &23     &2      &2          &16     &55     &2      \\\hline
F+R+R    &1     &0      &66 (99\%)     &0      &0      &0          &0      &0      &0      \\\hline
plat     &60    &66     &1      &519 (71\%)    &7      &1          &13     &59     &1      \\\hline
plat+F   &15    &9      &0      &11     &372 (89\%)    &0          &8      &9      &0      \\\hline
plat+F+R &4     &1      &0      &1      &0      &40 (83\%)         &0      &2      &0      \\\hline
plat+R   &15    &15     &0      &6      &1      &0          &463 (78\%)    &40     &55     \\\hline
R        &51    &54     &3      &32     &9      &4          &23     &1515 (89\%)   &3      \\\hline
R+plat   &0     &0      &0      &0      &0      &0          &0      &0      &62 (100\%)     \\\hline
\end{tabular}
\caption{\label{tab:markedness}A confusion matrix for sonority profile types; model accuracy = \SI{82}{\percent}. F = fall, R = rise, plat = plateau.}
\end{table}
\end{landscape}

\subsubsection{Cluster size}\label{sec:analysis_size}

\begin{figure}[H]
\centering
\includegraphics[width=0.6\textwidth]{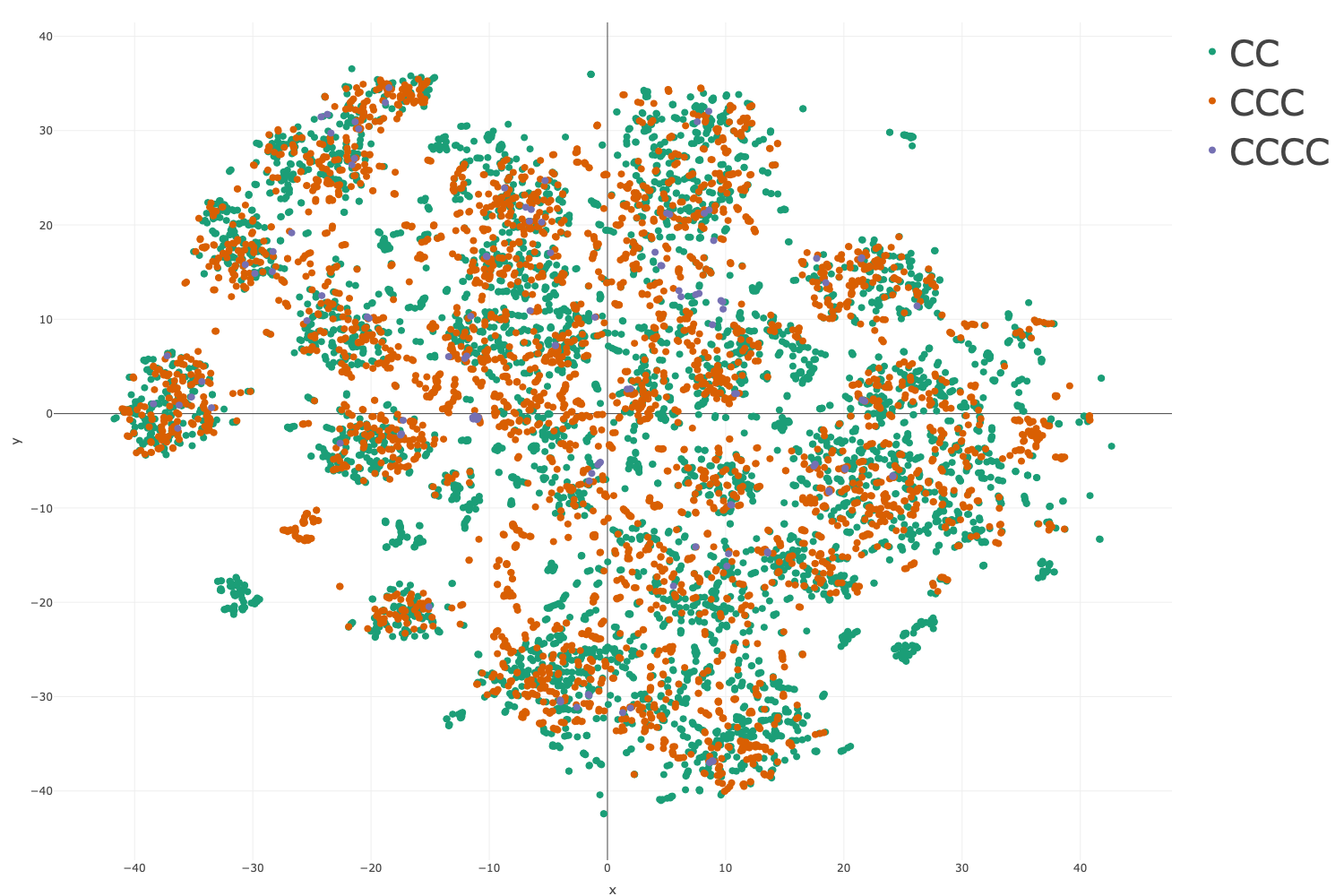}
\caption{\label{fig:size}Scatterplot of words in a 2-dimensional t-SNE space, colour-coded for three cluster sizes.  An interactive map is found [\href{run:Cluster_size_tsne.html}{here}].}
\end{figure}

Similarly to other (mor)phonological properties, a t-SNE unsupervised clustering analysis does not reveal any clear groupings for cluster length.  As shown in Figure~\ref{fig:size}, consonant clusters CC, CCC, and CCCC are scattered across the t-SNE plane.  Several small groupings are found, in particular on the periphery of the plane. The groupings map words belonging to the same morphological family or inflectional paradigm. CC clusters (green) include /sp vz zm zg/, while CCC clusters (orange) include /fpw zvj vzm sp\textesh/. For example, two /sp/ groupings (bottom left) comprise inflected verbs, gerunds, and participles of \textit{spędzać} `to spend' and \textit{spełniać} `to fulfil'. Similarly, the /fs/ and /zg/ groupings (bottom right) map various forms of \textit{wsypać} `to pour into' and \textit{zgasnąć} 'to go out (fire)'. By contrast, four-member clusters are spread out relatively evenly across the map. 

The results for the linear discriminant analysis are presented in Table~\ref{tab:size} (for a comparison with \textit{Word2vec} embeddings, see Table~\ref{tab:size_2vec} in the Appendix).  The accuracy of the predictions reaches 81\% (majority baseline = 57\%, $p < 0.0001$, proportions text). Misclassifications are present in each class.  CCs are confusable with CCCs (14\%), CCCs are confusable with CCs (24\%), and CCCCs tend to be misclassified as CCs (10\%).  This difference between the relatively accurate LDA results and the t-SNE visualization aligns with other (mor)phonological properties of clusters presented in Sections~\ref{sec:analysis_prefix} through~\ref{sec:analysis_markedness}. 

\begin{table}
\centering
\begin{tabular}{ |c||c|c|c| } 
 \hline
        &CC             &CCC            &CCCC       \\\hline
  \hline
CC      &3877 (85\%)    &635            &31         \\\hline
CCC     &804            &2528 (76\%)    &10         \\\hline
CCCC    &12             &7              &96 (83\%)  \\\hline
\end{tabular}
\caption{\label{tab:size}A confusion matrix for cluster size; model accuracy = \SI{81}{\percent}.}
\end{table}

\subsection{Comparison of \textit{fastText} and \textit{Word2vec} embeddings}\label{sec:analysis_summary}

\begin{figure}[H]
\centering
\includegraphics[width=0.8\textwidth]{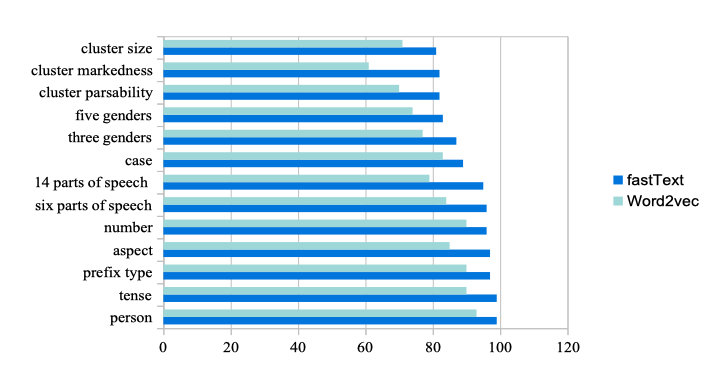}
\caption{\label{fig:comparison_accuracies}A comparison of accuracy results for two sets of embeddings:\textit{fastText} and \textit{Word2vec}.}
\end{figure}

Let us now compare the accuracies for \textit{fastText} with the results for \textit{Word2vec},  visualized in Figure~\ref{fig:comparison_accuracies}. We can observe a weaker performance of the LDA classifier for \textit{Word2vec} embeddings: for each category, the overall accuracy of \textit{fastText}-based LDA analyses is higher compared to \textit{Word2vec}  (mean difference = 10,5\% points). Several shared classification patterns emerge. First, in both analyses, the highest prediction accuracies are obtained for person.  Second, high accuracy results are also reported for tense, prefix type, aspect, and number. Finally, the least accurate predictions are made for the (mor)phonological properties of words, namely cluster size, cluster markedness, and morphotactic parsability.  

Moreover, for all categories, the accuracies achieved for the LDA classifier are much higher compared to the baseline, as evidenced in Table~\ref{tab:embeddigs_summary}.  In particular, the analysis of 14 parts of speech using \textit{fastText} shows that the prediction accuracy of LDA and a majority baseline differ by 73\% points (95\% vs. 22\%).  Although the results for \textit{Word2vec} are lower compared to \textit{fastText}, they are always also better compared to a majority baseline. Importantly, the comparison of \textit{fastText} and \textit{Word2vec} clarifies that the general pattern of results does not hinge on embeddings that make use of subword n-grams that are in partial correspondence with stems and exponents. The results for \textit{Word2vec} indicate that the collocational patterns in utterances are reflecting to a considerable extent aspects of the words' phonological and morphological structure.

\begin{table}[H]
\centering
\begin{tabular}{ |c||c|c|c|c|c|c| } 
 \hline
category           & \multicolumn{3}{c|}{\textit{fastText}} &   \multicolumn{3}{c|}{\textit{Word2vec}}  \\ \hline 
                   & LDA      & baseline & N & LDA & baseline & N \\ \hline
person              &99\%           &60\% &2853     &93\%           &84\% &1596 
 \\\hline
tense               &99\%           &49\% &2481     &90\%           &65\% &1505    \\\hline
prefix type         &97\%           &40\% &7997     &90\%           &41\% &5936   \\\hline
aspect              &97\%           &60\% &6842     &85\%           &59\% &4905   \\\hline
number              &96\%           &65\% &7392     &90\%           &68\% &5393   \\\hline
six parts of speech &96\%           &41\% &7997     &84\%           &34\% &5936   \\\hline
14 parts of speech  &95\%           &22\% &7997     &79\%           &23\% &5936   \\\hline
case                &89\%           &28\% &4535     &83\%           &29\% &3797   \\\hline
three genders       &87\%           &38\% &5761     &77\%           &39\% &4782   \\\hline
five genders        &83\%           &33\% &5761     &74\%           &32\% &4782   \\\hline
cluster parsability &82\%           &56\% &7997     &70\%           &54\% &5936   \\\hline
cluster markedness  &82\%           &30\% &7997     &61\%           &29\% &5936   \\\hline
cluster size        &81\%           &57\% &7997     &71\%           &55\% &5936   \\\hline
\end{tabular}
\caption{\label{tab:embeddigs_summary}A comparison of the results (model accuracy and majority baseline) for two sets of embeddings.  The data are ordered according to the descending accuracy results for \textit{fastText}, followed by \textit{Word2vec} embeddings. $N$:total number of observations.}
\end{table}

\section{Orthogonalization of semantic space}\label{sec:orthogonalization_fastText}

In the analyses reported above, we have observed good classification accuracy with linear discriminant analysis across all morpho-syntactic and (mor)phonological features that we considered. However, t-SNE visualizations sometimes reveal clear clusters, such as for the shift vectors discussed in Section~\ref{sec:shift_vectors}, and for the majority of morpho-syntactic properties in Sections~\ref{sec:analysis_POS} through~\ref{sec:analysis_size}, but also may not show any clear clustering, as observed for cluster size in Section~\ref{sec:analysis_size}.  In what follows, we address the question of why this divergence between supervised learning with cross-validation, and unsupervised learning, is possible.  Following up on \cite{Stupak-baayen-2022-particle-verbs}, we argue that, for clusters to emerge in a t-SNE analysis, the features in question must differentiate along latent dimensions of the semantic space that account for substantial variance. By contrast, features that do not explain much of the variance in semantic space are invisible to the t-SNE algorithm.  To support this interpretation, we used principal components analysis (henceforth PCA, \cite{Pearson-1901-pca}) to rotate the original embedding space. In PCA, each component captures a certain portion of the variance in the data. The first dimension of the rotated semantic space captures most of the variance, the second dimension captures the next largest amount of variance, and so on.  We can now ask whether the categories under consideration are well-represented in the most highly ranked dimensions of PCA, or whether they are thinly and evenly spread across multiple dimensions. 

\begin{figure}
\centering
\includegraphics[width=0.8\textwidth]{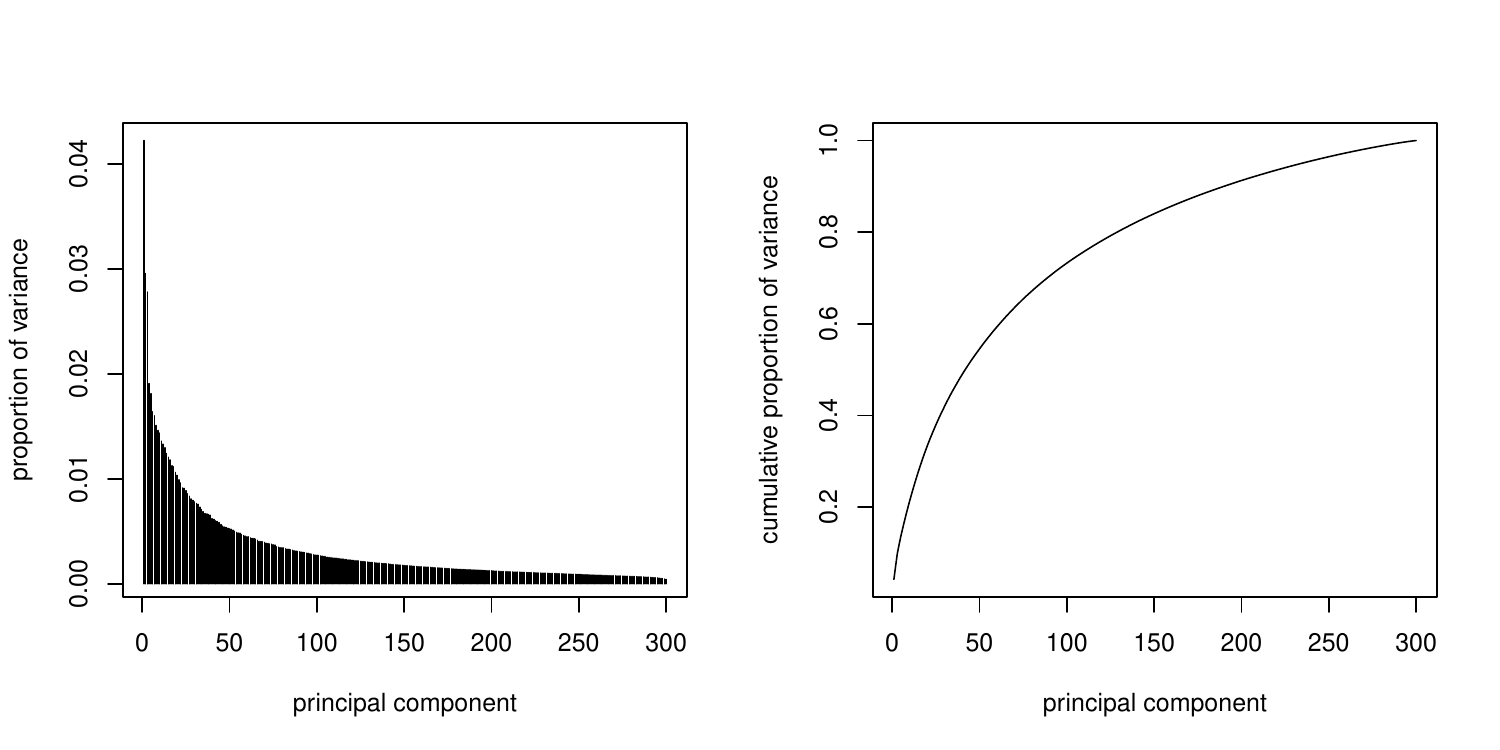}
\caption{The representation of the spread of the data points in principal components. The left panel shows the proportion of variance as explained by components PC=1 through PC=300. The right panel shows how the total variance explained increases as more components are added.}
\label{fig:ldapca}
\end{figure}

Figure~\ref{fig:ldapca} juxtaposes the proportion of variance as explained by the individual 300 principal components (PCs), and the cumulative proportion of variance captured by these components.  The left panel clarifies that even the highest ranked principal components capture a relatively small part of the variance and that -- conversely -- high-ranked dimensions still capture some variance. Nevertheless, as shown in the right-hand panel, only the first 50 components jointly capture as much as 55\% of the variance, and the first 100 principal components capture 73\% of the variance.  If a feature can be classified with high accuracy by a linear discriminant analysis on the basis of only the first 50 PCs, this implies that this feature is well-represented on these components, and is entangled with substantial variance in semantic space.  Conversely, if a feature can only be predicted well when the LDA is given to all 300 dimensions, the implication is that this feature is not explaining large amounts of variance in semantic space but is more subtly represented by all latent dimensions jointly.  

Figure~\ref{fig:ldapca_all} below presents the classification accuracy of linear discriminant analyses (with leave-one-out cross-validation) as a function of the number of principal components available to the LDA for selected morpho-syntactic and (mor)phonological features. 

\begin{figure}
\centering
\includegraphics[width=0.8\textwidth]{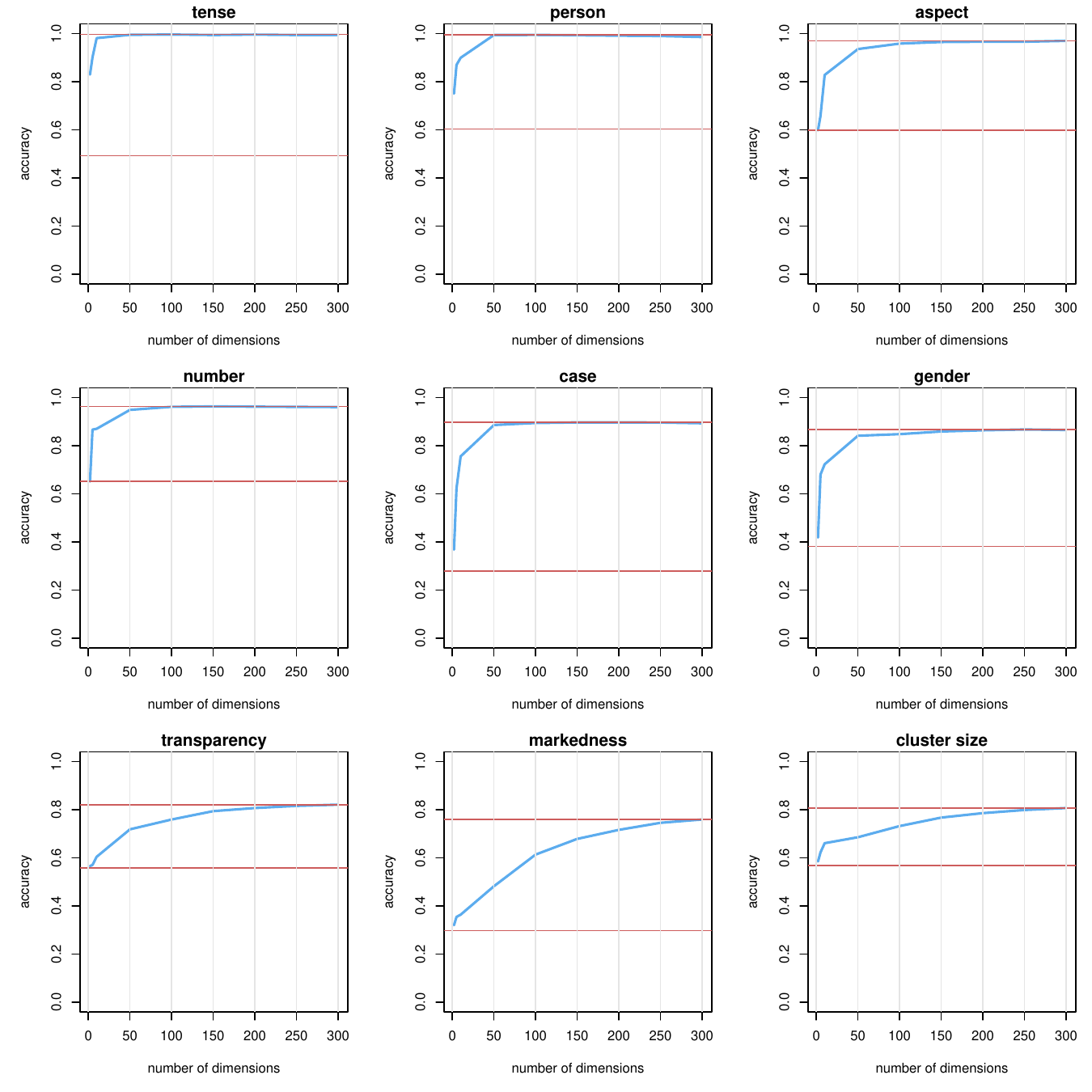}
\caption{Accuracy of linear discriminant analyses applied to the first n = 2, 5, 10, 50, 100, 150, 200, 250, 300 dimensions of PCA semantic vectors.}
\label{fig:ldapca_all}
\end{figure}

As can be observed, the semantics of person, number, and case are captured to a large extent by the first 50 dimensions. For gender and aspect, the first 50 PCs provide high-quality information, but the first 100 components are required to get very close to the best possible classification accuracy. Remarkably, for the category of tense, the first 10 PCs basically suffice in (possibly) fully capturing word meaning. The (mor)phonological characteristics of words constitute more dispersed categories in that word meaning is achieved with more PCs.  For cluster size and transparency, at least 200 dimensions are needed to approximate meaning. For markedness, at least 250 dimensions are required to obtain a good classification accuracy. 

\begin{figure}
\centering
\includegraphics[width=0.8\textwidth]{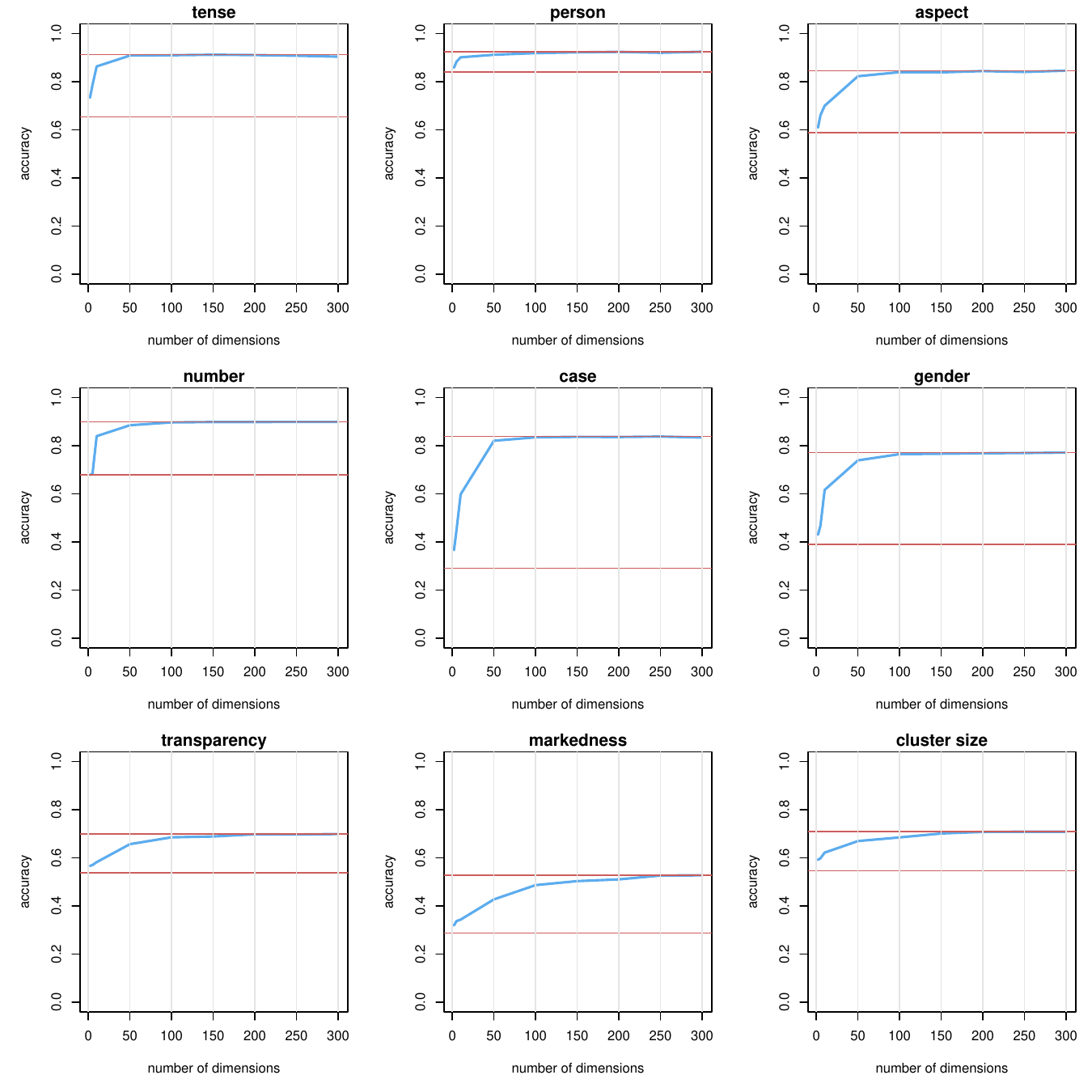}
\caption{Accuracy of linear discriminant analyses for \textit{Word2vec} applied to the first n = 2, 5, 10, 50, 100, 150, 200, 250, 300 dimensions of PCA semantic vectors.}
\label{fig:ldapca_all_2vec}
\end{figure}

\begin{figure}[H]
\centering
\includegraphics[width=0.8\textwidth]{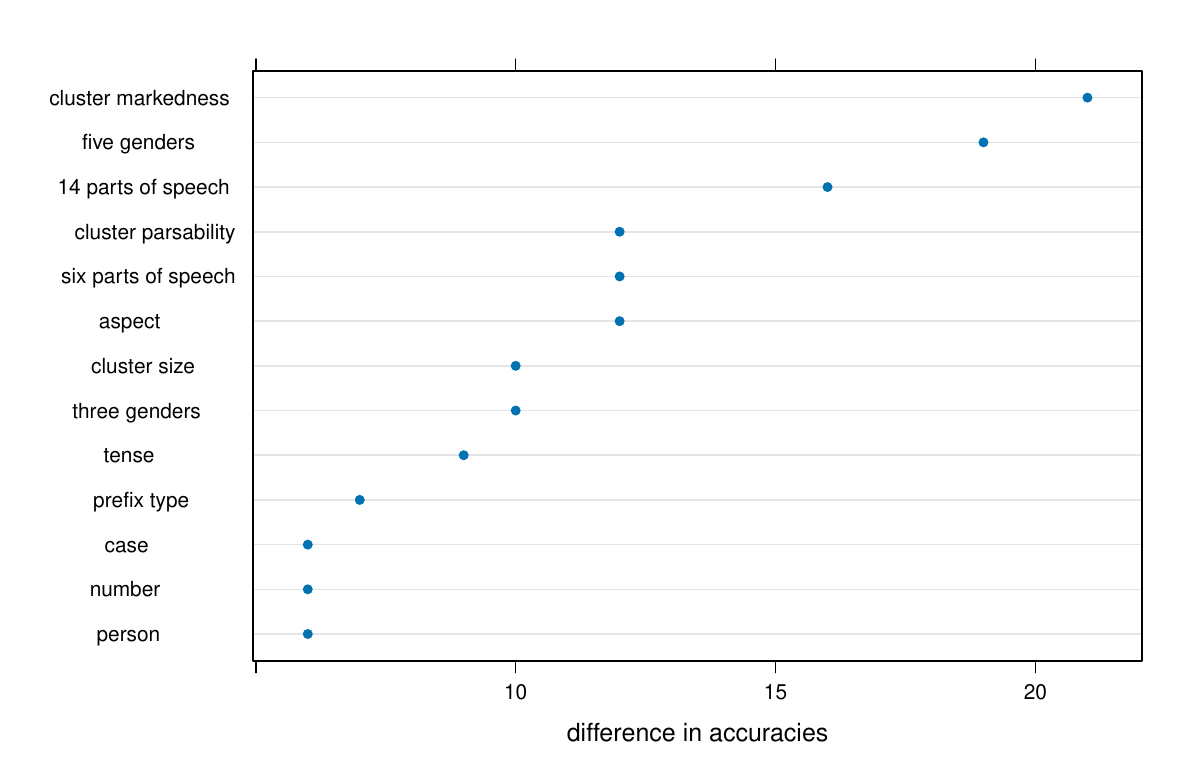}
\caption{\label{fig:Fasttext_vs_2vec} Difference in accuracy between LDA accuracy using  \textit{fastText} and \textit{Word2vec}. This difference quantifies the advantage of having sublexical co-occurrence information over and above lexical co-occurrence information.}
\end{figure}

At this point, the question arises to what extent the predictivity of the \textit{fastText} embeddings crucially hinge on \textit{fastText} having access to subword n-grams. To address this question, we present the results for \textit{Word2vec}, using otherwise the same procedure. Figure \ref{fig:ldapca_all_2vec} shows that the main trends are largely comparable to those obtained with \textit{fastText} vectors.  Fewer dimensions are required for case, number, gender, tense, aspect, and person, as compared to cluster size, markedness, and transparency, replicating the results obtained with \textit{fastText} embeddings.  However, accuracies with \textit{Word2vec} are lower overall, which aligns with previous findings (e.g. \cite{Chuang-etal-2023-defective-nouns-rus, Nikolaev-etal-2022-nouns-fin}).  What our analyses using \textit{Word2vec} clarify is that considerable structure in semantic space is detectable even when embeddings are used that do not `look inside words' as do \textit{fastText} vectors.  In other words, `looking inside words' is not critical for the conceptualization of word meaning. However, as can be seen in Figure~\ref{fig:Fasttext_vs_2vec}, the advantage of \textit{fastText} over \textit{Word2vec} is greatest for cluster markedness, and fine-grained classifications of gender and part of speech, indicating that for these categories, subword co-occurrence statistics are the most important for accurate classification.

\begin{figure}
\centering
\includegraphics[width=0.8\textwidth]{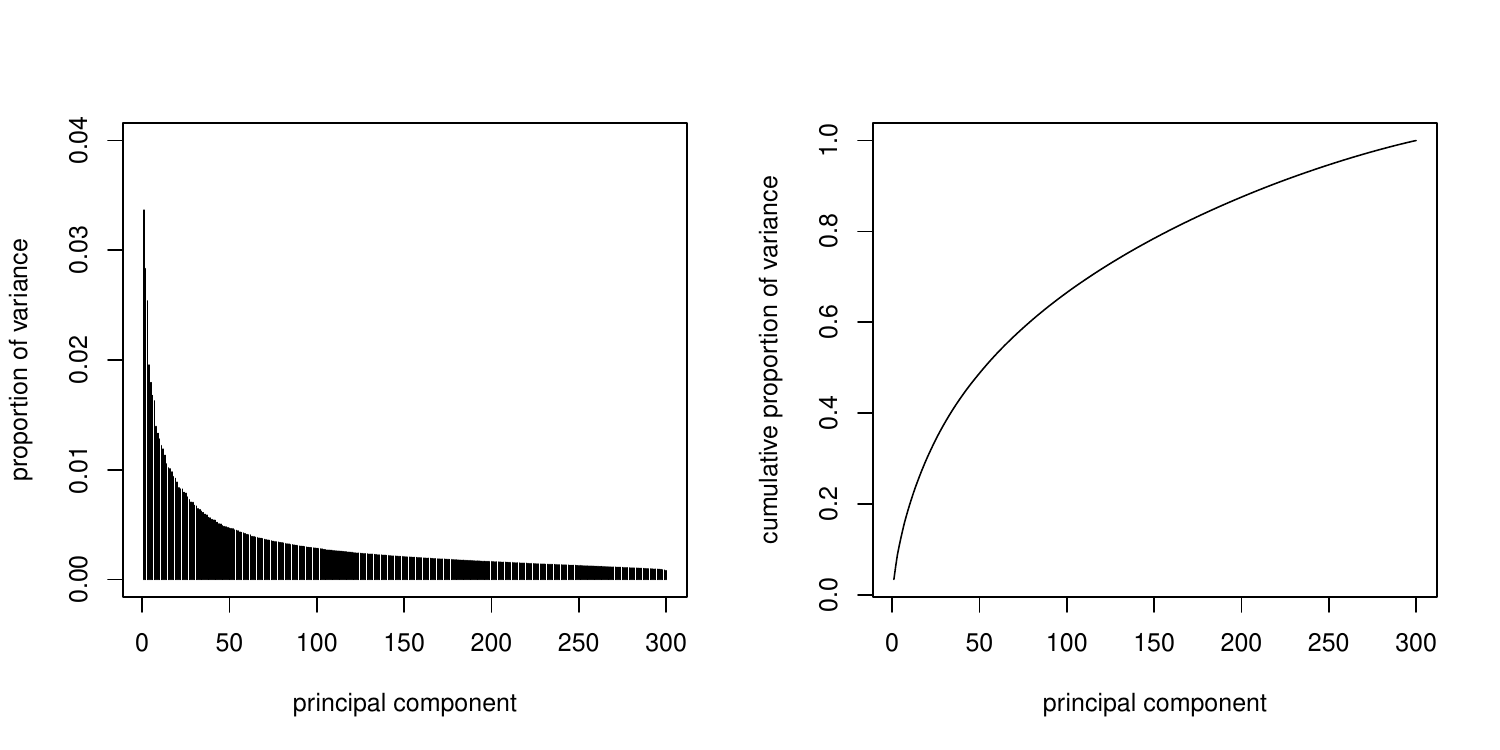}
\caption{The representation of the spread of the data points in principal components for \textit{Word2vec}. The left panel shows the proportion of variance as explained by components PC=1 through PC=300. The right panel shows how the total variance explained increases as more components are added.}
\label{fig:screeplot_all_2vec}
\end{figure}

\section{General Discussion}\label{sec:gendisc}

In this study, we first laid out the complexities of Polish phonotactics and morphology. This language has been characterized as having ‘enormous complexity of consonantal sequences’ \citep{Rochon-2000-OT} (p.1) and a ‘formidable array’ of ‘odd-looking’ \citep{Cyran-gussmann-1999-clusters-GP}(p.219f.) and ‘unusual phonotactic possibilities’ \citep{Rubach-booij-1990-syllable-pl} (p.454), many of which are triggered by morphology \citep{Dressler-KDK-2006}. 

We then showed that a cognitively motivated computational model of the mental lexicon, the Discriminative Lexicon Model \citep{Heitmeneier-etal-book}, understands and produces Polish words with high accuracy. The central question that we addressed is why this is possible, as mathematically, the model is very simple, using straightforward linear mappings, while being constrained to work with numerical representations of words' forms and words' meanings (using embeddings). Our general hypothesis has been that the form space and the semantic space of Polish must be aligned to a considerable extent. In other words, the high accuracy of the model predictions suggested that the morpho-syntactic and (mor)phonological properties of Polish words have mirror images on the semantic side.  

To follow up on this possibility, we conducted a series of computational experiments in which we employed Linear Discriminant Analysis to clarify to what extent (mor)phonotactic and morphosyntactic features can be predicted from word embeddings. We reported that the accuracy ranges from 81\% (for phonotactic features) to 99\% (for morphosyntactic features). Prediction accuracies consistently substantially exceeded majority baselines. Accuracies were higher when \textit{fastText} embeddings were used compared to \textit{Word2vec} embeddings. But the accuracies obtained with \textit{Word2vec} were also far above the majority baselines. Apparently, it is possible to predict the internal structure of Polish words from their semantics, based solely on simple word co-occurrence patterns captured by \textit{Word2vec} embeddings. \textit{FastText} vectors offer a further advantage by incorporating subword information, which provides an additional layer of informativity. This subword sensitivity gives \textit{fastText} an extra leverage in capturing morphosyntactic, morphonological, and phonological regularities. 

We also observed that for features such as person, tense, and aspect, high accuracy is obtained with less than 100 principal components of the semantic space, whereas for transparency, markedness, and cluster size, 200 to 300 dimensions are essential to reach maximum prediction accuracy.  This clarifies that features such as, among others, person, tense, and aspect are responsible for substantial variation in the semantic space, in line with hypothesis 4, whereas (mor)phonological features such as cluster size, markedness and parsability are reflected much more subtly and evenly across all dimensions of the semantic space. But even phonotactic and morphonotactic features are still highly discriminable, as anticipated by hypotheses 2 and 3.  

The present paper adds to a relatively small pool of studies showing a closely-knit relationship between meaning and form.  The first and --- as far as we know --- the only study to address the relationship between semantics and phonotactics for Polish is \cite{KDK-etal-2011-lexicon}. The authors made an attempt at classifying the lexicon based on word-medial combinations of consonants, and showed that specific consonant clusters can map derivational categories, inflectional categories, parts of speech and morphotactic possibilities in the language. For instance, /-jpfr-/ is found exclusively in the derivational nest of \textit{grapefruit}, /-dvst-/ maps the inflectional paradigm of \textit{przedwstępny} `preliminary', and /-nt\textipa{\textesh}r-/ occurs exclusively in adjectives with one prefix (e.g. \textit{wewnątrz+redakcyjna} `internal to the editorial team', \textit{wewnątrz+resortowa} `inter-departmental').  Finally, the model assumes that a morpheme boundary may appear within a consonant cluster that maps a subset of words of a specific morphological composition, such as /-zvzgl-/, which is found in adverbs and adjectives featuring the prefix \textit{bez-} `without' and the basic form \textit{wzgląd} `consideration'. The overarching tendency that according to \citet{KDK-etal-2011-lexicon} characterizes the whole lexicon is that an increase in cluster size is more likely to be associated with one or more semantic categories. These conclusions dovetail well with the findings reported in Section~\ref{sec:analysis_size} on the predictability of cluster types within a cluster.

The form – meaning parallels reported in the paper is consistent with a line of research that argues in favor of a more tightly-knit relationship between semantics of words and their phonetic and phonological properties. Under this view, even simple coursebook examples illustrate how the two modules of grammar interact. A minute change in the phonological shape of a word affects its meaning. For example, a minimal change in feature [±voice] in \textit{tip - dip} leads to entirely different semantic interpretation of the words, including a change in a part of speech. The reverse influence has also been explored. A battery of studies have provided evidence in favour of the semantic support for allophony and phonological contrast. \cite{Peperkamp-dupoux-2007-zipfian-paradigm} studied the allophonic variation of consonants in two related forms that followed different determiners and found that such alternations are learned based on both phonological and semantic cues. That is, speakers used a specific determiner as a cue to predict allophones. Other studies have investigated the role of semantic properties in the acquisition of non-native phonological contrast. \cite{Yeung-werker-2009-learning} have found that 9-month-old infants learn to recognize the contrast between a dental and retroflex stop /d/ when the signal is accompanied by specific visual cues. In \cite{Creel-2012-phonological-similarity}, children were shown to use semantic cues to deal with odd pronunciations. For example, pre-schoolers select known referents such as ‘fish’ when searching for a referent for odd pronunciations such as \textit{feesh}. In a similar fashion, \cite{Swingley-aslin-2000-word-recognition} have found that infants, when hearing an odd phonetic variant such as /ve\textbackslash{}textipa\{\i\}bi/, tend to look at a known referent, namely \textit{baby}. Other studies on the phonology-semantics interface have explored homophony (e.g. \cite{Kaplan-2011-homophony, Kaplan-2015-homophony-avoidance, Kaplan-muratani-2015-homophony-avoidance}, or sound symbolism (e.g. \cite{Hinton-ohala-2006-sound, Wrembel-2010-symbol}. 

The final aspect is particularly interesting as the results contribute to our understanding of the symbolic nature of sounds by providing evidence that the relationship between linguistic form and meaning is less arbitrary than traditionally argued (e.g. Hinton et al., 2006; Dingemanse et al., 2015; Winter et al., 2024). Our results show that the phonotactic properties of Polish words, such as cluster size, markedness, and morphonotactic transparency, are partially reflected in semantic space. Although these correspondences are weaker than those observed for morphosyntactic categories, their consistent presence suggests that sublexical phonological structure contributes to meaning representation. This is particularly noteworthy in Polish, where highly complex consonant clusters might be expected to be semantically opaque. Instead, the findings support emerging evidence that form–meaning mappings can arise as gradient, distributed patterns across the lexicon, rather than being limited to isolated cases of iconicity (cf. Dziubalska-Kołaczyk et al., 2011). Overall, the results challenge strong versions of the arbitrariness assumption and point toward a more integrated view of phonology and semantics in the mental lexicon.

As to the accuracy of computational models, we have shown that the (mor)phonotactic and morphosyntactic properties of a large set of morphologically complex Polish words can be predicted with high accuracy not only from \textit{fastText} embeddings, but also from \textit{Word2vec} embeddings.  We have also demonstrated that the Discriminative Lexicon Model, implemented with simple linear mappings, achieves high prediction accuracies for form and meaning. We propose that these high accuracies would be impossible without the phonological and morphological reflexes in the semantic space of Polish, on the one hand, and the intricate and diverse forms with which meaning is realized in form, on the other hand. Apparently, meaning and form are much more finely calibrated than has often been assumed.

\appendix

\section{Results for \textit{Word2vec}}\label{sec:word2vec}

\begin{table}[H]
\centering
\begin{tabular}{ |c||c|c|c|} 
 \hline
                & full data & training data & test data\\\hline
   \hline
comprehension   & 77        & 79          & 75.6 \\\hline
production      & 85.6      & 87          & 53 \\\hline
\end{tabular}
\caption{\label{tab:LDL_accuracy_2vec} Accuracy (in percentages) for comprehension and production for the full dataset, a training dataset and a held-out test data set for \textit{Word2vec} embeddings.}
\end{table}

\begin{figure}[H]
\centering
\includegraphics[width=0.6\textwidth]{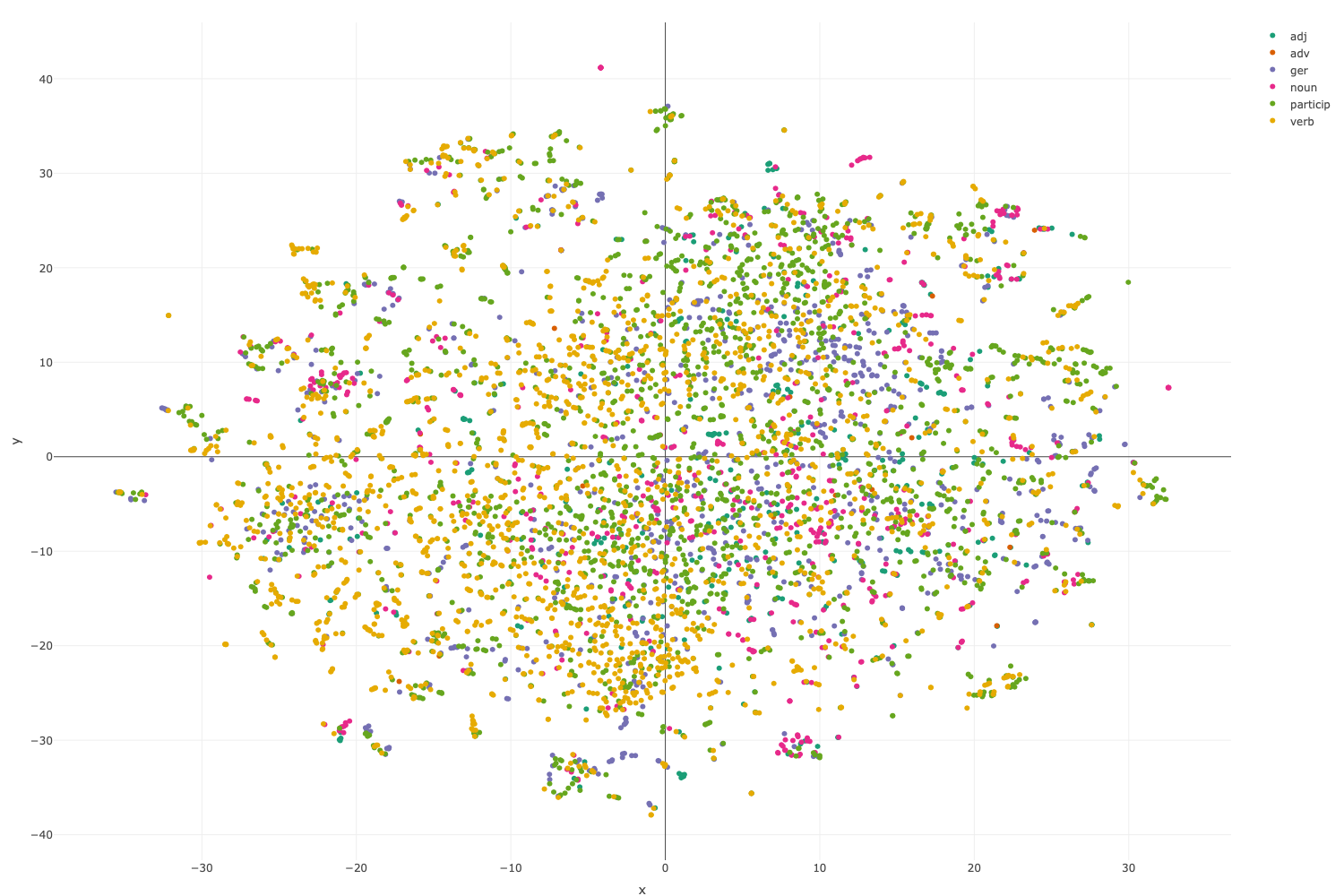}
\caption{\label{fig:POS_tsne_2vec}Scatterplot of words in a 2-dimensional t-SNE space, colour-coded for six parts of speech.}
\end{figure}

\begin{table}[H]
\centering
\begin{tabular}{ |c||c|c|c|c|c|c| } 
 \hline
            &adjective  &adverb   &gerund   &noun       &participle &verb \\\hline
  \hline
adjective   &230 (58\%) &0       &14        &27         &110        &17  \\\hline
adverb      &1         &7 (47\%) &0         &2          &5          &0   \\\hline
gerund      &3         &0       &693 (79\%) &63         &108        &11  \\\hline
noun        &32        &2       &70         &406 (66\%) &101        &6   \\\hline
participle  &62        &0       &33         &38         &1810 (89\%) &93  \\\hline
verb        &4         &0       &3          &6          &152        &1827 (92\%)\\\hline
\end{tabular}
\caption{\label{tab:POS_2vec}A confusion matrix for six parts of speech; model accuracy = \SI{84}{\percent}.}
\end{table}

\begin{figure}[H]
\centering
\includegraphics[width=0.6\textwidth]{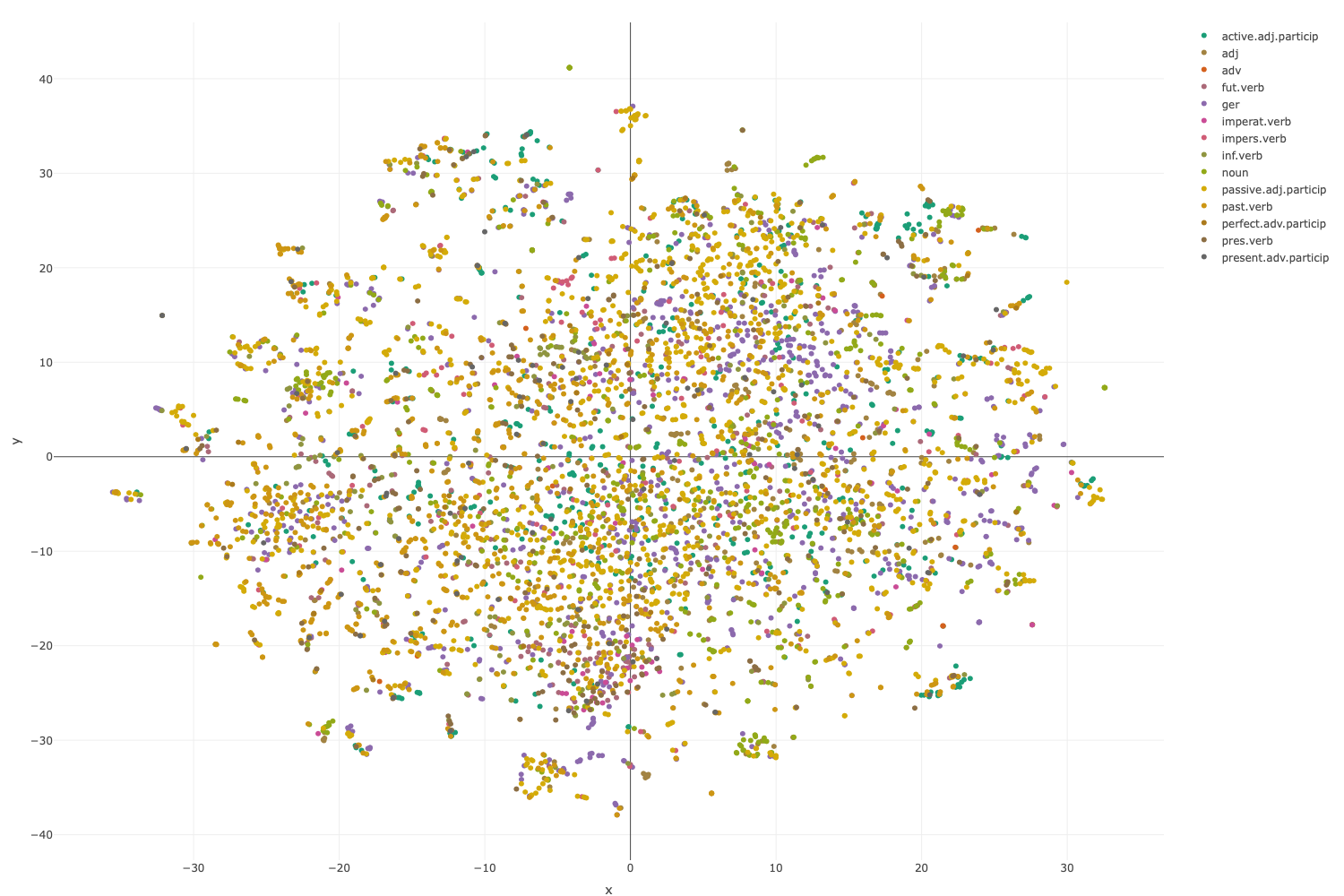}
\caption{\label{fig:POS_detailed_tsne_2vec}Scatterplot of words in a 2-dimensional t-SNE space, colour-coded for 14 parts of speech.}
\end{figure}

\begin{landscape}
\begin{table}[H]
\centering
\begin{tabular}{ |c||c|c|c|c|c|c|c|c|c|c|c|c|c|c| } 
 \hline
                    &active.adj.particip &adj  &adv &fut.verb  &ger  &imperat.verb  &impers.verb &inf.verb &noun   &passive.adj.particip &past.verb &perf.adv.particip &pres.verb &pres.adv.particip \\\hline
  \hline
active.adj.particip & 353 (68\%) &23  &0  &1   &25  &2   &0    &0   &20   &85  &5     &0   &1     &4 \\\hline
adj                 & 32   &247 (62\%) &0  &1   &13  &4   &0    &0   &33   &61  &4     &2   &1     &0 \\\hline
adv                 &3     &1   &7 (47\%) &0   &0   &0   &0    &0   &2    &2   &0     &0   &0     &0 \\\hline
fut.verb            &4     &0   &0  &193 (76\%) &3   &24  &0    &1   &0    &10  &13    &0   &6     &0 \\\hline
ger                 &22    &7   &0  &1   &728 (83\%) &3   &0    &0   &73   &41  &2     &1   &0     &0 \\\hline
imperat.verb        &6     &1   &0  &8   &2   &64 (70\%) &0    &1   &3    &6   &0     &0   &0     &0 \\\hline
impers.verb         &3     &0   &0  &1   &1   &1   &104 (67\%) &0   &0    &14  &31    &0   &0     &0 \\\hline
inf.verb            &3     &0   &0  &11  &2   &0   &0    &207 (86\%) &0    &14  &4     &0   &0     &0 \\\hline
noun                &25    &36  &2  &1   &76  &1   &0    &0   &417 (68\%) &57  &0     &1   &1     &0 \\\hline
passive.adj.particip &52   &42  &0  &6   &19  &5   &2    &0   &26  &1223 (88\%) &7     &6   &0     &1 \\\hline
past.verb           &20    &2   &0  &10  &1   &2   &8    &0   &2    &43  &888 (90\%)  &7   &2     &0 \\\hline
perf.adv.particip   &2     &0   &0  &1   &1   &0   &0    &0   &0    &5   &8     &22 (54\%) &0     &2 \\\hline
pres.verb           &16    &1   &0  &33  &3   &19  &1    &0   &1    &11  &18    &0   &163 (61\%)  &0 \\\hline
pres.adv.particip   &4     &0   &0  &1   &2   &0   &0    &0   &0    &6   &11    &6   &2     &55 (63\%) \\\hline
\end{tabular}
\caption{\label{tab:POS_detailed_2vec}A confusion matrix for 14 parts of speech; model accuracy = \SI{79}{\percent}.}
\end{table}
\end{landscape}

\begin{figure}[H]
\centering
\includegraphics[width=0.6\textwidth]{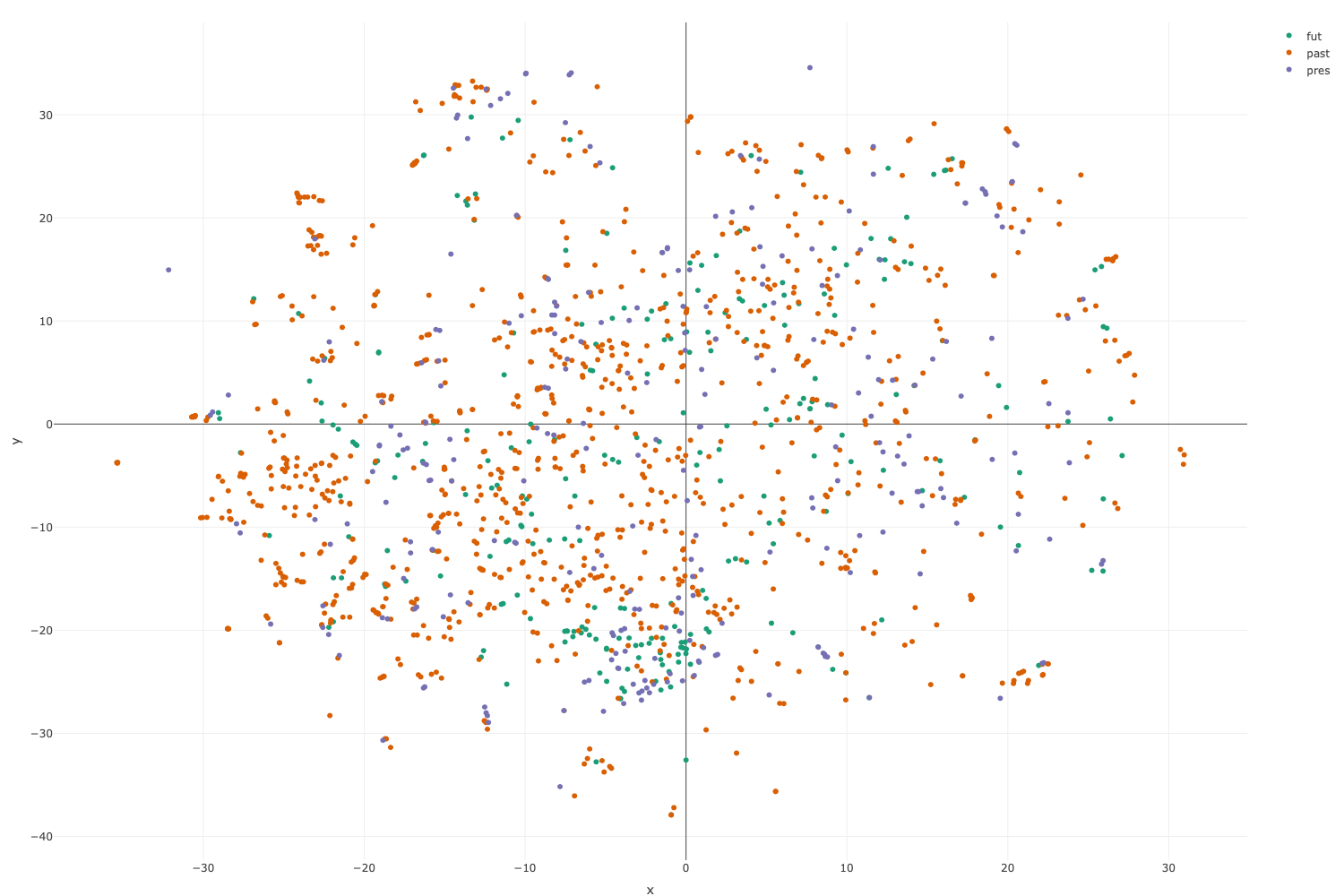}
\caption{\label{fig:Tense_tsne_2vec}Scatterplot of words in a 2-dimensional t-SNE space, colour-coded for three tenses.}
\end{figure}

\begin{table}[H]
\centering
\begin{tabular}{ |c||c|c|c| } 
 \hline
            &future     &past           &present        \\\hline
   \hline
future      &214 (84\%) &28             &12             \\\hline
past        &16         &958 (97\%)     &11             \\\hline
present     &46         &32             &188 (71\%)     \\\hline
\end{tabular}
\caption{\label{tab:tense_2vec}A confusion matrix for tense; model accuracy = \SI{90}{\percent}.}
\end{table}

\begin{figure}[H]
\centering
\includegraphics[width=0.6\textwidth]{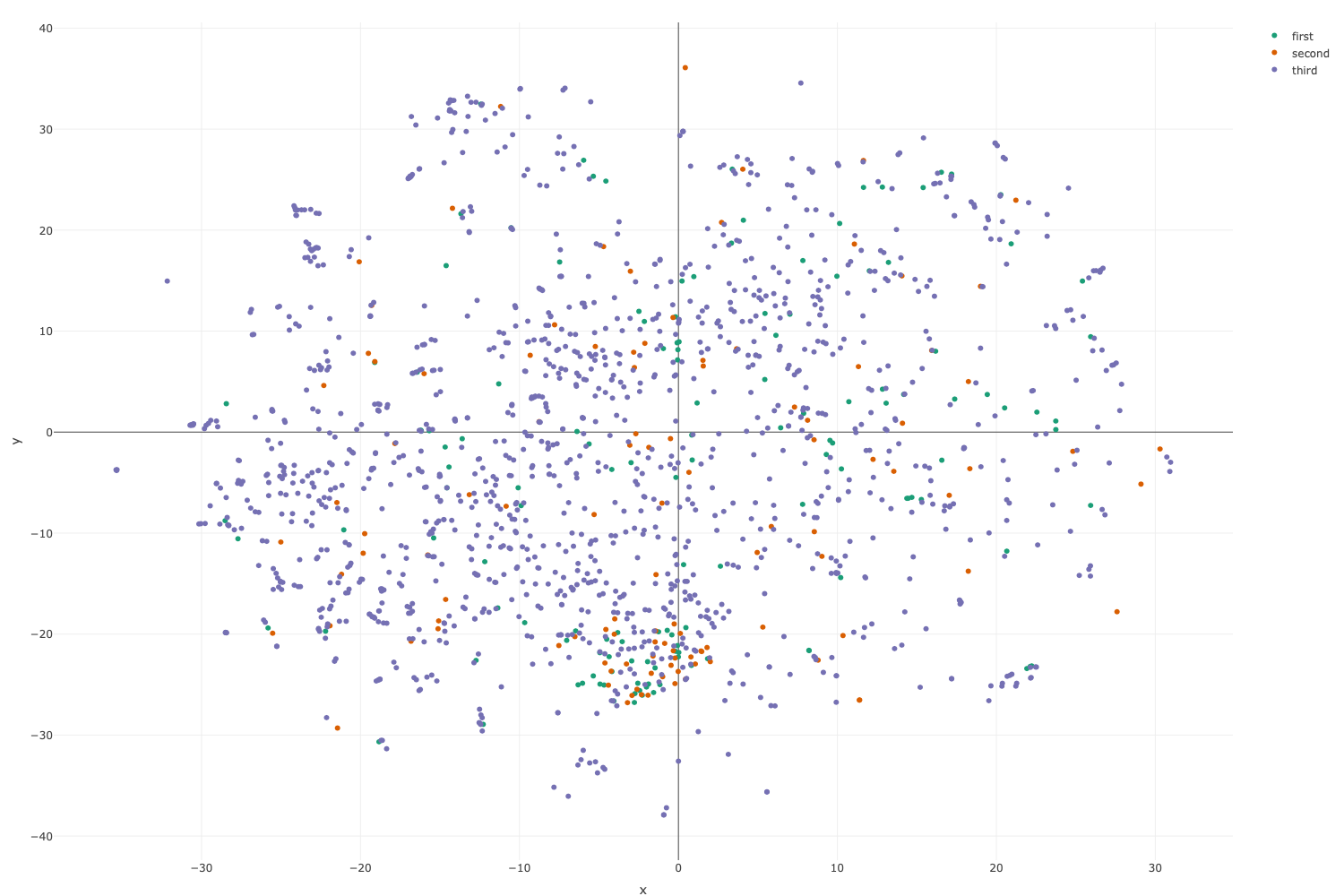}
\caption{\label{fig:Person_tsne_2vec}Scatterplot of words in a 2-dimensional t-SNE space, colour-coded for three persons.}
\end{figure}

\begin{table}[H]
\centering
\begin{tabular}{ |c||c|c|c| } 
 \hline
            &first      &second     &third          \\\hline
   \hline
first       &95 (67\%)  &18         &20             \\\hline
second      &31         &71 (63\%)  &10             \\\hline
third       &25         &7          &1311 (98\%)   \\\hline
\end{tabular}
\caption{\label{tab:person_2vec}A confusion matrix for person; model accuracy = \SI{93}{\percent}.}
\end{table}

\begin{figure}[H]
\centering
\includegraphics[width=0.6\textwidth]{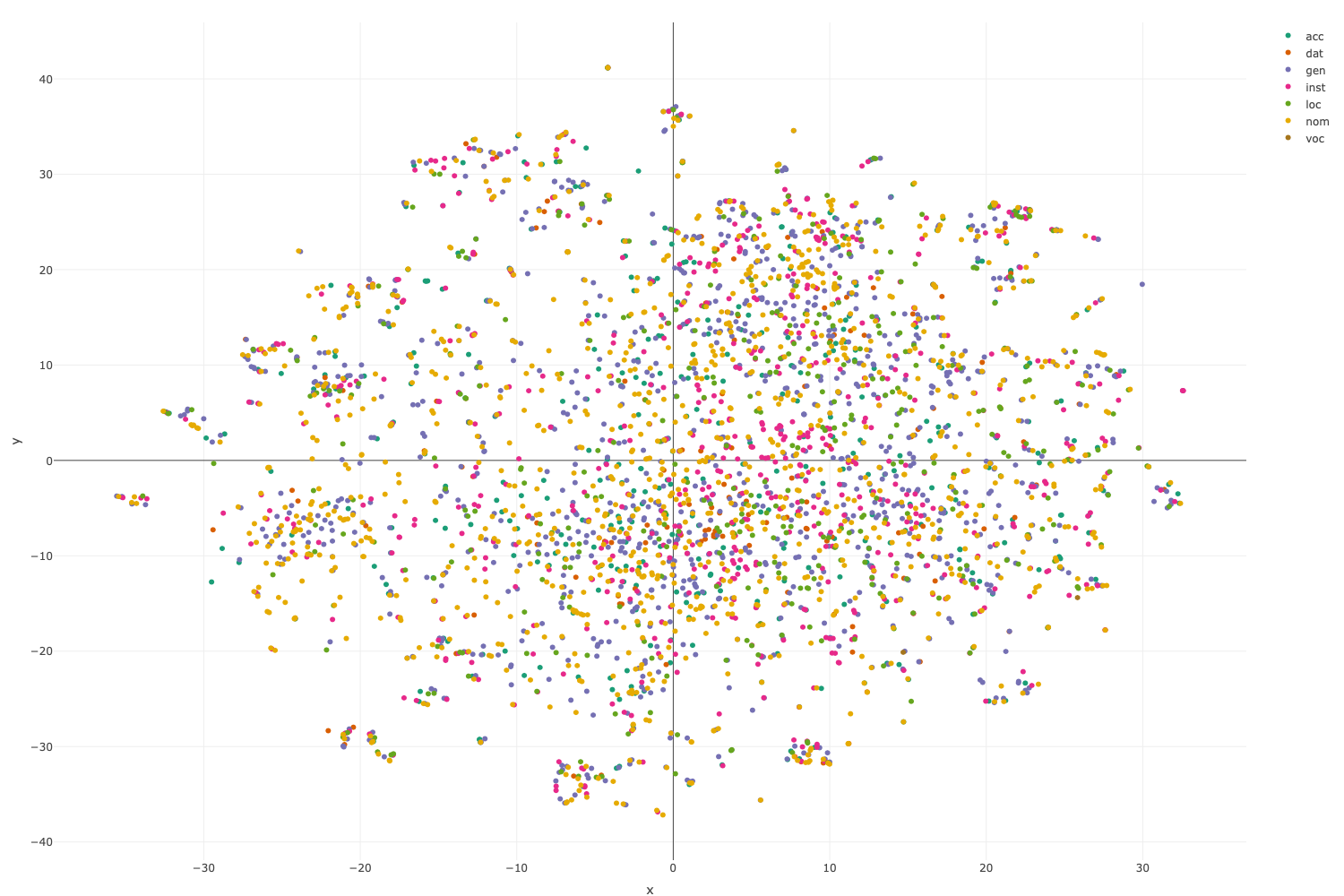}
\caption{\label{fig:Case_tsne_2vec}Scatterplot of words in a 2-dimensional t-SNE space, colour-coded for seven cases.}
\end{figure}

\begin{landscape}
\begin{table}[H]
\centering
\begin{tabular}{ |c||c|c|c|c|c|c|c| } 
 \hline
        &accusative &dative &genitive &instrumental &locative &nominative &vocative\\\hline
  \hline
accusative      &255 (62\%)   &0      &28       &7         &2         &117       &1       \\\hline
dative          &2     &119 (77\%)   &3        &11        &10        &9         &0       \\\hline
genitive        &16    &3      &996 (91\%)     &8         &11        &55        &0       \\\hline
instrumental    &31    &6      &13       &528 (84\%)      &22        &25        &0       \\\hline
locative        &6     &14     &35       &39        &299 (72\%)      &19        &1       \\\hline
nominative      &83    &3      &34       &15        &4         &964 (87\%)      &1       \\\hline
vocative        &0     &0      &0        &0         &0         &2 (100\%)        &0       \\\hline
\end{tabular}
\caption{\label{tab:case_2vec}A confusion matrix for case; model accuracy = \SI{83}{\percent}.}
\end{table}
\end{landscape}

\begin{figure}[H]
\centering
\includegraphics[width=0.6\textwidth]{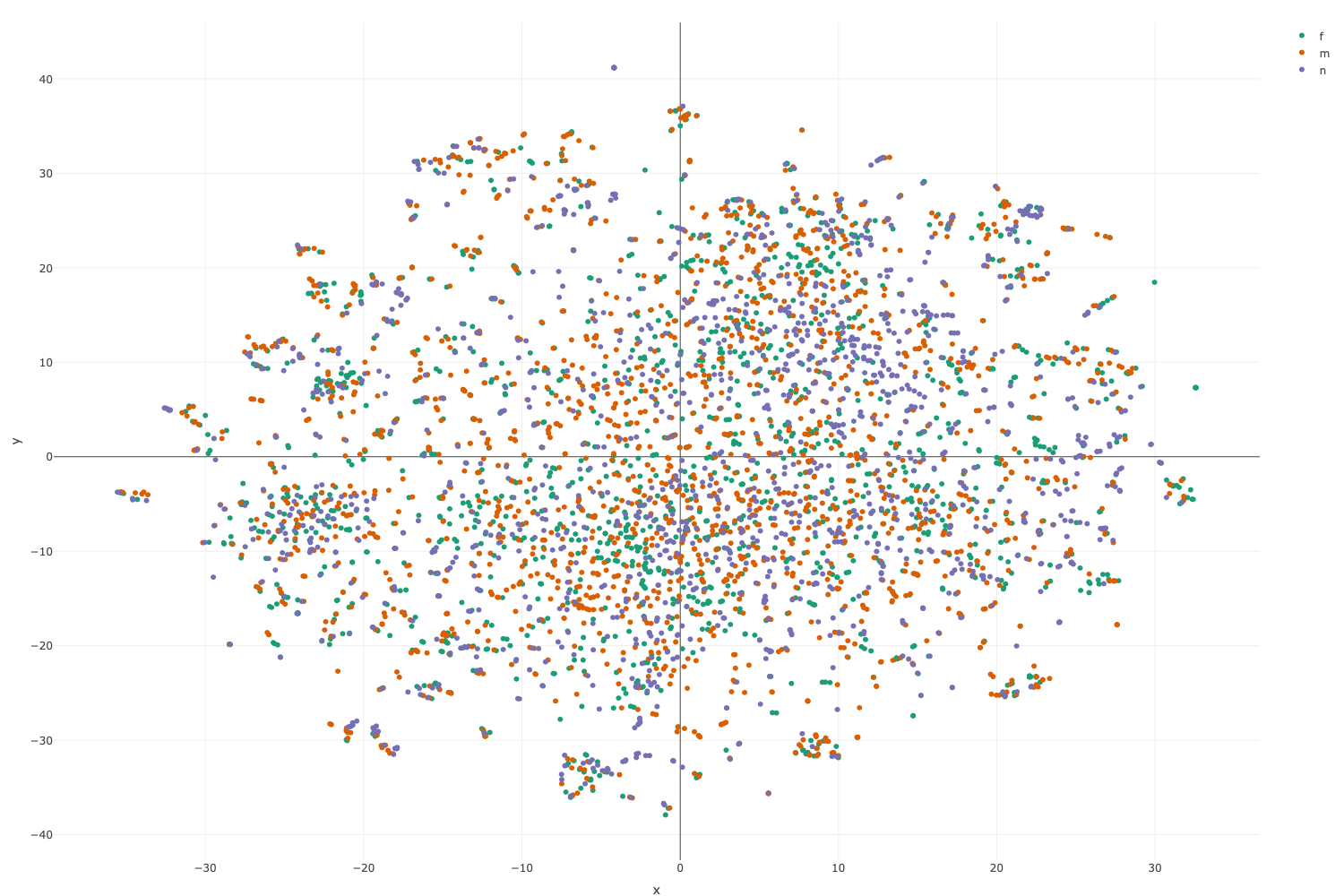}
\caption{\label{fig:Gender_tsne_2vec}Scatterplot of words in a 2-dimensional t-SNE space, colour-coded for three genders.}
\end{figure}

\begin{table}[H]
\centering
\begin{tabular}{ |c||c|c|c| } 
 \hline
            &feminine   &masculine       &neuter     \\\hline
 \hline
feminine    &1177 (77\%) &209            &148        \\\hline
masculine   &175        &1477 (79\%)     &213        \\\hline
neuter      &122        &225             &1036 (75\%) \\\hline
\end{tabular}
\caption{\label{tab:gender_gen_2vec}A confusion matrix for three genders; model accuracy = \SI{77}{\percent}.}
\end{table}

\begin{figure}[H]
\centering
\includegraphics[width=0.6\textwidth]{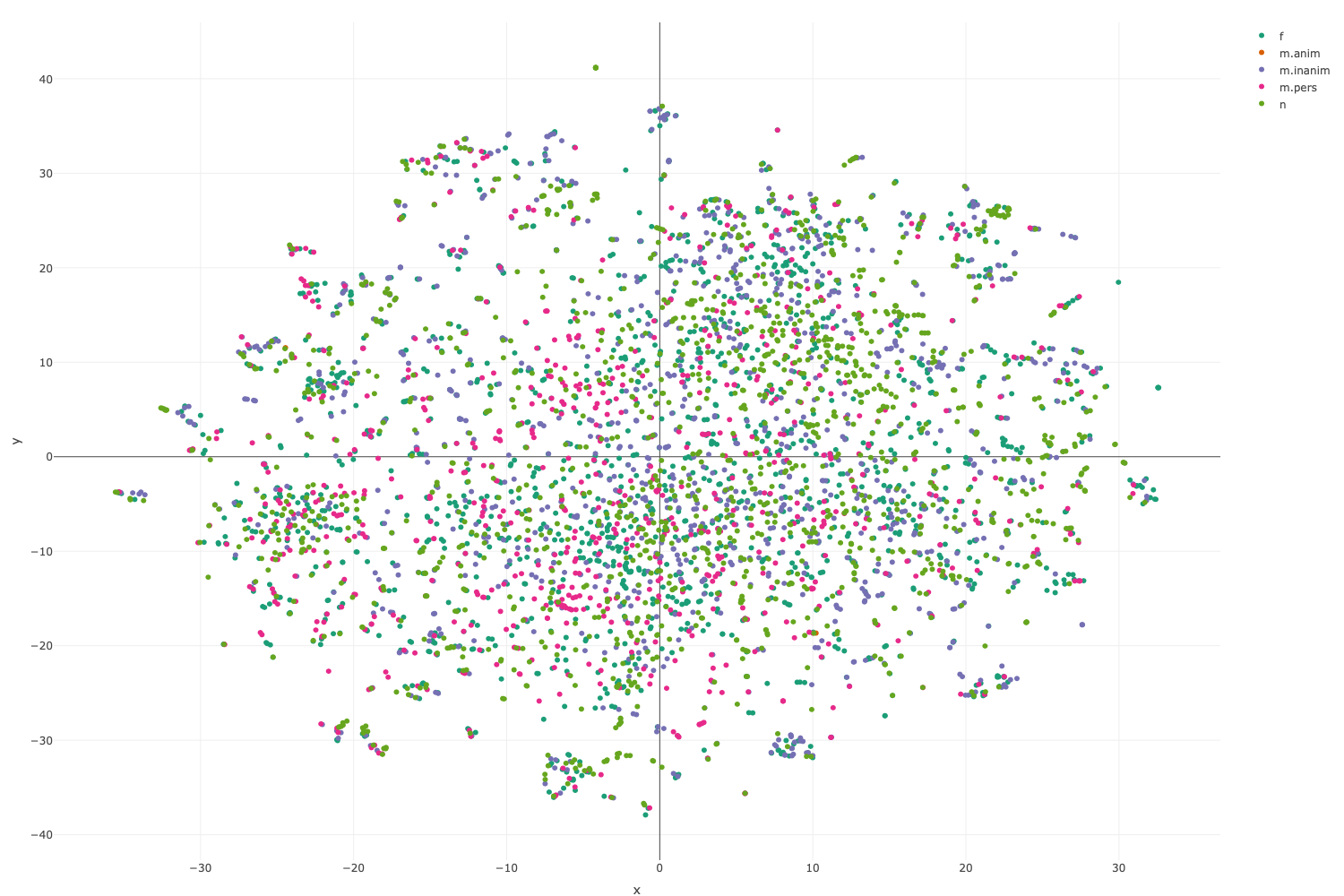}
\caption{\label{fig:Gender_spec_tsne_2vec}Scatterplot of words in a 2-dimensional t-SNE space, colour-coded for five genders.}
\end{figure}

\begin{table}[H]
\centering
\begin{tabular}{ |c||c|c|c|c|c| } 
 \hline
            &feminine   &m.animate &m.inanimate &m.personal &neuter     \\\hline
 \hline
feminine    &1206 (79\%) &0       &150          &15         &163        \\\hline
m.animate   &2 (67\%)   &0        &0            &1          &0          \\\hline
m.inanimate &157        &1        &694 (63\%)   &53         &199        \\\hline
m.personal  &47         &0        &104          &572 (75\%) &35         \\\hline
neuter      &127        &0        &169          &25         &1062 (77\%) \\\hline
\end{tabular}
\caption{\label{tab:gender_spec_2vec}A confusion matrix for five genders; model accuracy = \SI{74}{\percent}.}
\end{table}

\begin{figure}[H]
\centering
\includegraphics[width=0.6\textwidth]{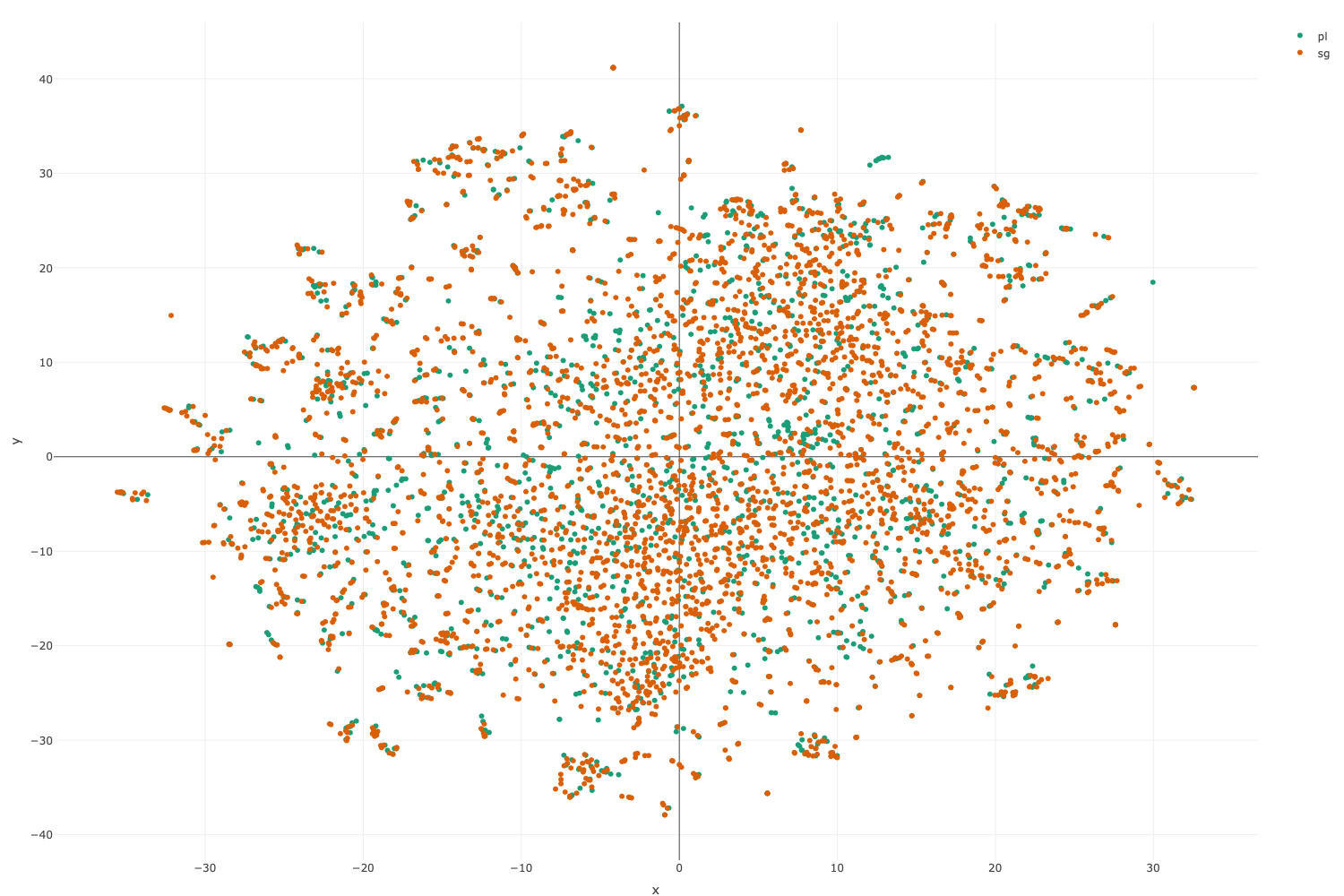}
\caption{\label{fig:Number_tsne_2vec}Scatterplot of words in a 2-dimensional t-SNE space, colour-coded for two numbers.}
\end{figure}

\begin{table}[H]
\centering
\begin{tabular}{ |c||c|c| } 
 \hline
            &singular       &plural         \\\hline
 \hline
singular    &1401 (81\%)    &330            \\\hline
plural      &215            &3447 (94\%)      \\\hline
\end{tabular}
\caption{\label{tab:number_2vec}A confusion matrix for number; model accuracy = \SI{90}{\percent}.}
\end{table}

\begin{figure}[H]
\centering
\includegraphics[width=0.6\textwidth]{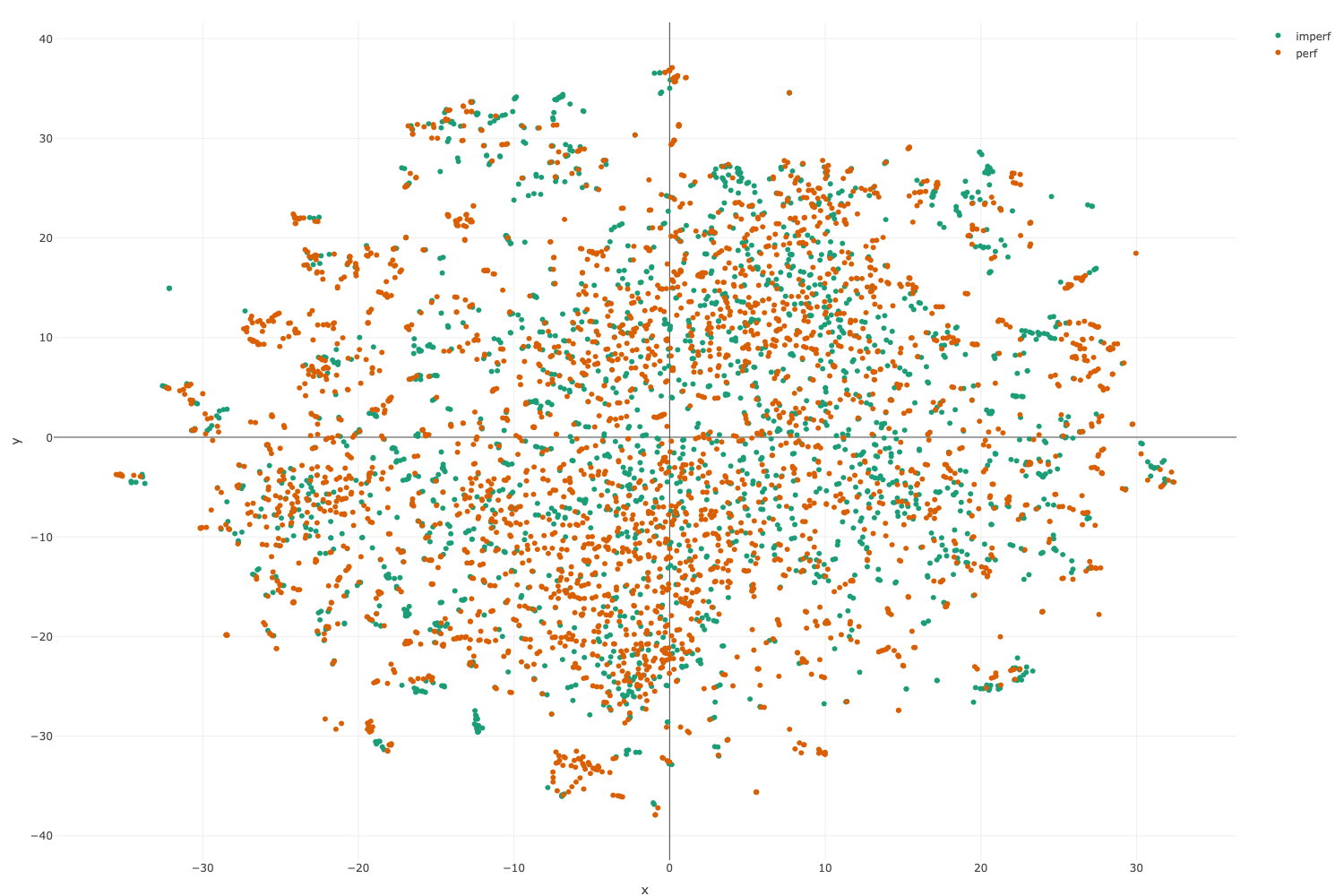}
\caption{\label{fig:Aspect_tsne_2vec}Scatterplot of words in a 2-dimensional t-SNE space, colour-coded for two aspects.}
\end{figure}

\begin{table}
\centering
\begin{tabular}{ |c||c|c| } 
 \hline
                &imperfective  &perfective    \\\hline
   \hline
imperfective    &1572 (78\%)   &445           \\\hline
perfective      &309           &2579 (89\%)   \\\hline
\end{tabular}
\caption{\label{tab:aspect_2vec}A confusion matrix for aspect; model accuracy = \SI{85}{\percent}.}
\end{table}

\begin{figure}[H]
\centering
\includegraphics[width=0.6\textwidth]{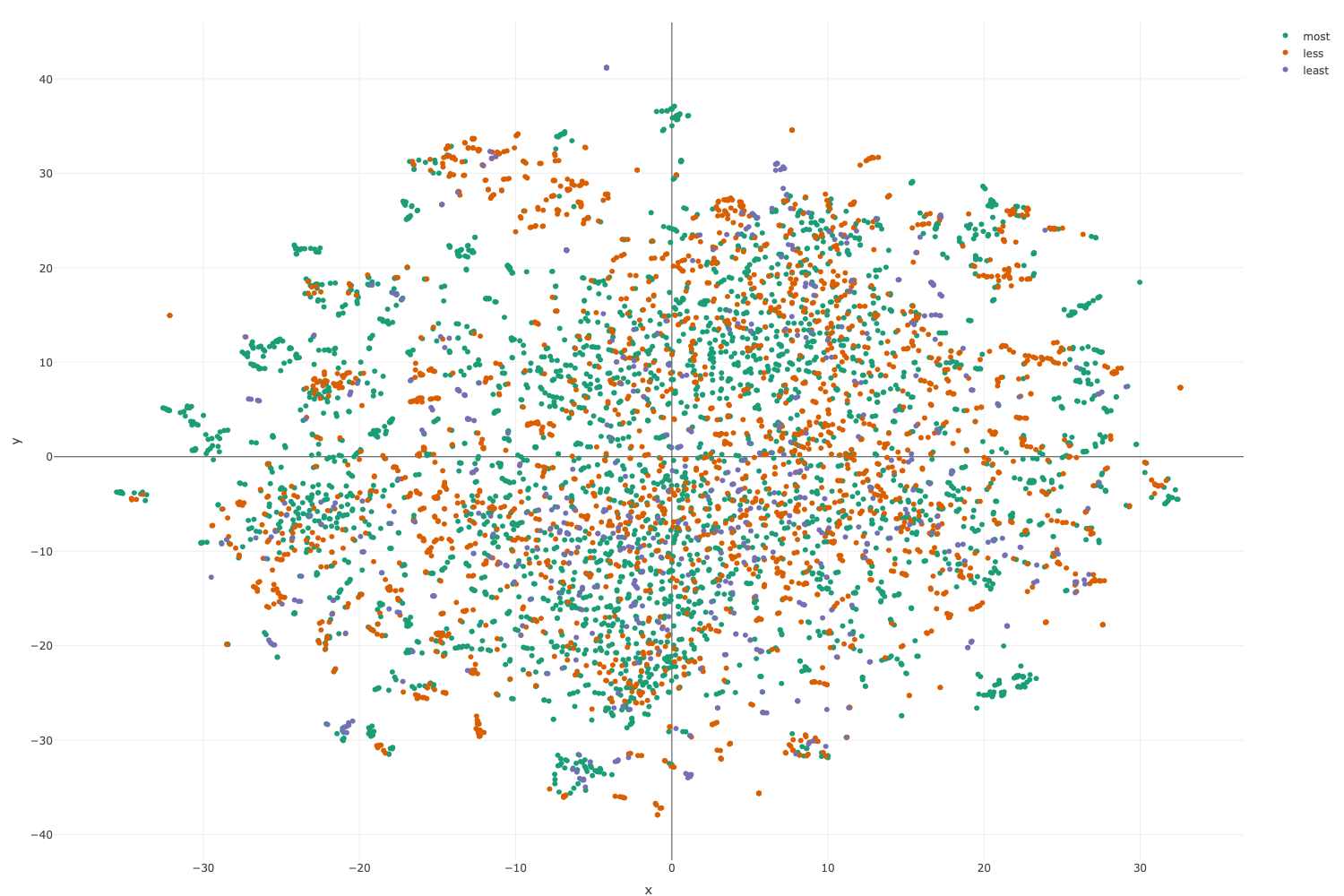}
\caption{\label{fig:Transparency_tsne_2vec}Scatterplot of words in a 2-dimensional t-SNE space, colour-coded for three degrees of morphotactic parsability.}
\end{figure}

\begin{table}[H]
\centering
\begin{tabular}{ |c||c|c|c| } 
 \hline
                &most parsable &less parsable &least parsable    \\\hline
 \hline
most parsable   &2625 (80\%)   &480           &89                \\\hline
less parsable   &796           &1314 (60\%)   &77                \\\hline
least parsable  &183           &150           &222 (40\%)        \\\hline
\end{tabular}
\caption{\label{tab:parsability_2vec}A confusion matrix for the degrees of morphotactic parsability; model accuracy = \SI{70}{\percent}.}
\end{table}

\begin{figure}[H]
\centering
\includegraphics[width=0.6\textwidth]{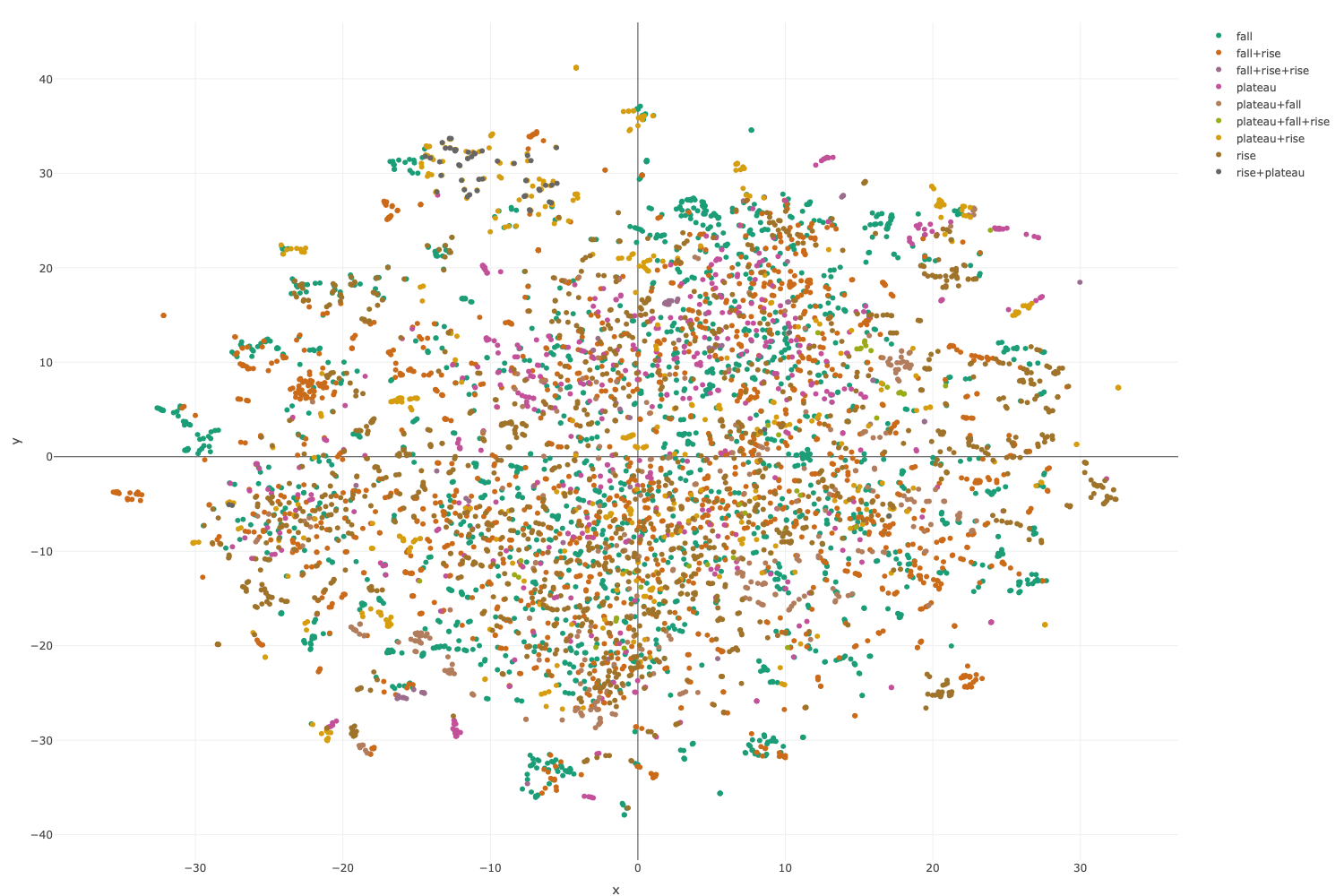}
\caption{\label{fig:Markedness_tsne_2vec}Scatterplot of words in a 2-dimensional t-SNE space, colour-coded for nine sonority profile types.}
\end{figure}

\begin{table} [H]
\centering
\begin{tabular}{ |c||c|c|c|c|c|c|c|c|c| } 
 \hline
            &F      &F+R    &F+R+R  &plat   &plat+F &plat+F+R &plat+R   &R      &R+plat\\\hline
  \hline
F           &1081 (63\%)  &305    &3      &62     &25     &5        &30       &194    &1     \\\hline
F+R         &437    &899 (59\%)   &4      &45     &28     &2        &23       &86     &0     \\\hline
F+R+R       &11     &2      &44 (77\%)    &0      &0      &0        &0        &0      &0     \\\hline
plat        &98     &65     &0      &159 (32\%)    &11     &0        &14       &143    &0     \\\hline
plat+F      &59     &36     &4      &12     &180 (55\%)   &1        &5        &29     &0     \\\hline
plat+F+R    &11     &3      &0      &0      &4      &19 (48\%)      &1        &2      &0     \\\hline
plat+R      &32     &23     &0      &11     &12     &2        &237 (50\%)     &107    &47    \\\hline
R           &87     &90     &1      &47     &24     &3        &38       &981 (77\%)   &0     \\\hline
R+plat      &0      &0      &0      &0      &0      &0        &6        &5      &40 (78\%)   \\\hline
\end{tabular}
\caption{\label{tab:markedness_2vec}A confusion matrix for sonority profile types; model accuracy = \SI{61}{\percent}.}
\end{table}

\begin{figure}[H]
\centering
\includegraphics[width=0.6\textwidth]{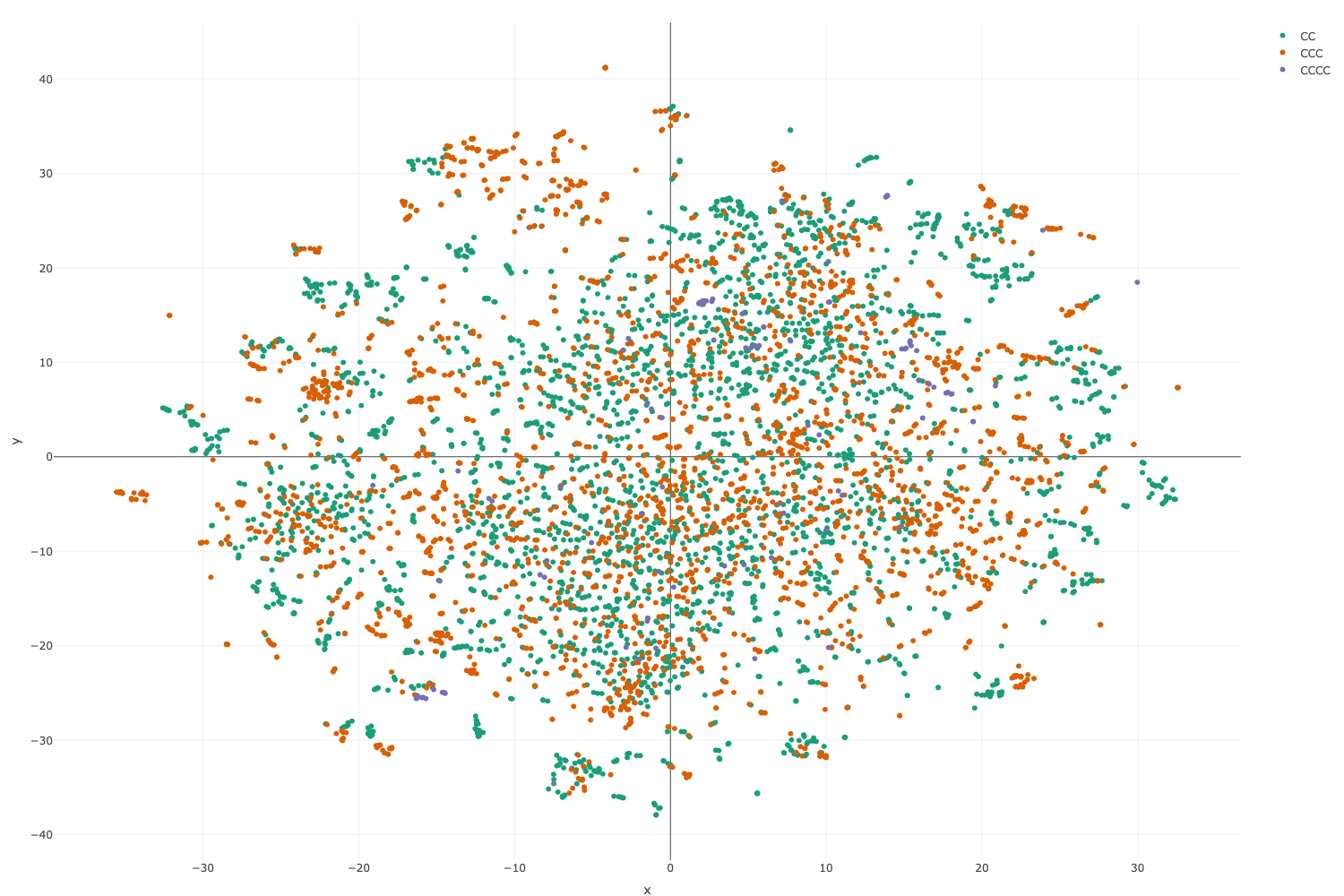}
\caption{\label{fig:Cluster_size_tsne_2vec}Scatterplot of words in a 2-dimensional t-SNE space, colour-coded for three cluster sizes.}
\end{figure}

\begin{table}
\centering
\begin{tabular}{ |c||c|c|c| } 
 \hline
            &CC             &CCC            &CCCC    \\\hline
  \hline
CC          &2584 (80\%)    &649            &15      \\\hline
CCC         &985            &1586 (61\%)    &20      \\\hline
CCCC        &24             &8              &65 (67\%) \\\hline
\end{tabular}
\caption{\label{tab:size_2vec}A confusion matrix for cluster size; model accuracy = \SI{71}{\percent}.}
\end{table}

\begin{figure}[H]
\centering
\includegraphics[width=0.6\textwidth]{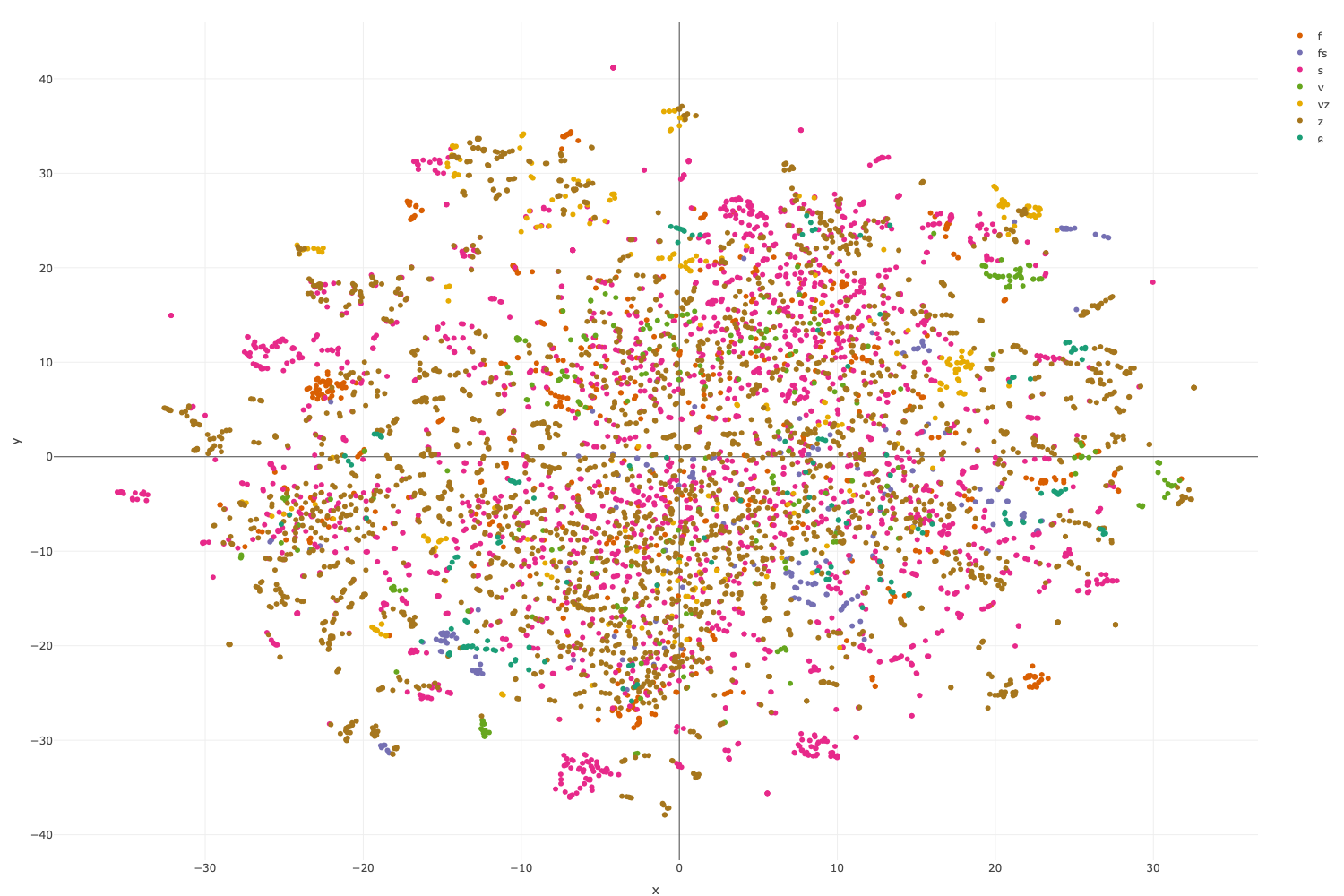}
\caption{\label{fig:Prefix_tsne_2vec}Scatterplot of words in a 2-dimensional t-SNE space, colour-coded for seven prefixe types.}
\end{figure}

\begin{table} [H]
\centering
\begin{tabular}{ |c||c|c|c|c|c|c|c| } 
 \hline  
        &\textipa{\textctc} &/f/    &/fs/   &/s/    &/v/    &/vz/   &/z/      \\\hline
      \hline
\textipa{\textctc}  &96 (63\%)    &0      &0      &57     &0      &0      &0       \\\hline
/f/                 &1      &406 (77\%)    &3      &55     &43     &3      &14     \\\hline
/fs/                &0      &14     &156 (74\%)    &10     &3      &8      &20     \\\hline
/s/                 &18     &17     &14     &2015 (96\%)  &3      &5      &30      \\\hline
/v/                 &0      &62     &5      &55     &136 (51\%)    &2      &6      \\\hline
/vz/                &0      &10     &7      &3      &1      &199 (73\%)   &52      \\\hline
/z/                 &0      &19     &9      &1      &7      &39     &2332 (97\%)   \\\hline
\end{tabular}
\caption{\label{tab:prefix_2vec}A confusion matrix for prefix type; model accuracy = \SI{90}{\percent}.}
\end{table}

\begin{figure}[H]
\centering
\includegraphics[width=1.0\textwidth]{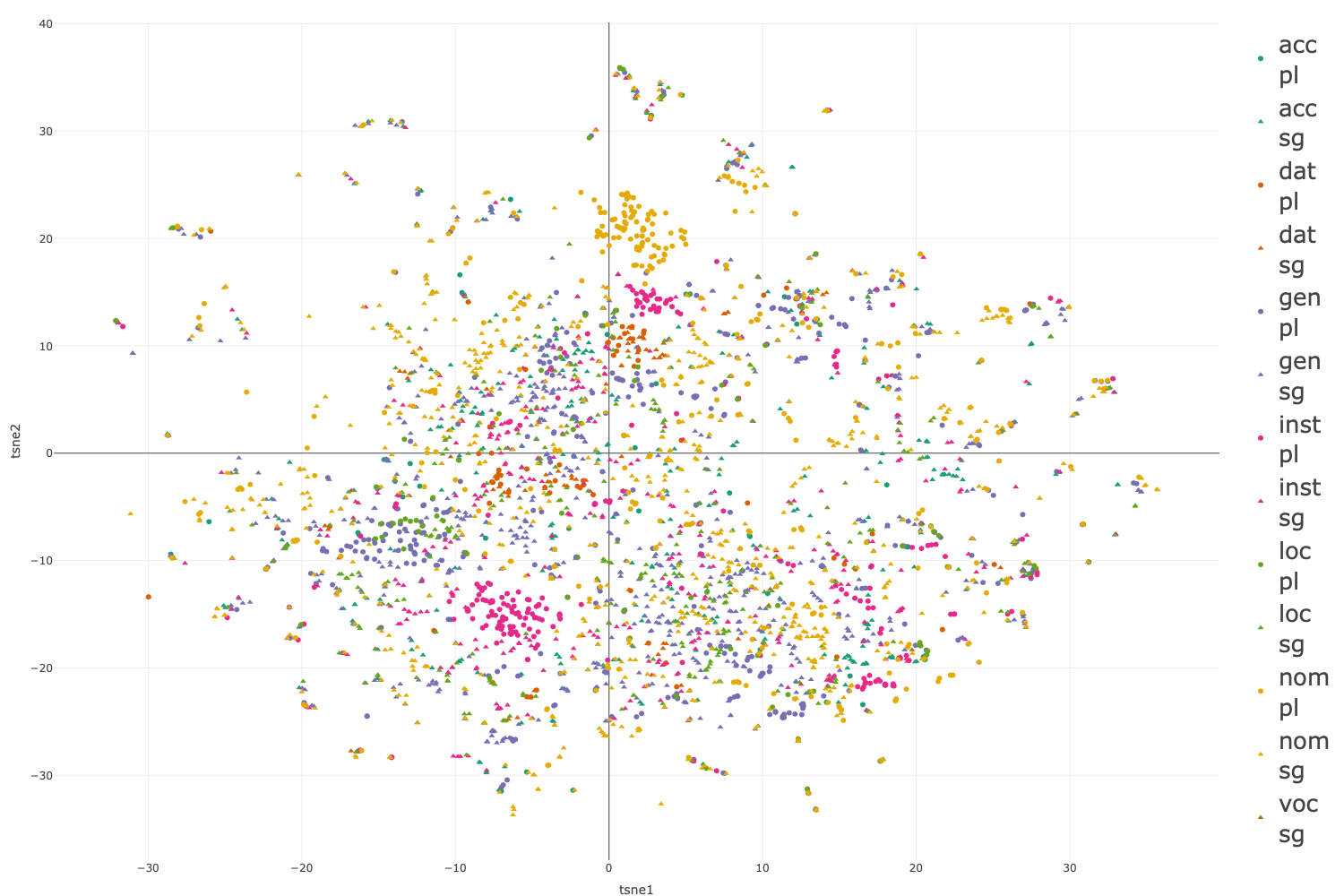}
\caption{\label{fig:casenum_2vec}Scatterplot of words in a 2-dimensional t-SNE space for \textit{Word2vec}, colour-coded for case, and symbol-coded for number. Location of plural and singular forms differs by case.}
\end{figure}

\begin{figure}
\centering
\includegraphics[width=1.0\textwidth]{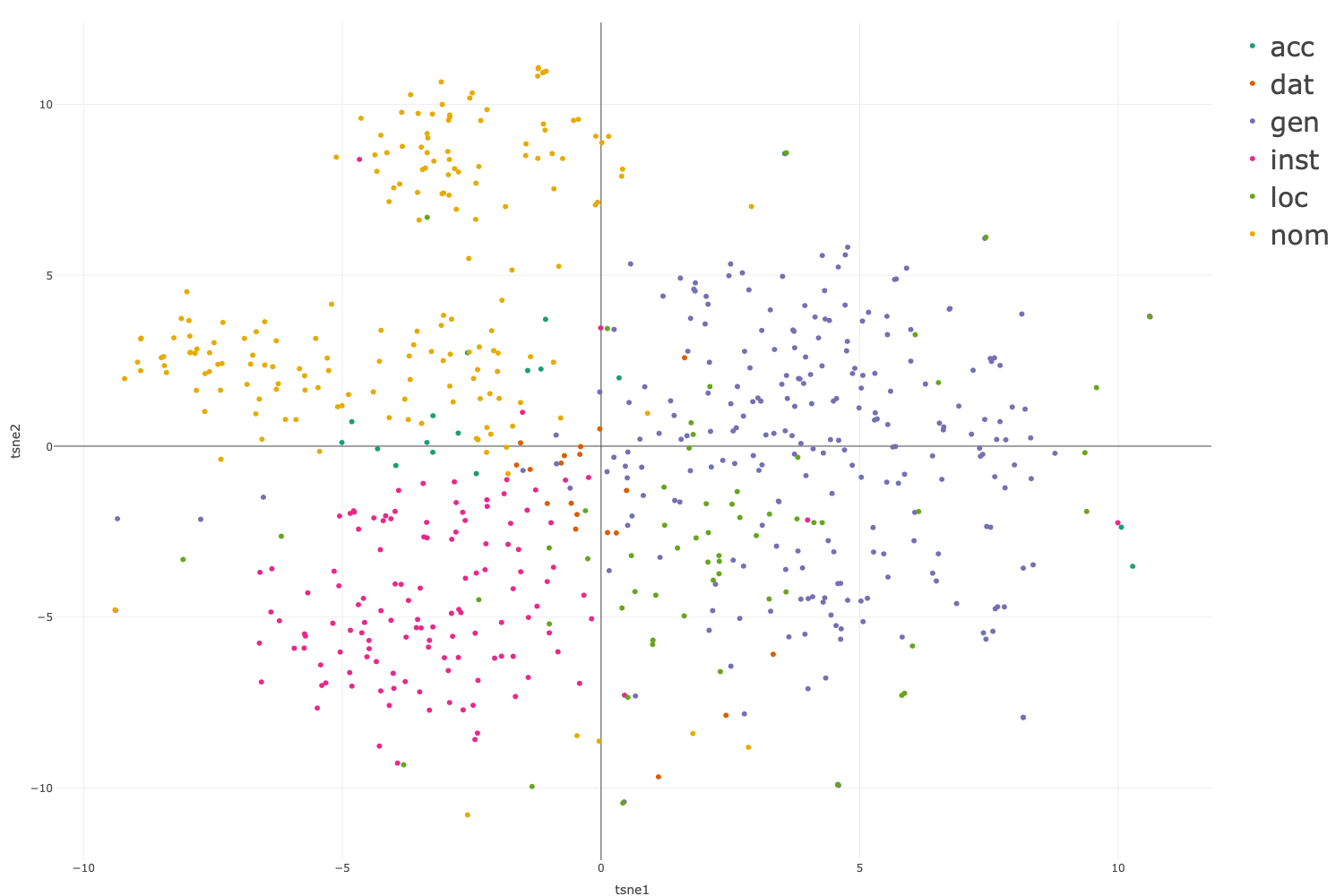}
\caption{\label{fig:shiftcectors_casenum_2vec}Scatterplot of the plural shift vectors of words in t-SNE space for \textit{Word2vec}, colour-coded for case.  Data points cluster by case.  Sub-clusters within the nominative cluster are differentiated by gender and part of speech.}
\end{figure}



\end{document}